\documentclass{article}

\usepackage[final]{neurips_2021}
\usepackage[utf8]{inputenc} 
\usepackage[T1]{fontenc}    
\usepackage{url}            
\usepackage{booktabs}       
\usepackage{amsfonts}       
\usepackage{nicefrac}       
\usepackage{microtype}      
\usepackage{epsfig}
\usepackage{amsmath}
\usepackage{mathtools}
\usepackage{caption}
\usepackage{subcaption}
\usepackage{algorithm}
\usepackage{algpseudocode}
\usepackage{authblk}
\usepackage{bbm}
\usepackage{pifont}
\usepackage{color}

\usepackage{amsmath,amsfonts,bm}

\def\eqref#1{equation~\ref{#1}}

\def\1{\bm{1}}

\def\rvx{{\mathbf{x}}}
\def\rvy{{\mathbf{y}}}
\def\rvz{{\mathbf{z}}}

\def\vf{{\bm{f}}}

\def\vv{{\bm{v}}}
\def\vw{{\bm{w}}}
\def\vx{{\bm{x}}}
\def\vy{{\bm{y}}}
\def\vz{{\bm{z}}}

\def\mW{{\bm{W}}}
\def\mX{{\bm{X}}}

\def\mZ{{\bm{Z}}}

\DeclareMathAlphabet{\mathsfit}{\encodingdefault}{\sfdefault}{m}{sl}
\SetMathAlphabet{\mathsfit}{bold}{\encodingdefault}{\sfdefault}{bx}{n}

\usepackage{multirow}
\usepackage{textcomp}
\usepackage{comment}
\usepackage{graphicx,wrapfig,lipsum}
\usepackage{diagbox}
\usepackage{amsthm}

\usepackage[x11names]{xcolor}
\usepackage{colortbl}
\usepackage[
  colorlinks,
]{hyperref}

\AtBeginDocument{\hypersetup{citecolor=DodgerBlue3}}

\usepackage{booktabs,arydshln}

\makeatletter
\def\adl@drawiv#1#2#3{%
        \hskip.5\tabcolsep
        \xleaders#3{#2.5\@tempdimb #1{1}#2.5\@tempdimb}%
                #2\z@ plus1fil minus1fil\relax
        \hskip.5\tabcolsep}
\newcommand{\cdashlinelr}[1]{%
  \noalign{\vskip\aboverulesep
           \global\let\@dashdrawstore\adl@draw
           \global\let\adl@draw\adl@drawiv}
  \cdashline{#1}
  \noalign{\global\let\adl@draw\@dashdrawstore
           \vskip\belowrulesep}}
\makeatother

\title{Rebooting ACGAN: Auxiliary Classifier GANs \\with Stable Training}
\author{%
\textbf{Minguk Kang} \hspace{1cm} \textbf{Woohyeon Shim} \hspace{1cm} \textbf{Minsu Cho} \hspace{1cm} \textbf{Jaesik Park}\\
Pohang University of Science and Technology~(POSTECH), South Korea\\
\texttt{\{mgkang, wh.shim, mscho, jaesik.park\}@postech.ac.kr} \\
}
\newcommand{\etal}{\textit{et al.~}}

\newtheorem{property}{Property}

\newcommand{\MYhref}[3][blue]{\href{#2}{\color{#1}{#3}}}
 
\begin{document}
\maketitle
\begin{abstract}
Conditional Generative Adversarial Networks~(cGAN) generate realistic images by incorporating class information into GAN. While one of the most popular cGANs is an auxiliary classifier GAN with softmax cross-entropy loss~(ACGAN), it is widely known that training ACGAN is challenging as the number of classes in the dataset increases. ACGAN also tends to generate easily classifiable samples with a lack of diversity. In this paper, we introduce two cures for ACGAN. First, we identify that gradient exploding in the classifier can cause an undesirable collapse in early training, and projecting input vectors onto a unit hypersphere can resolve the problem. Second, we propose the Data-to-Data Cross-Entropy loss~(D2D-CE) to exploit relational information in the class-labeled dataset. On this foundation, we propose the Rebooted Auxiliary Classifier Generative Adversarial Network~(ReACGAN). The experimental results show that ReACGAN achieves state-of-the-art generation results on CIFAR10, Tiny-ImageNet, CUB200, and ImageNet datasets. We also verify that ReACGAN benefits from differentiable augmentations and that D2D-CE harmonizes with StyleGAN2 architecture.~Model weights and a software package that provides implementations of representative cGANs and all experiments in our paper are available at~\MYhref[magenta]{https://github.com/POSTECH-CVLab/PyTorch-StudioGAN}{https://github.com/POSTECH-CVLab/PyTorch-StudioGAN}.
\end{abstract}
\section{Introduction}
Generative Adversarial Networks~(GAN)~\cite{Goodfellow2014GenerativeAN} are known for the forefront approach to generating high-fidelity images of diverse categories~\cite{Radford2016UnsupervisedRL, Miyato2018SpectralNF, Brock2019LargeSG, karras2019style, Wu2019LOGANLO, karras2020analyzing, Karras2020TrainingGA, zhao2020differentiable}. Behind the sensational generation ability of GANs, there has been tremendous effort to develop adversarial objectives free from the vanishing gradient problem~\cite{Mao2017LeastSG, Arjovsky2017WassersteinG, Arjovsky2017TowardsPM}, regularizations for stabilizing adversarial training~\cite{Arjovsky2017WassersteinG, Gulrajani2017ImprovedTO, Kodali2018OnCA, Miyato2018SpectralNF, zhou2019lipschitz, Zhang2019ConsistencyRF, Zhao2020ImprovedCR}, and conditioning techniques to support the adversarial training using category information of the dataset~\cite{Odena2017ConditionalIS, Miyato2018cGANsWP, NIPS2019_8414, Siarohin2019WhiteningAC, Liu_2019_ICCV, kang2020contragan, shim2020circlegan, zhou2020omni, kavalerov2021multi, hou2021cgans, Han_2021_ICCV}. Subsequently, the conditioning techniques have become the \emph{de~facto} standard for high-quality image generation. The models with the conditioning methods are called conditional Generative Adversarial Networks~(cGAN), and cGANs can be divided into two groups depending on the discriminator's conditioning way:~classifier-based GANs~\cite{Odena2017ConditionalIS, NIPS2019_8414, kang2020contragan, zhou2020omni, hou2021cgans} and projection-based GANs~\cite{Miyato2018cGANsWP, Miyato2018SpectralNF, Brock2019LargeSG, Han_2021_ICCV}. 

The classifier-based GANs facilitate an auxiliary classifier to generate class-specific images by penalizing the generator if the synthesized images are not consistent with the conditioned labels. ACGAN~\cite{Odena2017ConditionalIS} 
has been one of the widely used classifier-based GANs for its simple design and satisfactory generation performance. While ACGAN can exploit class information by pushing and pulling classifier's weights~(proxies) against image embeddings~\cite{kang2020contragan}, it is well known that ACGAN training is prone to collapsing at the early stage of training as the number of classes increases~\cite{Miyato2018cGANsWP, kang2020contragan, zhou2020omni, hou2021cgans}. In addition, the generator of ACGAN tends to generate easily classifiable images at the cost of reduced diversity~\cite{Odena2017ConditionalIS, Miyato2018cGANsWP, hou2021cgans}.
Projection-based GANs, on the other hand, have shown cutting-edge generation results on datasets with a large number of categories. SNGAN~\cite{Miyato2018SpectralNF}, SAGAN~\cite{Zhang2019SelfAttentionGA}, and BigGAN~\cite{Brock2019LargeSG} are representatives in this family and can generate realistic images on CIFAR10~\cite{Krizhevsky2009LearningML} and ImageNet~\cite{Deng2009ImageNetAL} datasets. However, projection-based GANs only consider a pairwise relationship between an image and its label proxy~(data-to-class relationships). As the result, the projection-based GANs can miss an additional opportunity to consider relation information between data instances~(data-to-data relationships) as discovered by~\cite{kang2020contragan}.

\begin{figure}[t]
    \centering
    \includegraphics[width=0.93\linewidth]{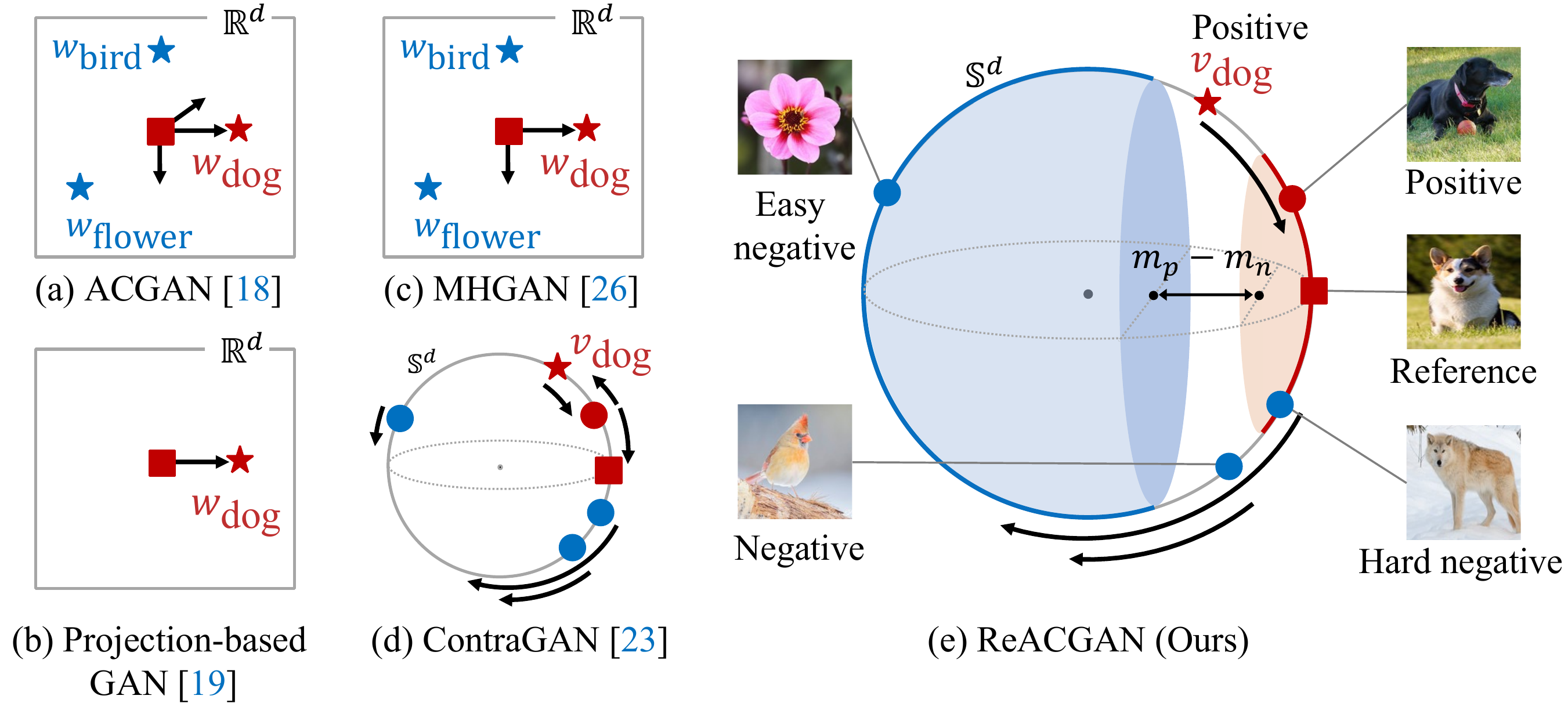}
    \caption{Schematics that depict how cGANs perform conditioning. The red color means positive samples/proxy, and the blue color indicates negative samples/proxies. Arrows represent push-pull forces based on the reference sample. The length of an arrow indicates the magnitude of the force.}
    \label{fig:Figure_cGANs}
\end{figure}
In this paper, we analyze why ACGAN training becomes unstable as the number of classes increases and propose remedies for (1) the instability and (2) the relatively poor generation performance of ACGAN compared with the projection-based models. First, we begin by analytically deriving the gradient of the softmax cross-entropy loss used in ACGAN. By examining the exact values of analytic gradients, we discover that the unboundedness of input feature vectors and poor classification performance in the early training stage can cause an undesirable gradient exploding problem. Second, with alleviating the instability, we propose the Rebooted Auxiliary Classifier Generative Adversarial Networks~(ReACGAN) using the Data-to-Data Cross-Entropy loss~(D2D-CE). ReACGAN projects image embeddings and proxies onto a unit hypersphere and computes similarities for data-to-data and data-to-class consideration. Additionally, we introduce two margin values for intra-class variations and inter-class separability. In this way, ReACGAN overcomes the training instability and can exploit additional supervisory signals by explicitly considering data-to-class and data-to-data relationships, and also by implicitly looking at class-to-class relationships in the same mini-batch.

To validate our model, we conduct image generation experiments on CIFAR10~\cite{Krizhevsky2009LearningML}, Tiny-ImageNet~\cite{Tiny}, CUB200~\cite{WelinderEtal2010}, and ImageNet~\cite{Deng2009ImageNetAL} datasets. Through extensive experiments, we demonstrate that ReACGAN beats both the classifier-based and projection-based GANs, improving over the state of the art by 2.5\%, 15.8\%, 5.1\%, and 14.5\% in terms of Fréchet Inception Distance~(FID)~\cite{Heusel2017GANsTB} on the four datasets, respectively. We also verify that ReACGAN benefits from consistency regularization~\cite{Zhang2019ConsistencyRF} and differentiable augmentations~\cite{zhao2020differentiable, Karras2020TrainingGA} for limited data training. Finally, we confirm that D2D-CE harmonizes with the StyleGAN2 architecture~\cite{karras2020analyzing}.
\section{Background: Generative Adversarial Networks}
\label{sec:Backgroud_GAN}
Generative Adversarial Network~(GAN)~\cite{Goodfellow2014GenerativeAN} is an implicit generative model that aims to generate a sample indistinguishable from the real. GAN consists of two networks: a \emph{Generator} $G: \mathcal{Z}\longrightarrow\mathcal{X}$ that tries to map a latent variable~$\rvz \sim p(\rvz)$ into the real data space~$\mathcal{X}$ and a \emph{Discriminator} $D:\mathcal{X}\longrightarrow[0,1]$ that strives to discriminate whether a given sample$~\rvx$ is from the real data distribution~$p_{\text{real}}{(\rvx)}$ or from the implicit distribution~$p_{\text{gen}}{(G(\rvz))}$ derived from the generator~$G(\rvz)$. The objective of a vanilla GAN~\cite{Goodfellow2014GenerativeAN} can be expressed as follows:
\begin{align*}
    \min_{G}\max_{D}~\mathbb{E}_{\rvx \sim {p(\rvx)}}[\log(D(\rvx))] + \mathbb{E}_{\rvz \sim {p(\rvz)}}[\log(1 - D(G(\rvz)))].
    \label{eq:eq1}\tag{1}
\end{align*}
While GANs have shown impressive results in the image generation task~\cite{Radford2016UnsupervisedRL, Nowozin2016fGANTG, Arjovsky2017WassersteinG, Gulrajani2017ImprovedTO}, training GANs often ends up encountering a mode-collapse problem~\cite{srivastava2017veegan, Arjovsky2017WassersteinG, Arjovsky2017TowardsPM}. As one of the prescriptions for stabilizing and reinforcing GANs, training GANs with categorical information, named conditional Generative Adversarial Networks~(cGAN), is suggested~\cite{Mirza2014ConditionalGA, Odena2017ConditionalIS, Miyato2018cGANsWP}. Depending on the presence of explicit classification losses, cGAN can be divided into two groups: Classifier-based GANs~\cite{Odena2017ConditionalIS, NIPS2019_8414, kang2020contragan, zhou2020omni, hou2021cgans} and Projection-based GANs~\cite{Miyato2018cGANsWP, Miyato2018SpectralNF, Brock2019LargeSG, Han_2021_ICCV}. One of the widely used classifier-based GANs is ACGAN~\cite{Odena2017ConditionalIS}, and ACGAN utilizes softmax cross-entropy loss to perform classification task with adversarial training. 
Although ACGAN has shown satisfactory generation results, training ACGAN becomes unstable as the number of classes in the training dataset increases~\cite{Miyato2018cGANsWP, kang2020contragan, zhou2020omni, hou2021cgans}. Besides, ACGAN tends to generate easily classifiable images at the cost of limited diversity~\cite{Odena2017ConditionalIS, Miyato2018cGANsWP, hou2021cgans}. To alleviate those problems, Zhou~\etal~\cite{zhou2018activation} have proposed performing adversarial training on the classifier. Gong~\etal~\cite{NIPS2019_8414} have introduced
an additional classifier to eliminate a conditional entropy minimization process in the adversarial training. However, ACGAN training still suffers from the early-training collapse issue and the reduced diversity problem when trained on datasets with a large number of class categories, such as Tiny-ImageNet~\cite{Tiny} and ImageNet~\cite{Deng2009ImageNetAL}.

In these circumstances, Miyato~\etal~\cite{Miyato2018cGANsWP} have proposed a projection discriminator for cGANs and have shown significant improvement in generating the ImageNet dataset. Motivated by the promising result of the projection discriminator, many projection-based GANs~\cite{Miyato2018SpectralNF, Zhang2019SelfAttentionGA, Brock2019LargeSG, Zhang2019ConsistencyRF, Wu2019LOGANLO, Zhao2020ImprovedCR} have been proposed and become the standard for conditional image generation. In this paper, we revisit ACGAN and unveil why ACGAN training is so unstable. Coping with the instability, we propose the Rebooted Auxiliary Classifier GANs~(ReACGAN) for high-quality and diverse image generation.
\section{Rebooting Auxiliary Classifier GANs}
\label{sec:ACGAN}

\subsection{Feature Normalization}
\label{sec:Feature_Normalize}
\begin{figure}[t]
    \centering
    \begin{subfigure}{0.32\textwidth}
    \includegraphics[width=\linewidth]{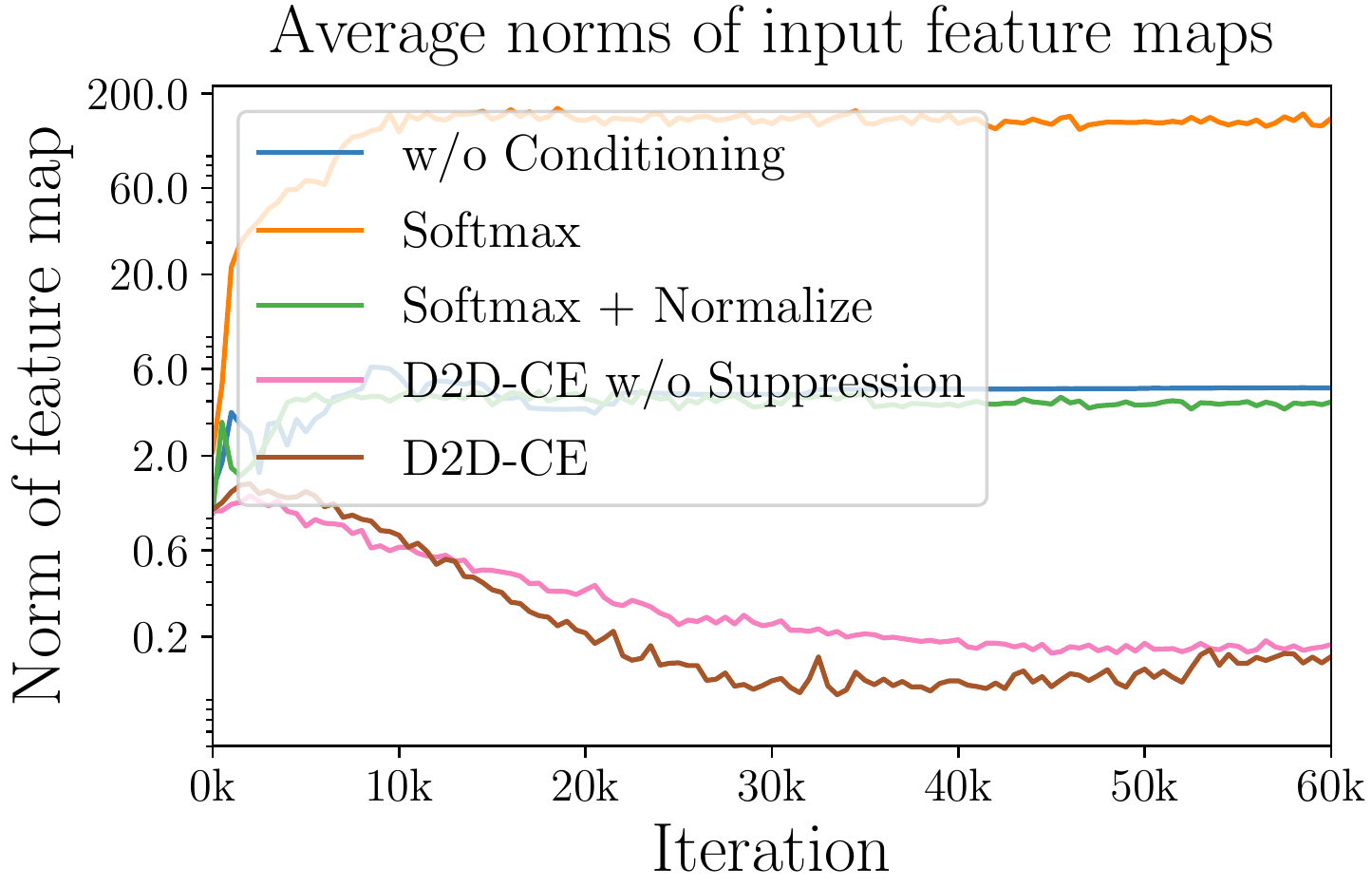}
    \caption{Feature norm} \label{fig:acgan_fnorm}
    \end{subfigure}
    \begin{subfigure}{0.32\textwidth}
    \includegraphics[width=\linewidth]{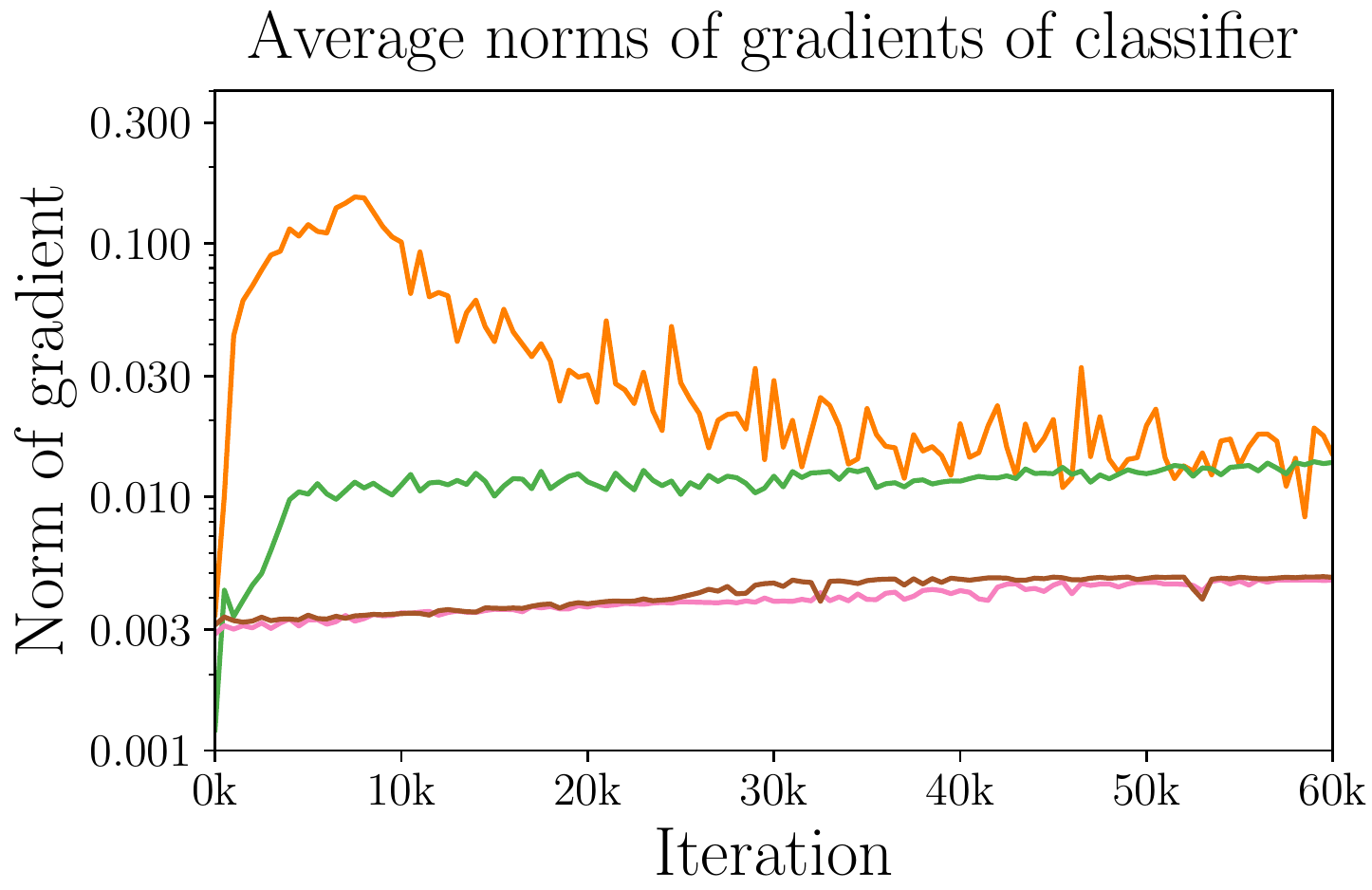}
    \caption{Gradient norm} \label{fig:acgan_gnorm}
    \end{subfigure}
    \begin{subfigure}{0.32\textwidth}
    \includegraphics[width=\linewidth]{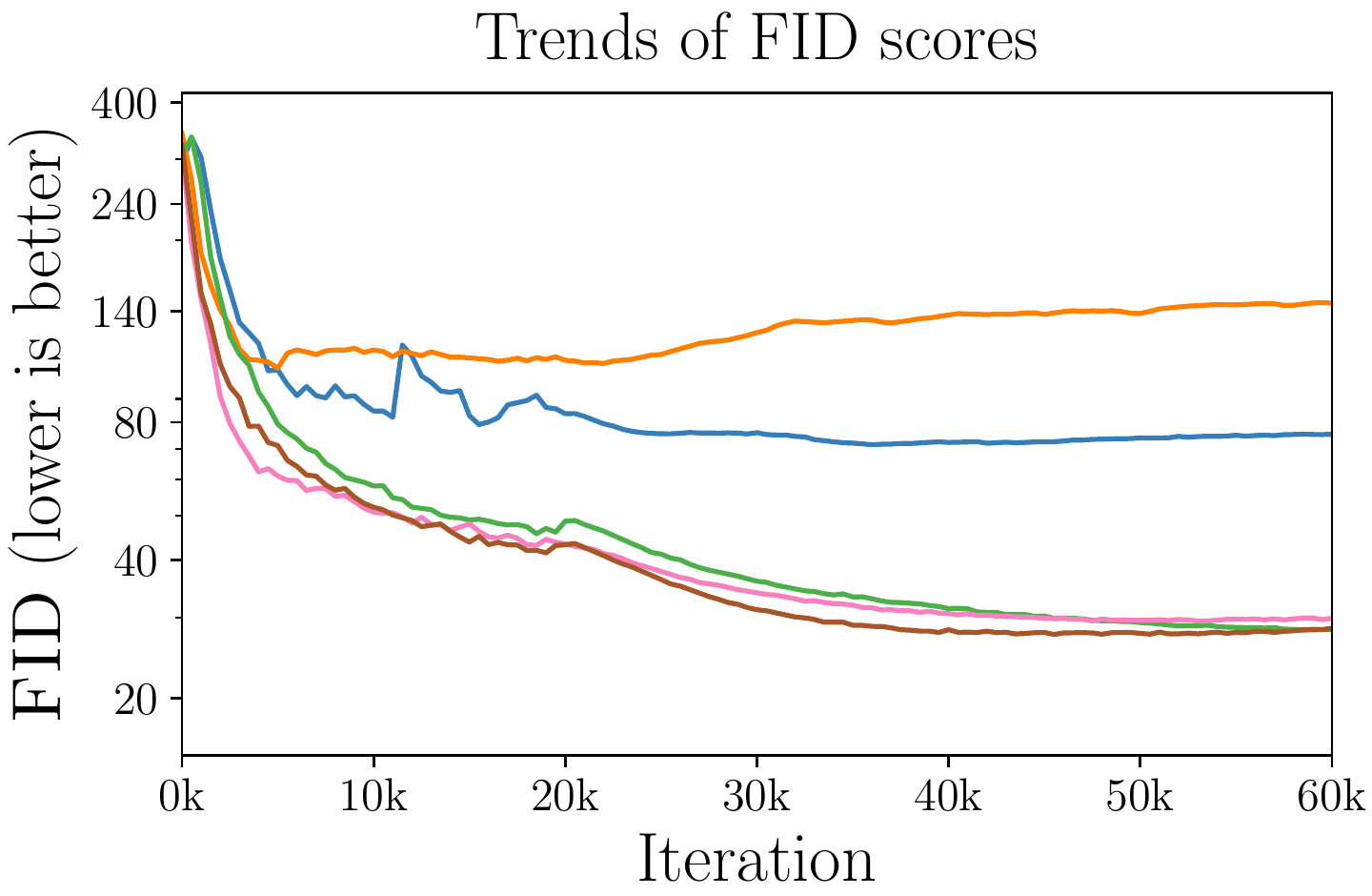}
    \caption{FID score} \label{fig:acgan_fid}
    \end{subfigure}
    \caption{Merits of integrating feature normalization and data-to-data relationship consideration. All experiments are conducted on Tiny-ImageNet~\cite{Tiny} dataset. (a) Average norms of input feature maps $F(\cdot)$, (b) average norms of gradients of classification losses, and (c) trends of FID scores. Compared with ACGAN~\cite{Odena2017ConditionalIS}, the proposed ReACGAN does not experience the training collapse problem caused by excessively large norms of feature maps and gradients at the early training stage. In addition, ReACGAN can converge to better equilibrium by considering data-to-data relationships with easy positive and negative sample suppression.}
    \label{fig:Figure_exploding}
\end{figure}

 To uncover a nuisance that can cause the instability of ACGAN, we start by analytically deriving the gradients of weight vectors in the softmax classifier. Let the part of the discriminator before the fully connected layer be a $\text{\emph{Feature extractor}}~F:\mathcal{X}\longrightarrow\ \mathcal{F} \in \mathbb{R}^d$ and let \emph{Classifier}~$C:\mathcal{F}\longrightarrow\ \mathbb{R}^c$ be a single fully connected layer parameterized by~$\mW = [\vw_{1},...,\vw_{c}]\in\mathbb{R}^{d \times c}$, where $c$ denotes the number of classess. We sample training images~$\mX=\{\vx_1,...,\vx_{N}\}$ and integer labels~$\vy=\{y_1,...,y_{N}\}$ from the joint distribution~$p(\rvx, \rvy)$. Using the notations above, we can express the empirical cross-entropy loss used in ACGAN~\cite{Odena2017ConditionalIS} as follows:
\begin{align*}
    \mathcal{L_{\text{CE}}} = -
    \frac{1}{N} \sum_{i=1}^{N}\log{\Big(\frac{\exp{(F(\vx_i)^\top  \vw_{y_i})}}
    {\sum_{j=1}^{c}\exp{(F(\vx_i)^\top \vw_{j})}}\Big)}.
    \label{eq:eq3}\tag{3}
\end{align*}
Based on Eq.~(\ref{eq:eq3}), we can derive the derivative of the cross-entropy loss, w.r.t $\vw_{k \in \{1,...,c\}}$ as follows:
\begin{align*}
    \frac{\partial \mathcal{L_{\text{CE}}}}{\partial \vw_{k}}=
    -\frac{1}{N}\sum_{i=1}^{N}\bigg\{ F(\vx_i)\bigg(\1_{y_i=k} - 
    p_{i,k}\bigg)\bigg\},
    \label{eq:eq4}\tag{4}
\end{align*}
where $\1_{y_i=k}$ is an indicator function that will output 1 if $y_i = k $ is satisfied, and $p_{i,k}$ is a class probability that represents the probability that $i$-th sample belongs to class $k$, mathematically $\frac{\exp{(F(\vx_i)^\top \vw_{k})}}{\sum_{j=1}^{c}\exp{(F(\vx_i)^\top \vw_{j})}}$.
The equation above implies that the norm of the gradient of the softmax cross-entropy loss is coupled with the norms and directions of each input feature map~$F(\vx_{i})$ and the class probabilities. In the early training stage, the classifier is prone to making incorrect predictions, resulting in low probabilities. This phenomenon occurs more frequently as the number of categories in the dataset increases. As the result, the norm of the gradient~$|\frac{\partial \mathcal{L_{\text{CE}}}}{\partial \vw_{k}}|$ begins to explode as the vector $F(\vx_i)$ stretches out to $\vw_{k}$ direction but being located close to the the other vectors $\vw_{j\in\{1,...,c\}\backslash \{k\}}$. This often breaks the balance between adversarial learning and classifier training, leading to an early-training collapse. Once the early-training collapse occurs, ACGAN training concentrates on classifying categories of images instead of discriminating the authenticity of given samples. We experimentally demonstrate that the average norm of ACGAN's input feature maps increases as the training progresses (Fig.~\ref{fig:acgan_fnorm}).
Accordingly, the average norm of the gradients increases sharply at the early training stage and decreases with the high class probabilities of the classifier~(Fig.~\ref{fig:acgan_gnorm} in the main paper and Fig.~\ref{fig:acgan_gnorm_sp} and~\ref{fig:biased_p} in Appendix~\ref{additional_experiments}). While the average norm of gradients decreases at some point, the FID value of ACGAN does not decrease, implying the collapse of ACGAN training~(Fig.~\ref{fig:acgan_fid}).

As one of the remedies for the gradient exploding problem, we find that simply \emph{normalizing the feature embeddings onto a unit hypersphere}~$\frac{F(\vx_i)}{||F(\vx_i)||}$ effectively resolves ACGAN's early-training collapse~(see Fig.~\ref{fig:Figure_exploding}). The motivation is that normalizing the features onto the hypersphere makes the norms of feature maps equal to 1.0. Thus, the discriminator does not experience the gradient exploding problem. From the next section, we will deploy a linear projection layer~$P$ on the feature extractor~$F$. And, we will normalize both the embeddings from the projection layer and the weight vectors $(\vw_{1},...,\vw_{c})$ in the classifier since normalizing both the embeddings and the weight vectors does not degrade image generation performance. We denote the normalized embedding $\frac{P(F(\vx_i))}{||P(F(\vx_i))||}$ as $\vf_i$ and the normalized weight vector $\frac{\vw_{y_i}}{||\vw_{y_i}||}$ as $\vv_{y_i}$.

\subsection{Data-to-Data Cross-Entropy Loss~(D2D-CE)}
\label{sec:Dada_Softmax}
We expand the feature normalized softmax cross-entropy loss described in Sec.~\ref{sec:Feature_Normalize} to the Data-to-Data Cross-Entropy~(D2D-CE) loss. The motivations are summarized into two points: (1) replacing data-to-class similarities in the denominator of Eq.~(\ref{eq:eq3}) with data-to-data similarities and (2) introducing two margin values to the modified softmax cross-entropy loss. We expect that point (1) will encourage the feature extractor to consider data-to-class as well as data-to-data relationships, and that point (2) will guarantee inter-class separability and intra-class variations in the feature space while preventing ineffective gradient updates induced by easy negative and positive samples.
\begin{figure}[t!]
    \centering
    \includegraphics[width=0.96\linewidth]{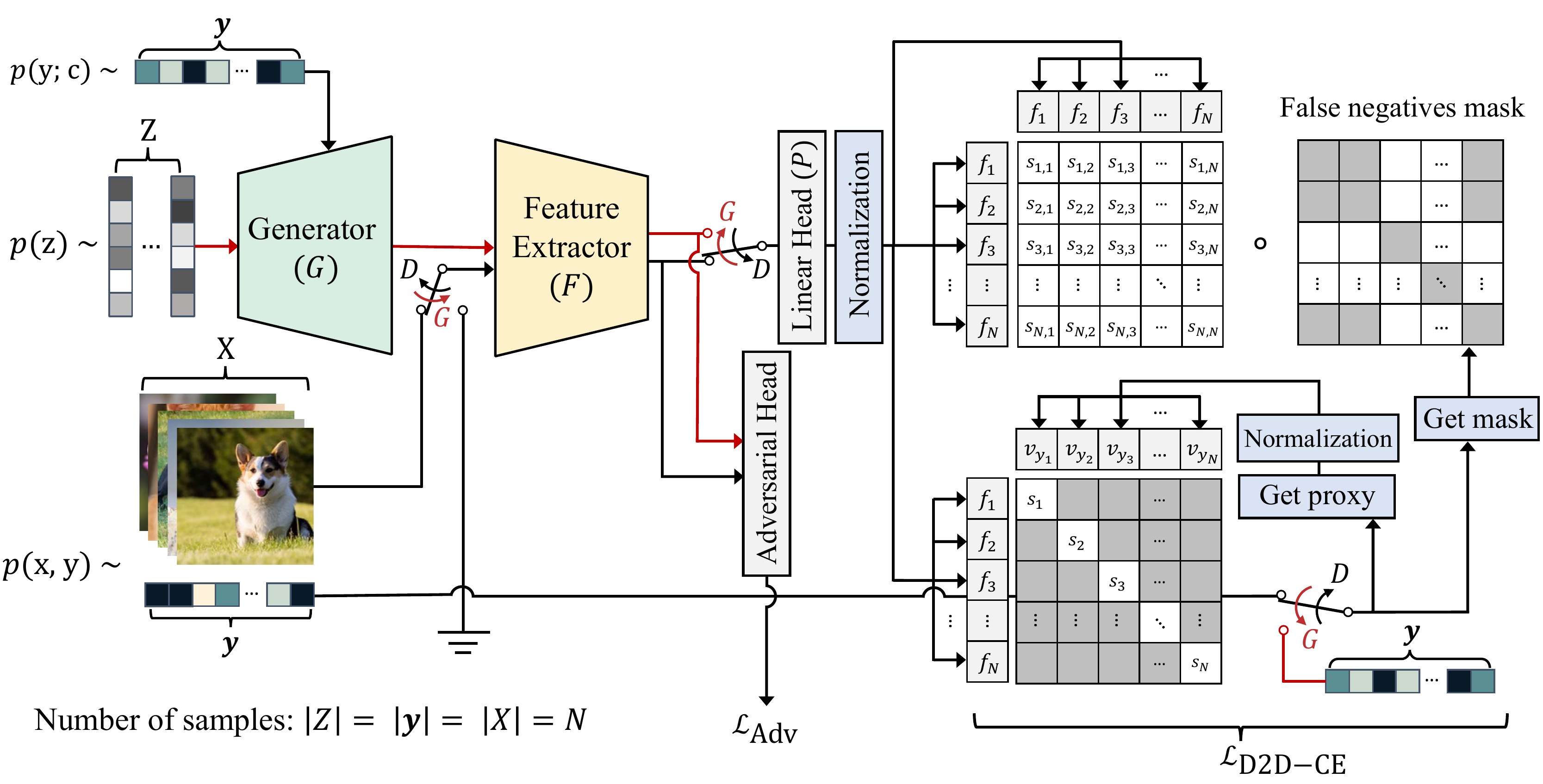}
    \caption{Overview of ReACGAN. ReACGAN performs adversarial training using the loss $\mathcal{L}_{\text{Adv}}$. At the same time, ReACGAN tries to minimize data-to-data cross-entropy loss~(D2D-CE) on the linear head ($P$) to exploit relational information in the labeled dataset. $\scriptstyle{\circ}$ means element-wise product. Note that \emph{False negatives mask} operates on the similarity matrix between two batches of sample embeddings to compute the similarities between negative samples in the denominator of Eq.~(\ref{eq:eq6}).}
    \label{fig:Figure_ReACGAN}
\end{figure}
To develop the feature normalized cross-entropy loss into D2D-CE, we replace the similarities between a sample embedding and all proxies except for the positive one in the denominator,~$\sum_{j\in \{1,...,c\} \backslash \{y_i\}}\exp{(\vf_i^\top \vv_j)}$ with similarities between negative samples in the same mini-batch. The modified cross-entropy loss can be expressed as follows:
\begin{align*}
    \mathcal{L_{\text{CE}}^{'}} = -
    \frac{1}{N}\sum_{i=1}^{N}\log{\bigg(\frac{\exp{(\vf_i^\top \vv_{y_i}/\tau)}}
    {\exp{(\vf_i^\top\vv_{y_{i}}}/\tau)+\sum_{j\in\mathcal{N}(i)}\exp{(\vf_i^\top\vf_j/\tau)}}\bigg)},
    \label{eq:eq5}\tag{5}
\end{align*}
where $\tau$ is a temperature, and $\mathcal{N}(i)$ is the set of indices that point locations of the negative samples whose labels are different from the reference label $\vv_{y_i}$ in the mini-batch. The self-similarity matrix of samples in the mini-batch is used to calculate the similarities between negative samples~$\vf_i^\top\vf_{j\in\mathcal{N}(i)}$ with a false negative mask~(see~Fig.~\ref{fig:Figure_ReACGAN}). 
Thus, Eq.~(\ref{eq:eq5}) enables the discriminator to contrastively compare visual differences between multiple images and can supply more informative supervision for image conditioning. Finally, we introduce two margin hyperparameters to $\mathcal{L_{\text{CE}}^{'}}$ and name it \emph{Data-to-Data Cross-Entropy loss}~(D2D-CE). The proposed D2D-CE can be expressed as follows:
\begin{align*}
    \mathcal{L_{\text{D2D-CE}}}=-\frac{1}{N}\sum_{i=1}^{N}\log{\bigg(\frac{\exp{\big([\vf_i^\top\vv_{y_i}-m_p]_{-}/\tau}\big)}
    {\exp{\big([\vf_i^\top\vv_{y_i}-m_p]_{-}/\tau}\big)+\sum_{j\in \mathcal{N}(i)}\exp{\big([\vf_i^\top \vf_j-m_n]_{+}/\tau\big)}}\bigg)},
    \label{eq:eq6}\tag{6}
\end{align*}
where $m_p$ is a margin for suppressing a high similarity value between a reference sample and its corresponding proxy~(easy positive), $m_n$ is a margin for suppressing low similarity values between negatives samples~(easy negatives). $[\cdot]_{-}$ and $[\cdot]_{+}$ denote min($\cdot$,0) and max($\cdot$,0) functions, respectively. 

\subsection{Useful Properties of Data-to-Data Cross-Entropy Loss}
\label{sec:Properties}
\begin{wrapfigure}{r}{0.47\textwidth}
  \vspace{-25pt}
  \begin{center}
    \includegraphics[width=0.45\textwidth]{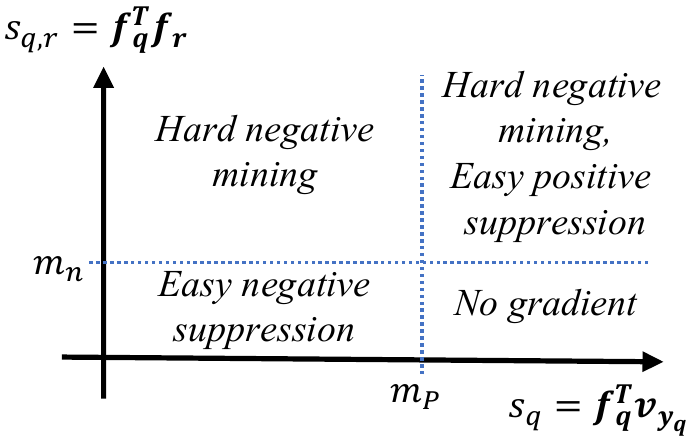}
  \end{center}
  \vspace{-12pt}
  \caption{Graph showing regions where hard negative mining, easy positive and negative suppression, and no gradient update occur.}
  \vspace{-10pt}
\end{wrapfigure}
In this subsection, we explain four useful properties of D2D-CE. Let $s_{q}$ be a similarity between the normalized embedding $\vf_q$ and the corresponding normalized proxy $\vv_{y_q}$, $s_{q, r}$ be a similarity between $\vf_q$ and $\vf_r$, and $a$ and $b$ be arbitrary indices of negative samples, \textit{i.e.}, $a, b \in\mathcal{N}(q)$. Then the properties of D2D-CE can be summarized as follows:

\begin{property}
Hard negative mining.
If the value of $s_{q, a}$ is greater than $s_{q, b}$, the derivative of $\mathcal{L}_{\text{D2D-CE}}$ w.r.t $s_{q, a}$ is greater than or equal to the derivative w.r.t $s_{q, b}$; that is $\frac{\partial{\mathcal{L}_{\text{D2D-CE}}}}{\partial{s_{q, a}}} \geq \frac{\partial{\mathcal{L}_{\text{D2D-CE}}}}{\partial{s_{q, b}}}\geq 0$.
\end{property}

\begin{property}
Easy positive suppression.
If $s_{q} - m_p \geq 0$, the derivative of  $\mathcal{L}_{\text{D2D-CE}}$ w.r.t $s_{q}$ is $0$.
\end{property}

\begin{property}
Easy negative suppression. 
If $s_{q, r} -m_n \leq 0$, the derivative of  $\mathcal{L}_{\text{D2D-CE}}$ w.r.t $s_{q, r}$ is $0$.
\end{property}

\begin{property}
If $s_{q} - m_p \geq 0$ and $s_{q, r} - m_n \leq 0$ are satisfied, $\mathcal{L}_{\text{D2D-CE}}$ has the global minima of $~\frac{1}{N}\sum_{i=1}^{N}\log{(1 + |\mathcal{N}(i)|)}$.
\end{property}

We put proofs of the above properties in Appendix~\ref{proof_of_D2DCE}. Property 1 indicates that D2D-CE implicitly conducts hard negative mining and benefits from comparing samples with each other. Also, Properties 2 and 3 imply that samples will not affect gradient updates if the samples are trained sufficiently. Consequently, the classifier concentrates on pushing and pulling hard negative and hard positive examples without being dominated by easy negative and positive samples.

\subsection{Rebooted Auxiliary Classifier Generative Adversarial Networks~(ReACGAN)}
With the proposed D2D-CE objective, we propose the \textit{Rebooted Auxiliary Classifier Generative Adversarial Networks}~(ReACGAN). As ACGAN does, ReACGAN jointly optimizes an adversarial loss and the classification objective (D2D-CE). Specifically, the discriminator, which consists of the feature extractor, adversarial head, and linear head, of ReACGAN tries to discriminate whether a given image is sampled from the real distribution or not. At the same time, the discriminator tries to maximize similarities between the reference samples and corresponding proxies while minimizing similarities between negative samples using real images and D2D-CE loss. After updating the discriminator a predetermined number of times, the generator strives to deceive the discriminator by generating well-conditioned images that will output a low D2D-CE value. By alternatively training the discriminator and generator until convergence, ReACGAN generates high-quality images of diverse categories without early-training collapse. We attach the algorithm table in Appendix~\ref{algorithm}. 

\textbf{Differences between ReACGAN and ContraGAN.}
The authors of ContraGAN~\cite{kang2020contragan} propose a conditional contrastive loss~(2C loss) to cover data-to-data relationships when training cGANs. The main differences between 2C loss and D2D-CE objective are summed up into three points: (1) while 2C loss is derived from NT-Xent loss~\cite{Chen2020ASF}, which is popularly used in the field of the self-supervised learning, our D2D-CE is developed to resolve the early-training collapsing problem and the poor generation results  of ACGAN, (2) 2C loss holds false-negative samples in the denominator, and they can cause unexpected positive repulsion forces, and (3) 2C loss contains the similarities between all positive samples in the numerator, and they give rise to large gradients on pulling easy positive samples. Consequently, GANs with 2C loss tend to synthesize images of unintended classes as reported in the author's software document~\cite{studiogan}. However, D2D-CE does not contain the false negatives in the denominator and considers only the similarity between a sample and its proxy in the numerator. Therefore, ReACGAN is free from the undesirable repelling forces and does not conduct unnecessary easy positive mining. More detailed explanations are attached in Appendix~\ref{difference_reac_contra}.

\textbf{Consistency Regularization and D2D-CE Loss.} Zhang~\etal~\cite{Zhang2019ConsistencyRF} propose a consistency regularization to force the discriminator to make consistent predictions if given two images are visually close to each other. They create a visually similar image pair by augmenting a reference image with pre-defined augmentations. After that, they let the discriminator minimize L2 distance between the logit of the reference image and the logit of the augmented counterpart. While ReACGAN locates an image embedding nearby its corresponding proxy but far apart multiple image embeddings of different classes, consistency regularization only pulls a reference and the augmented image towards each other. Since consistency regularization and D2D-CE can be applied together, we will show that ReACGAN benefits from consistency regularization in the experiments section~(Table~\ref{table:main_table1}).
\section{Experiments}
\subsection{Datasets and Evaluation Metrics}
\label{sec:experiment_datasets}
To verify the effectiveness of ReACGAN, we conduct conditional image generation experiments using five datasets: \text{CIFAR10}~\cite{Krizhevsky2009LearningML}, \text{Tiny-ImageNet}~\cite{Tiny}, \text{CUB200}~\cite{WelinderEtal2010}, \text{ImageNet}~\cite{Deng2009ImageNetAL}, and \text{AFHQ}~\cite{choi2020starganv2} datasets and four evaluation metrics: Inception Score~(\text{IS})~\cite{Salimans2016ImprovedTF}, Fr\'echet Inception Distance~(\text{FID})~\cite{Heusel2017GANsTB}, and $\text{F}_{\text{0.125}}$~(Precision) and $\text{F}_{\text{8}}$~(Recall)~\cite{sajjadi2018assessing}. The details on the training datasets are in Appendix~\ref{dataset}.

\textbf{Inception Score (IS)}~\cite{Salimans2016ImprovedTF}\textbf{~and Fr\'echet Inception Distance (FID)}~\cite{Heusel2017GANsTB} are widely used metrics for evaluating generative models. We utilize IS and FID together because some studies~\cite{Brock2019LargeSG, Wu2019LOGANLO, zhou2020omni} have shown that IS has a tendency to measure the fidelity of images better while FID tends to weight capturing the diversity of images.

\textbf{Precision~($\text{F}_{\text{0.125}}$) and Recall~($\text{F}_{\text{8}}$)}~\cite{sajjadi2018assessing} are metrics for estimating precision and recall of the approximated distribution~$p(G(\rvz ))$ against the true data distribution~$p(\rvx)$. Instead of evaluating generative models using one-dimensional scores, such as IS and FID, Sajjadi~\etal~\cite{sajjadi2018assessing} have suggested using a two-dimensional score~$\text{F}_{\text{0.125}}$ and $\text{F}_{\text{8}}$ that can quantify how precisely the generated images are and how well the generated images cover the reference distribution.

\subsection{Experimental Details}
\label{sec:experimental_details}
The implementation of ReACGAN basically follows details of PyTorch-StudioGAN library~\cite{studiogan}\footnote{PyTorch-StudioGAN is an open-source library under the MIT license (MIT) with the exception of StyleGAN2 and StyleGAN2 + ADA related implementations, which are under the NVIDIA source code license.} that supports various experimental setups from ACGAN~\cite{Odena2017ConditionalIS} to StyleGAN2~\cite{karras2020analyzing} + ADA~\cite{Karras2020TrainingGA} with different scales of datasets~\cite{Krizhevsky2009LearningML, Tiny, WelinderEtal2010, Deng2009ImageNetAL, choi2020starganv2} and architectures~\cite{Radford2016UnsupervisedRL, Gulrajani2017ImprovedTO, Brock2019LargeSG, karras2020analyzing}. For a fair comparison, we use the same backbones over all baselines, except otherwise noted, for both the discriminator and generator. For stable training, we apply spectral normalization (SN)~\cite{Miyato2018SpectralNF} to the generator and discriminator except for experiments using SNGAN (in this case, we apply SN to the discriminator only) and StyleGAN2~\cite{karras2020analyzing}. We also use the same conditioning method for generators with conditional batch normalization~(cBN)~\cite{Dumoulin2017ALR, de_Vries, Miyato2018cGANsWP}. This allows us to investigate solely the conditioning methods of the discriminator, which are the main interest of our paper. If not specified, we use hinge loss~\cite{Lim2017GeometricG} as a default for the adversarial loss.

Before conducting main experiments, we perform hyperparameter search with candidates of a temperature $\tau \in \{0.125, 0.25, 0.5, 0.75, 1.0\}$ and a positive margin $m_p \in~\{0.5, 0.75, 0.9, 0.95, 0.98, 1.0\}$. We set a negative margin $m_n$ as $1 - m_p$ to reduce search time. Through extensive experiments with 3 runs per each setting, we select $\tau$ with $\{0.5, 0.75, 0.25, 0.5, 0.25\}$ and $m_p$ with $\{0.98, 1.0, 0.95, 0.98, 0.90\}$ for $\text{CIFAR10}, \text{Tiny-ImageNet}, \text{CUB200}$, $\text{ImageNet 256 B.S.}$, and $\text{ImageNet 2048 B.S.}$ experiments, respectively.
A low temperature seems to work well on fine-grained image generation tasks, but generally, ReACGAN is robust to the choice of hyperparameters. The results of the hyperparameter search and other hyperparameter setups are provided in Appendix~\ref{hyperparameter_setup}. 

We evaluate all methods through the same protocol of~\cite{zhao2020differentiable, Zhang2019ConsistencyRF, Zhao2020ImprovedCR}, which uses the same amounts of generated images from the reference split specialized for each dataset.\footnote{We use the validation split as the default reference set, but we use the test split of~\text{CIFAR10} and the training split of~\text{CUB200} and~\text{AFHQ} due to the absence or lack of the validation dataset.} Besides, we run all the experiments three times with random seeds and report the averaged best performances for reliable evaluation with the lone exception of ImageNet and StyleGAN2 related experiments. Please refer to Appendix~\ref{hyperparameter_setup} for other experimental details. The numbers in bold-faced denote the best performance and in underline indicate that the values are in one standard deviation from the best.

\subsection{Evaluation Results}
\begin{table}[t!]
\caption{Comparison with classifier-based GANs~\cite{Odena2017ConditionalIS, kang2020contragan} and projection-based GANs~\cite{Miyato2018SpectralNF,Brock2019LargeSG,Zhang2019SelfAttentionGA} on CIFAR10~\cite{Krizhevsky2009LearningML}, Tiny-ImageNet~\cite{Tiny}, and CUB200~\cite{WelinderEtal2010} datasets using IS~\cite{Salimans2016ImprovedTF}, FID~\cite{Heusel2017GANsTB}, $\text{F}_{\text{0.125}}$, and $\text{F}_{\text{8}}$~\cite{sajjadi2018assessing} metrics. For baselines, both the numbers from the cited paper (denoted as * in method) and from our experiments using StudioGAN library~\cite{studiogan} are reported. The numbers in bold-faced denote the best performance and in underline indicate the values are in one standard deviation from the best.}
\setlength\tabcolsep{4.0pt}
\vspace{2mm}
  \resizebox{1.0\textwidth}{!}{
\begin{tabular}{lcccccccccccc}
\cmidrule[1.0pt]{1-13}
\multirow{2}*[-0.5ex]{\large{Method}} & \multicolumn{4}{c}{\text{CIFAR10~\cite{Krizhevsky2009LearningML}}} & \multicolumn{4}{c}{\text{Tiny-ImageNet~\cite{Tiny}}} & \multicolumn{4}{c}{\text{CUB200~\cite{WelinderEtal2010}}} \\
\cmidrule[1.0pt]( r){2-5}
\cmidrule[1.0pt](lr){6-9}
\cmidrule[1.0pt](l ){10-13}
& \text{IS}~$\uparrow$ & \text{FID}~$\downarrow$ & $\text{F}_{\text{0.125}}$~$\uparrow$ & $\text{F}_{\text{8}}$ ~$\uparrow$ & \text{IS}~$\uparrow$ & \text{FID}~$\downarrow$ & $\text{F}_{\text{0.125}}$~$\uparrow$ & $\text{F}_{\text{8}}$~$\uparrow$ & \text{IS}~$\uparrow$ & \text{FID}~$\downarrow$ & $\text{F}_{\text{0.125}}$~$\uparrow$ & $\text{F}_{\text{8}}$ ~$\uparrow$ \\ 
\cmidrule[1.0pt]( r){1-5}
\cmidrule[1.0pt](lr){6-9}
\cmidrule[1.0pt](l ){10-13}
SNGAN$^{*}$~\cite{Miyato2018SpectralNF} & 8.22 & 21.7 & - & - & - & - & - & - & - & - & - & -\\
BigGAN$^{*}$~\cite{Brock2019LargeSG} & 9.22 & 14.73 & - & - & - & - & - & - & - & - & - & -\\
ContraGAN$^{*}$~\cite{kang2020contragan} & - & 10.60 & - & - & - & 29.49 & - & - & - & - & - & -\\
ACGAN~\cite{Odena2017ConditionalIS} & 9.84 & 8.45 & \underline{0.993} & \textbf{0.992} & 6.00 & 96.04 & 0.656 & 0.368 & \textbf{6.09} & 60.73 & 0.726 & 0.891 \\
SNGAN~\cite{Miyato2018SpectralNF} & 8.67 & 13.33 & 0.985 & 0.976 & 8.71 & 51.15 & 0.900 & 0.702 & 5.41 & 47.75 & 0.754 & 0.912 \\
SAGAN~\cite{Zhang2019SelfAttentionGA} & 8.66 & 14.31 & 0.983 & 0.973 & 8.74 & 49.90 & 0.872 & 0.712 & 5.48 & 54.29 & 0.728 & 0.882 \\
BigGAN~\cite{Brock2019LargeSG} & \underline{9.81} & 8.08 & \underline{0.993} & \textbf{0.992} & 12.78 & 32.03 & 0.948 & 0.868 & 4.98 & 18.30 & 0.924 & \textbf{0.967}\\
ContraGAN~\cite{kang2020contragan} & 9.70 & 8.22 & \underline{0.993} & \underline{0.991} & 13.46 & 28.55 & \textbf{0.974} & 0.881 & 5.34 & 21.16 & 0.935 & 0.942 \\
\rowcolor{yellow!20}ReACGAN & \textbf{9.89} & \textbf{7.88} & \textbf{0.994} & \textbf{0.992} & \textbf{14.06} & \textbf{27.10} & 0.970 & \textbf{0.894} & 4.91 & \textbf{15.40} & \textbf{0.970} & 0.954\\
\midrule
BigGAN + CR$^{*}$~\cite{Zhang2019ConsistencyRF} & - & 11.48 & - & - & - & - & - & - & - & - & - & -\\
BigGAN~\cite{Brock2019LargeSG} + CR~\cite{Zhang2019ConsistencyRF} & \underline{9.97} & \textbf{7.18} & \underline{0.995} & \underline{0.993} & 15.94 & 19.96 & 0.972 & \textbf{0.950} & \textbf{5.14} & 11.97 & 0.978 & \textbf{0.981} \\
ContraGAN~\cite{kang2020contragan} + CR~\cite{Zhang2019ConsistencyRF} & 9.59 & 8.55 & 0.992 & 0.972 & 15.81 & \textbf{19.21} & \underline{0.983} & 0.941 & 4.90 & 11.08 &  \underline{0.984} & 0.967 \\
\rowcolor{yellow!20}ReACGAN + CR~\cite{Zhang2019ConsistencyRF} & \textbf{10.11} & \underline{7.20} & \textbf{0.996} & \textbf{0.994} & \textbf{16.56} & 19.69 & \textbf{0.984} & 0.940  & 4.87 & \textbf{10.72} & \textbf{0.985} & 0.971 \\
\midrule
BigGAN + DiffAug$^{*}$~\cite{zhao2020differentiable} & 9.17 & 8.49 & - & - & - & - & - & - & - & - & - & -\\
BigGAN~\cite{Brock2019LargeSG} + DiffAug~\cite{zhao2020differentiable} & 9.94 & 7.17 & 0.995 & \underline{0.992} & 18.08 & 15.70 & 0.980 & \textbf{0.972} & \textbf{5.53} & 12.15 & 0.967 & 0.981\\
ContraGAN~\cite{kang2020contragan} + DiffAug~\cite{zhao2020differentiable} & 9.95 & 7.27 & 0.995 & \underline{0.992} & 18.20 & 15.40 & 0.986 & 0.963 & 5.39 & 11.02 & 0.978 & 0.970 \\ 
\rowcolor{yellow!20}ReACGAN + DiffAug~\cite{zhao2020differentiable} & \textbf{10.22} & \textbf{6.79} & \textbf{0.996} & \textbf{0.993} & \textbf{20.60} & \textbf{14.25} & \textbf{0.988} & \textbf{0.972} & 5.22 & \textbf{9.27} & \textbf{0.985} & \textbf{0.983} \\
\cmidrule[1.0pt]{1-13}
Real Data & 11.54 &  &  &  & 34.11 &  &  &  & 5.49 &  &  &  \\ \cmidrule[1.0pt]{1-13}
\end{tabular}}
\label{table:main_table1}
\end{table}

\begin{table}[t!]
\caption{Experiments using ImageNet~\cite{Deng2009ImageNetAL} dataset. B.S. means batch size for training. (Left):~Comparisons with previous cGAN approaches. (Right): Learning curves of BigGAN~\cite{Brock2019LargeSG}, ContraGAN~\cite{kang2020contragan}, and ReACGAN~(ours) which are trained using the batch size of 256.}
\vspace{2mm}
\begin{minipage}[b]{0.50\linewidth}
    \centering
    \resizebox{1.0\textwidth}{!}{
    \begin{tabular}{clccccc}
    \cmidrule[1.0pt]{2-6}
    & \multirow{2}*[-0.5ex]{\large{Method}} & \multicolumn{4}{c}{\text{ImageNet~\cite{Deng2009ImageNetAL}}}\\
    \cmidrule[1.0pt]{3-6}
    & & \text{IS}~$\uparrow$ & \text{FID}~$\downarrow$ & $\text{F}_{\text{0.125}}$~$\uparrow$ & $\text{F}_{\text{8}}$ ~$\uparrow$\\
    \cmidrule[1.0pt]{2-6}
    \parbox[t]{2mm}{\multirow{14}{*}{\rotatebox[origin=c]{90}{\text{B.S.$=256$}}}}
    & ACGAN$^{*}$~\cite{NIPS2019_8414} & 7.26 & 184.41 & - & - &\\
    & SNGAN$^{*}$~\cite{Miyato2018SpectralNF} & 36.80 & 27.62 & - & - &\\
    & SAGAN$^{*}$~\cite{Zhang2019SelfAttentionGA} & 52.52 & 18.28 & - & - &\\
    & BigGAN$^{*}$~\cite{NIPS2019_8414} & 38.05  & 22.77 & - & - &\\
    & TAC-GAN$^{*}$~\cite{NIPS2019_8414} & 28.86 &  23.75 & - & - &\\
    & ContraGAN$^{*}$~\cite{kang2020contragan} & 31.10 & 19.69 & 0.951 & 0.927 &\\
    & ACGAN~\cite{Odena2017ConditionalIS} & 62.99 & 26.35 & 0.935 & 0.963 &\\
    & SNGAN~\cite{Miyato2018SpectralNF} & 32.25 & 26.79 & 0.938 & 0.913 &\\
    & SAGAN~\cite{Zhang2019SelfAttentionGA} & 29.85 & 34.73 & 0.849 & 0.914 &\\
    & BigGAN~\cite{Brock2019LargeSG} & 43.97 & 16.36 & 0.964 & 0.955 &\\
    & ContraGAN~\cite{kang2020contragan} & 25.25 & 25.16 & 0.947 & 0.855  &\\
    &  \cellcolor{yellow!20}ReACGAN & \cellcolor{yellow!20}\textbf{68.27} & \cellcolor{yellow!20}\textbf{13.98} & \cellcolor{yellow!20}\textbf{0.976} & \cellcolor{yellow!20}\textbf{0.977} &\\
    \cmidrule[0.4pt]{2-6}
    & BigGAN~\cite{Brock2019LargeSG} + DiffAug~\cite{zhao2020differentiable} & 36.97 & 18.57 & 0.956 & 0.941 &\\
    & \cellcolor{yellow!20}ReACGAN + DiffAug~\cite{zhao2020differentiable} & \cellcolor{yellow!20}\textbf{69.74} & \cellcolor{yellow!20}\textbf{11.95} & \cellcolor{yellow!20}\textbf{0.977} & \cellcolor{yellow!20}\textbf{0.975} &\\
    \cmidrule[1.0pt]{2-6}
    \parbox[t]{2mm}{\multirow{4}{*}{\rotatebox[origin=c]{90}{\text{B.S.$=2048$}}}}
    & BigGAN$^{*}$~\cite{Brock2019LargeSG} & 99.31 & 8.51 & - & - &\\
    & BigGAN$^{*}$~\cite{zhou2020omni} & \textbf{104.57} & 9.18 & - & - &\\
    & BigGAN~\cite{Brock2019LargeSG} & 99.71  & \textbf{7.89} & 0.985 & 0.989 &\\
    & \cellcolor{yellow!20}ReACGAN & \cellcolor{yellow!20}92.74 & \cellcolor{yellow!20}8.23 & \cellcolor{yellow!20}\textbf{0.991} & \cellcolor{yellow!20}\textbf{0.990} &\\
    \cmidrule[1.0pt]{2-6}
    & Real Data & 173.33 &  &  & & \\ \cmidrule[1.0pt]{2-6}
    \end{tabular}}
    \label{table:student}
\end{minipage} 
\begin{minipage}[t]{0.49\linewidth}
    \centering
    \includegraphics[width=1.0\textwidth]{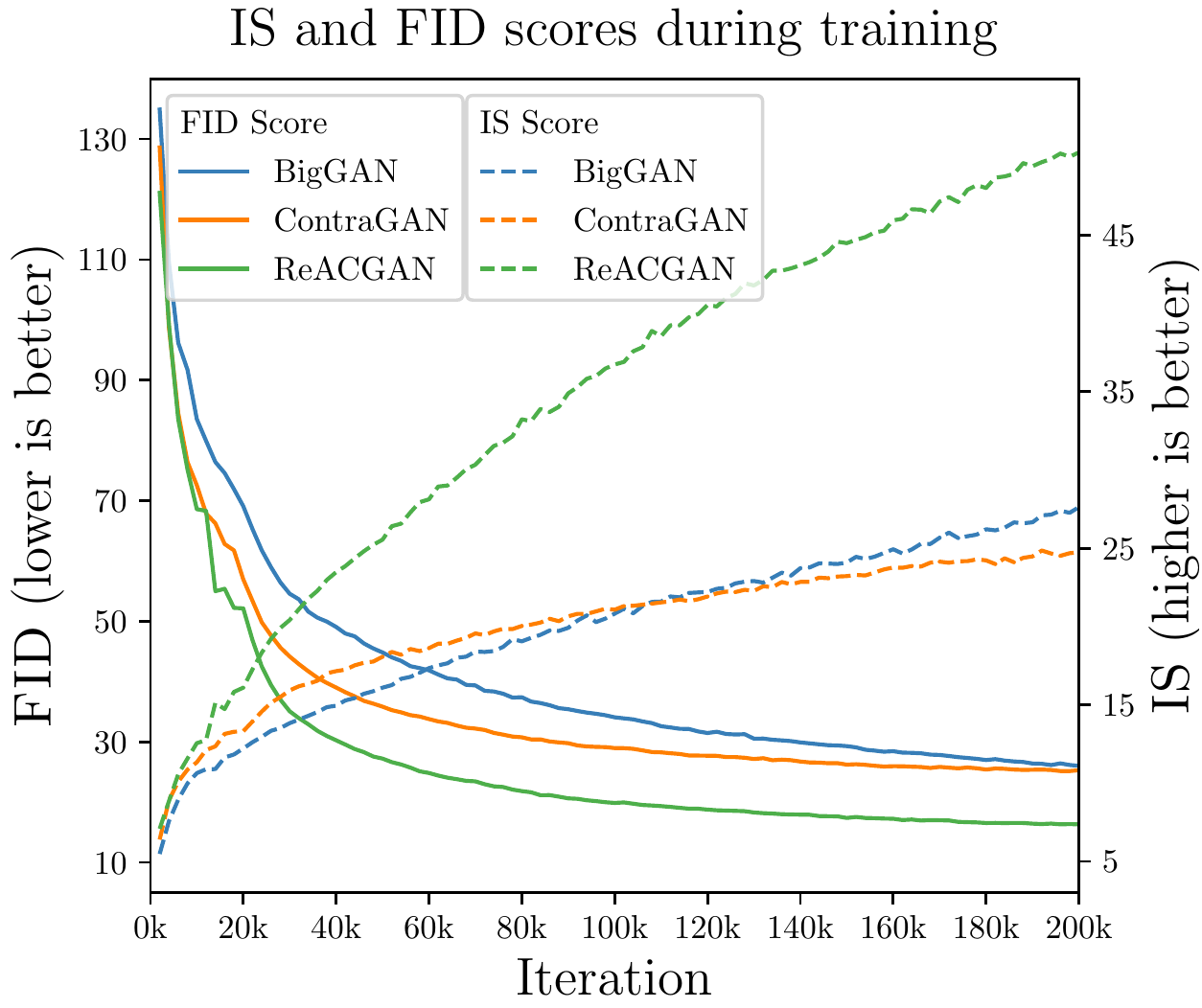}
    \label{fig:image}
\end{minipage}
\\
\label{table:main_table2}
\vspace{-7mm}
\end{table}
\textbf{Comparison with Other cGANs.}  We compare ReACGAN with previous state-of-the-art cGANs in Tables~\ref{table:main_table1} and~\ref{table:main_table2}. 
We employ the implementations of GANs in PyTorch-StudioGAN library as it provides improved results on standard benchmark datasets~\cite{Krizhevsky2009LearningML, Deng2009ImageNetAL}. For a fair comparison, we provide results from each original paper (denoted as * in method) as well as those from StudioGAN library~\cite{studiogan}. We also conduct experiments with popular augmentation-based methods: consistency regularization~(CR)~\cite{Zhang2019ConsistencyRF} and differentiable augmentation~(DiffAug)~\cite{zhao2020differentiable}.

Compared with other cGANs, ReACGAN performs the best on most benchmarks, surpassing the previous methods by 2.5\%, 5.1\%, 15.8\%, and 14.5\% in FID on CIFAR10, Tiny-ImageNet, CUB200, and ImageNet~(256~B.S.), respectively. ReACGAN also harmonizes with augmentation-based regularizations, bringing incremental improvements on all the metrics. For the ImageNet experiments using a batch size of 256, ReACGAN reaches higher IS and lower FID with fewer iterations than other models in comparison. Finally, we demonstrate that ReACGAN can learn with a larger batch size on ImageNet. While some recent methods~\cite{NIPS2019_8414, zhou2018activation, hou2021cgans} have been built on ACGAN to improve the generation performance of ACGAN, large-scale image generation experiments with the batch size of 2048 have never been reported. Table~\ref{table:main_table2} shows that our ReACGAN reaches FID score of 8.23 on ImageNet, being comparable with the value of 7.89 from BigGAN implementation in PyTorch-StudioGAN library. ReACGAN, however, provides better synthesis results than other implementations of BigGAN~\cite{Brock2019LargeSG, zhou2020omni}. Note that our result on ImageNet is obtained in only two runs while the training setup and architecture of BigGAN have been extensively searched and finely tuned.

\textbf{Comparison with Other Conditioning Losses.} 
We investigate how the generation qualities vary with different conditioning losses while keeping the other configurations fixed.
We compare D2D-CE loss with cross-entropy loss of ACGAN~(AC)~\cite{Odena2017ConditionalIS}, loss used in the projection discriminator~(PD)~\cite{Miyato2018cGANsWP}, multi-hinge loss~(MH)~\cite{kavalerov2021multi}, and conditional contrastive loss~(2C)~\cite{kang2020contragan}. The results are shown in Table~\ref{table:change_condition}.
AC- and MH-based models present decent results on \text{CIFAR10}, but undergo early-training collapse on \text{Tiny-ImageNet} and \text{CUB200} datasets.
Replacing them with PD, 2C, and D2D-CE losses produce satisfactory performances across all datasets, where PD loss makes the best $\text{F}_8$~(recall) on CUB200 dataset while giving third $\text{F}_{0.125}$ (precision) value.
The noticeable point is that the proposed D2D-CE shows consistent results across all datasets, showing the lowest FID and the highest $\text{F}_8$ and $\text{F}_{0.125}$ values in most cases. This means ReACGAN can generate high-fidelity images and is relatively free from the precision and recall trade-off~\cite{sajjadi2018assessing} than the others.

\begin{table}[t!]
\centering
\caption{Experiments on the effectiveness of D2D-CE loss compared with other conditioning losses.}
\setlength\tabcolsep{4.0pt}
\vspace{2mm}
  \resizebox{0.95\textwidth}{!}{
\begin{tabular}{lcccccccccccc}
\cmidrule[1.0pt]{1-13}
\multirow{2}*[-0.5ex]{\normalsize{Conditioning Method}} & \multicolumn{4}{c}{\text{CIFAR10~\cite{Krizhevsky2009LearningML}}} & \multicolumn{4}{c}{\text{Tiny-ImageNet~\cite{Tiny}}} & \multicolumn{4}{c}{\text{CUB200~\cite{WelinderEtal2010}}} \\
\cmidrule[1.0pt]( r){2-5}
\cmidrule[1.0pt](lr){6-9}
\cmidrule[1.0pt](l ){10-13}
& \textsc{IS}~$\uparrow$ & \textsc{FID}~$\downarrow$ & $\textsc{F}_{\text{0.125}}$~$\uparrow$ & $\textsc{F}_{\text{8}}$ ~$\uparrow$ &\textsc{IS}~$\uparrow$ & \textsc{FID}~$\downarrow$ & $\textsc{F}_{\text{0.125}}$~$\uparrow$ & $\textsc{F}_{\text{8}}$~$\uparrow$ & \textsc{IS}~$\uparrow$ & \textsc{FID}~$\downarrow$ & $\textsc{F}_{\text{0.125}}$~$\uparrow$ & $\textsc{F}_{\text{8}}$ ~$\uparrow$ \\
\cmidrule[1.0pt]( r){1-5}
\cmidrule[1.0pt](lr){6-9}
\cmidrule[1.0pt](l ){10-13}
\multirow{2}*[1.0ex]{BigGAN w/o Condition~\cite{Brock2019LargeSG}} & \multirow{2}*[-0.5ex]{9.46} & \multirow{2}*[-0.5ex]{12.21} & \multirow{2}*[-0.5ex]{0.987} & \multirow{2}*[-0.5ex]{0.982}& \multirow{2}*[-0.5ex]{7.38} & \multirow{2}*[-0.5ex]{76.15} & \multirow{2}*[-0.5ex]{0.804} & \multirow{2}*[-0.5ex]{0.576}& \multirow{2}*[-0.5ex]{5.16} & \multirow{2}*[-0.5ex]{35.17} & \multirow{2}*[-0.5ex]{0.852} & \multirow{2}*[-0.5ex]{0.936}\\
\multirow{2}*[1.0ex]{(Abbreviated to Big)} &  &  &  &  & & &  &  &  &  &  & \\
\midrule
Big + AC~\cite{Odena2017ConditionalIS} & 9.84 & 8.45 & \underline{0.993} & \textbf{0.992} & 6.00 & 96.04 & 0.656 & 0.368 & \textbf{6.09} & 60.73 & 0.726 & 0.891 \\
Big + PD~\cite{Miyato2018cGANsWP} & 9.81 & 8.08 & \underline{0.993} & \textbf{0.992} & 12.78 & 32.03 & 0.948 & 0.868 & 4.98 & 18.30 & 0.924 & \textbf{0.967}\\
Big + MH~\cite{kavalerov2021multi} & \textbf{10.05} & 7.94 & \textbf{0.994} & 0.990 & 4.37 & 140.74 & 0.282 & 0.156 & 5.18 & 245.69 & 0.625 & 0.832\\
Big + 2C~\cite{kang2020contragan} & 9.70 & 8.22 & \underline{0.993} & \underline{0.991} & 13.46 & 28.55 & \textbf{0.974} & 0.881 & 5.34 & 21.16 & 0.935 & 0.942\\
\rowcolor{yellow!20}Big + D2D-CE~(ReACGAN) & 9.89 & \textbf{7.88} & \textbf{0.994} & \textbf{0.992} & \textbf{14.06} & \textbf{27.10} & 0.970 & \textbf{0.894} & 4.91 & \textbf{15.40} & \textbf{0.970} & 0.954 \\ \cmidrule[1.0pt]{1-13}
\vspace{-0.4cm}
\end{tabular}}
\label{table:change_condition}
\end{table}

\begin{table}[ht]
\vspace{-1.5mm}
\centering
\caption{Experiments to identify the consistent performance of D2D-CE on adversarial loss selection.}
\vspace{2mm}
  \resizebox{0.95\textwidth}{!}{
\begin{tabular}{lccccccccccc}
\cmidrule[1.0pt]{1-12}
\multirow{2}*[-0.5ex]{\normalsize{Adversarial Loss}} & \multirow{2}*[0.5ex]{Conditioning} & \multicolumn{5}{c}{\text{CIFAR10~\cite{Krizhevsky2009LearningML}}} & \multicolumn{5}{c}{\text{Tiny-ImageNet~\cite{Tiny}}} \\
\cmidrule[1.0pt]( r){3-7}
\cmidrule[1.0pt]( l){8-12}
& \multirow{2}*[1.5ex]{Method} & \text{IS}~$\uparrow$ & \text{FID}~$\downarrow$ & $\text{F}_{\text{0.125}}$~$\uparrow$ & $\text{F}_{\text{8}}$ ~$\uparrow$ & Better? & \text{IS}~$\uparrow$ & \text{FID}~$\downarrow$ & $\text{F}_{\text{0.125}}$~$\uparrow$ & $\text{F}_{\text{8}}$ ~$\uparrow$ & Better? \\
\cmidrule[1.0pt]{1-12}
\multirow{3}*[0.5ex]{Non-saturation~\cite{Goodfellow2014GenerativeAN}} & PD~\cite{Miyato2018cGANsWP} & \underline{9.75} & \underline{8.29} & \textbf{0.993} & \textbf{0.991} & \checkmark & 8.27 & 58.85 & 0.816 & 0.713 &  \\
& 2C~\cite{kang2020contragan} & 9.30 & 10.47 & 0.990 & 0.959 &  & 6.57 & 84.27 & 0.745 & 0.556 &  \\
 & \cellcolor{yellow!20}D2D-CE & \cellcolor{yellow!20}\textbf{9.79} & \cellcolor{yellow!20}\textbf{8.27} & \cellcolor{yellow!20}\textbf{0.993} & \cellcolor{yellow!20}\textbf{0.991} & \cellcolor{yellow!20}\checkmark & \cellcolor{yellow!20}\textbf{11.76} & \cellcolor{yellow!20}\textbf{39.32} & \cellcolor{yellow!20}\textbf{0.942} & \cellcolor{yellow!20}\textbf{0.852} & \cellcolor{yellow!20}\checkmark \\
\cmidrule[0.4pt]{1-12}
\multirow{3}*[0.5ex]{Least square~\cite{Mao2017LeastSG}} & PD~\cite{Miyato2018cGANsWP} & \textbf{9.94} & \textbf{8.26} & \textbf{0.993} & \textbf{0.992} & \checkmark & \textbf{12.74} & \textbf{37.14} & \textbf{0.920} & \textbf{0.900} &  \checkmark\\
 & 2C~\cite{kang2020contragan} & 8.66 & 12.18 & 0.986 & 0.941 & & 9.58 & 53.10 & \underline{0.916} & 0.706 &\\
& \cellcolor{yellow!20}D2D-CE & \cellcolor{yellow!20}9.70 & \cellcolor{yellow!20}9.56 & \cellcolor{yellow!20}0.991 & \cellcolor{yellow!20}0.987 & \cellcolor{yellow!20}& \cellcolor{yellow!20}9.50 & \cellcolor{yellow!20}57.67 & \cellcolor{yellow!20}0.848 & \cellcolor{yellow!20}0.692 & \cellcolor{yellow!20}\\
\cmidrule[0.4pt]{1-12}
\multirow{3}*[0.5ex]{W-GP~\cite{Gulrajani2017ImprovedTO}} & PD~\cite{Miyato2018cGANsWP} & 5.71 & 64.75 & 0.792 & 0.652 & & 6.67 & 84.16 & 0.696 & 0.498 &\\
& 2C~\cite{kang2020contragan} & 5.93 & 55.99 & 0.842 & 0.709 &  & 6.89 & 74.45 & 0.812 & 0.536 &\\
 & \cellcolor{yellow!20}D2D-CE & \cellcolor{yellow!20}\textbf{7.30} & \cellcolor{yellow!20}\textbf{35.94} & \cellcolor{yellow!20}\textbf{0.942} & \cellcolor{yellow!20}\textbf{0.847} & \cellcolor{yellow!20}\checkmark & \cellcolor{yellow!20}\textbf{8.92} & \cellcolor{yellow!20}\textbf{52.74} & \cellcolor{yellow!20}\textbf{0.856} & \cellcolor{yellow!20}\textbf{0.689} & \cellcolor{yellow!20}\checkmark\\
\cmidrule[0.4pt]{1-12}
\multirow{3}*[0.5ex]{Hinge~\cite{Lim2017GeometricG}} & PD~\cite{Miyato2018cGANsWP} & \underline{9.81} & 8.08 & 0.993 & \textbf{0.992} & & 12.78 & 32.03 & 0.948 & 0.868 &\\
& 2C~\cite{kang2020contragan} & 9.70 & 8.22 & 0.993 & \underline{0.991} &  & 13.46 & 28.55 & \textbf{0.974} & 0.881 &\\
& \cellcolor{yellow!20}D2D-CE & \cellcolor{yellow!20}\textbf{9.89} & \cellcolor{yellow!20}\textbf{7.88} & \cellcolor{yellow!20}\textbf{0.994} & \cellcolor{yellow!20}\textbf{0.992} & \cellcolor{yellow!20}\checkmark & \cellcolor{yellow!20}\textbf{14.06} & \cellcolor{yellow!20}\textbf{27.10} & \cellcolor{yellow!20}\underline{0.970} & \cellcolor{yellow!20}\textbf{0.894} & \cellcolor{yellow!20}\checkmark\\
\cmidrule[1.0pt]{1-12}
\vspace{-0.7cm}
\end{tabular}}
\label{table:adversarial}
\end{table}
\textbf{Consistent Performance of ReACGAN on Adversarial Loss Selection.} We validate the consistent performance of ReACGAN on four adversarial losses: non-saturation loss~\cite{Goodfellow2014GenerativeAN}, least square loss~\cite{Mao2017LeastSG}, Wasserstein loss with gradient penalty regularization~(W-GP)~\cite{Gulrajani2017ImprovedTO}, and hinge loss~\cite{Lim2017GeometricG} on CIFAR10 and Tiny-ImageNet datasets in Table~\ref{table:adversarial}. The experimental results show that BigGAN + D2D-CE (ReACGAN) consistently outperforms the projection discriminator~(PD) and conditional contrastive loss~(2C) counterparts over three adversarial losses~\cite{Goodfellow2014GenerativeAN, Gulrajani2017ImprovedTO, Lim2017GeometricG}. However, for the experiments using the least square loss~\cite{Mao2017LeastSG}, ReACGAN exhibits inferior generation performances to the projection discriminator. We speculate that minimizing the least square distance between an adversarial logit and the target scalar (1 or 0) might affect the norms of feature maps and spoil the classifier training performed by D2D-CE loss.

\begin{table}[t!]
\centering
\caption{FID~\cite{Heusel2017GANsTB} values of conditional StyleGAN2~(cStyleGAN2)~\cite{karras2020analyzing} and StyleGAN2~\cite{karras2020analyzing} + D2D-CE on CIFAR10~\cite{Krizhevsky2009LearningML} and AFHQ~\cite{choi2020starganv2} datasets. ADA means the adaptive discriminator augmentation~\cite{Karras2020TrainingGA}. FID is exceptionally computed using the training split for calculating the reference moments since FID value of StyleGAN2 is often calculated using the moments of the training dataset.}
\vspace{2mm}
\resizebox{0.65\textwidth}{!}{
\begin{tabular}{lcc}
\cmidrule[1.0pt]{1-3}
Conditioning method & CIFAR10~\cite{Krizhevsky2009LearningML} & AFHQ~\cite{choi2020starganv2} \\
\cmidrule[1.0pt]{1-3}
\text{cStyleGAN2~\cite{karras2020analyzing}} & 3.88 & - \\
\rowcolor{yellow!20}\text{StyleGAN2~\cite{karras2020analyzing} + D2D-CE} & \textbf{3.34} & - \\
\midrule
\text{cStyleGAN2~\cite{karras2020analyzing} + ADA~\cite{Karras2020TrainingGA}} & 2.43 & \underline{4.99} \\
\rowcolor{yellow!20}\text{StyleGAN2~\cite{karras2020analyzing} + ADA~\cite{Karras2020TrainingGA} + D2D-CE} & \textbf{2.38} & \textbf{4.95} \\
\midrule
\rowcolor{yellow!20}\text{StyleGAN2~\cite{karras2020analyzing} + DiffAug~\cite{zhao2020differentiable} + D2D-CE + Tuning} & \textbf{2.26} & - \\
\cmidrule[1.0pt]{1-3}
\vspace{-0.7cm}
\end{tabular}}
\label{table:stylegan}
\end{table}
\textbf{Effect of D2D-CE for Different GAN Architectures.} We study the effect of D2D-CE with different GAN architectures. In Table~\ref{table:stylegan}, we validate that D2D-CE is effective for StyleGAN2~\cite{karras2020analyzing} backbone and also fits well with the adaptive discriminator augmentation~(ADA)~\cite{Karras2020TrainingGA}. StyleGAN2 with D2D-CE loss produces 13.9$\%$ better generation result than the conditional version of StyleGAN2~(cStyleGAN2) on CIFAR10. Moreover, StyleGAN2 with D2D-CE can be reinforced with ADA or DiffAug when train StyleGAN2 + D2D-CE under the limited data situation. Among GANs, StyleGAN2 + DiffAug + D2D-CE + Tuning achieves the best performance on CIFAR10, even outperforming some diffusion-based methods~\cite{nichol2021improved, kim2021score}.
Additional results with other architectures, i.e., a deep convolutional network~\cite{Radford2016UnsupervisedRL} and a resnet style backbone~\cite{Gulrajani2017ImprovedTO}, are provided in Table~\ref{table:sup_architecture} in Appendix~\ref{additional_experiments}.  
\\

\subsection{Ablation Study}
\label{sec:ablation}
We study how each component of ReACGAN affects ACGAN training. By adding or ablating each part of ReACGAN, as shown in Table~\ref{table:ReAC_ablation}, we identify four major observations. (1) Feature and weight normalization greatly stabilize ACGAN training and improve generation performances on Tiny-ImageNet and CUB200 datasets. (2) D2D-CE enhances the generation performance by considering data-to-data relationships and by performing easy sample suppression~(3th and 4th rows). (3) the suppression technique does not work well on the feature normalized cross-entropy loss~(5th row). (4) While ACGAN shows high Inception score on ImageNet experiment, 
it shows relatively poor FID, $\text{F}_{0.125}$, and $\text{F}_{8}$ values compared with the model trained with normalization, data-to-data consideration, and the suppression technique. This result is consistent with the qualitative results on ImageNet, where the images from ACGAN are easily classifiable, but the images from ReACGAN are high quality and diverse. We attribute the improvement to the discriminator that successfully leverages informative data-to-data and data-to-class relationships with easy sample suppression.
\begin{figure}[t!]    
    
    \centering
    \includegraphics[width=0.96\linewidth]{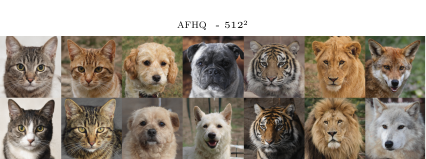}
    \centering
    \includegraphics[width=0.31\linewidth]{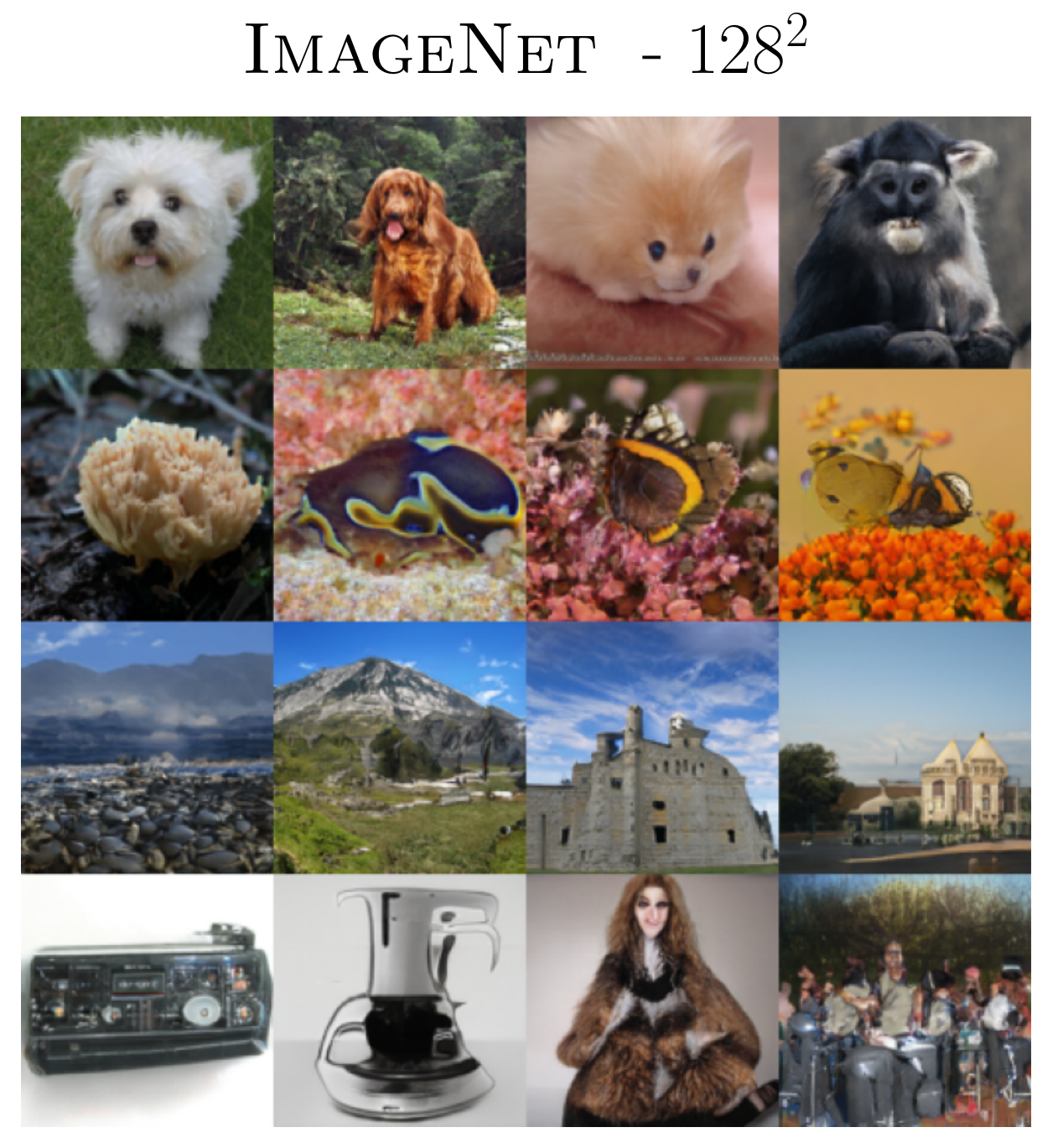}
    \hspace{-0.20cm}
    \includegraphics[width=0.31\linewidth]{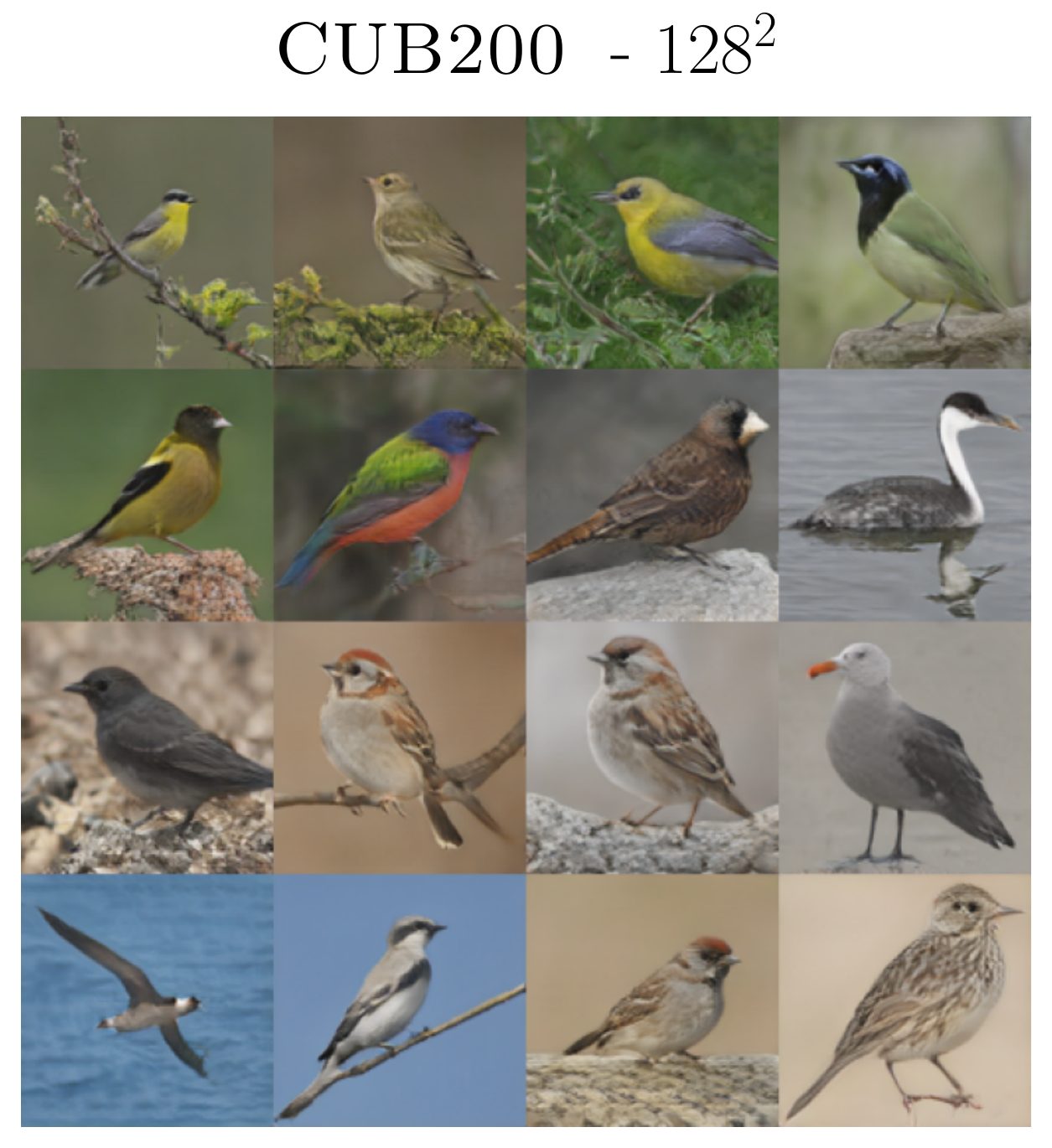}
    \hspace{-0.30cm}
    \includegraphics[width=0.205\linewidth]{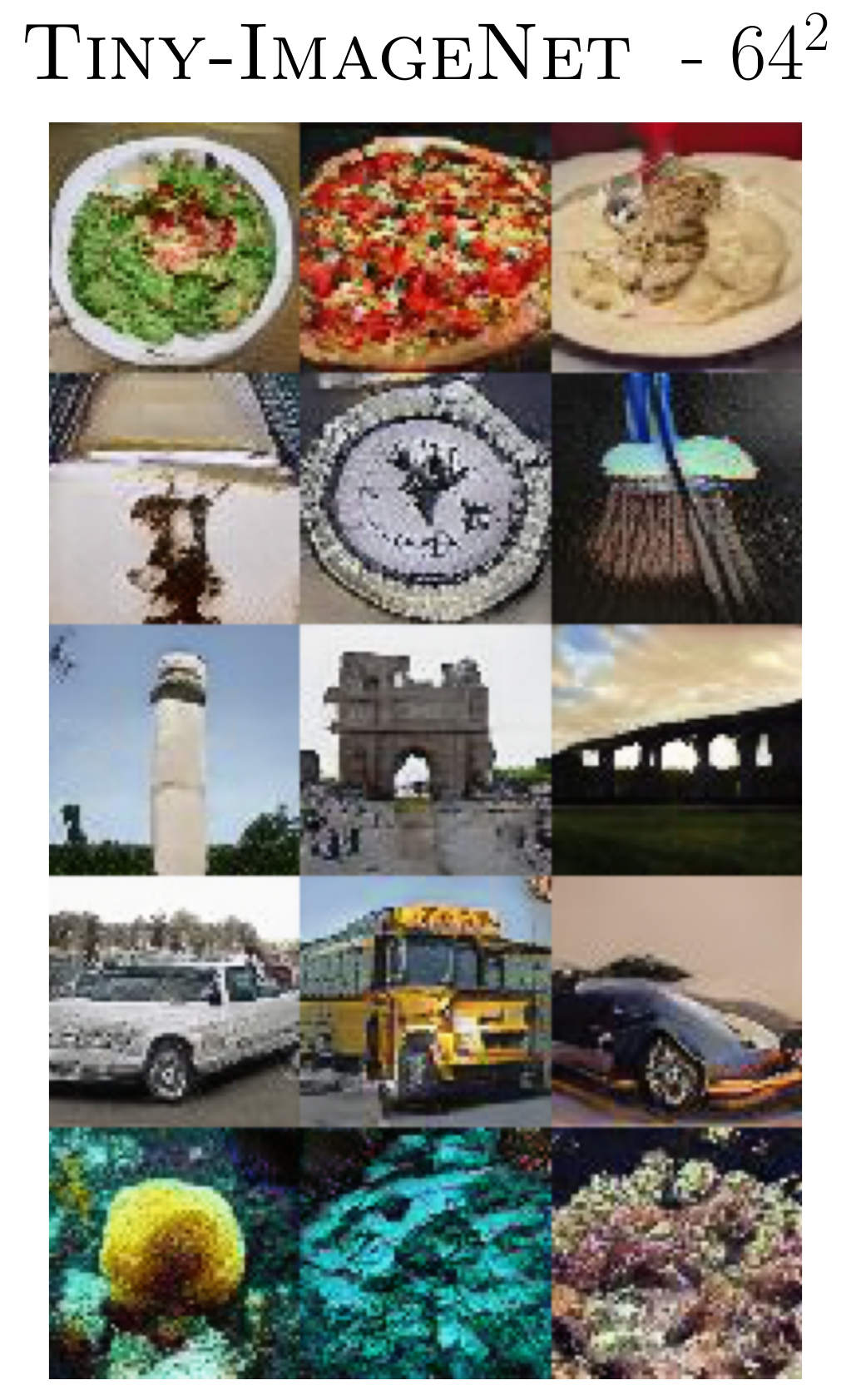}
    \hspace{-0.25cm}
    \includegraphics[width=0.16\linewidth]{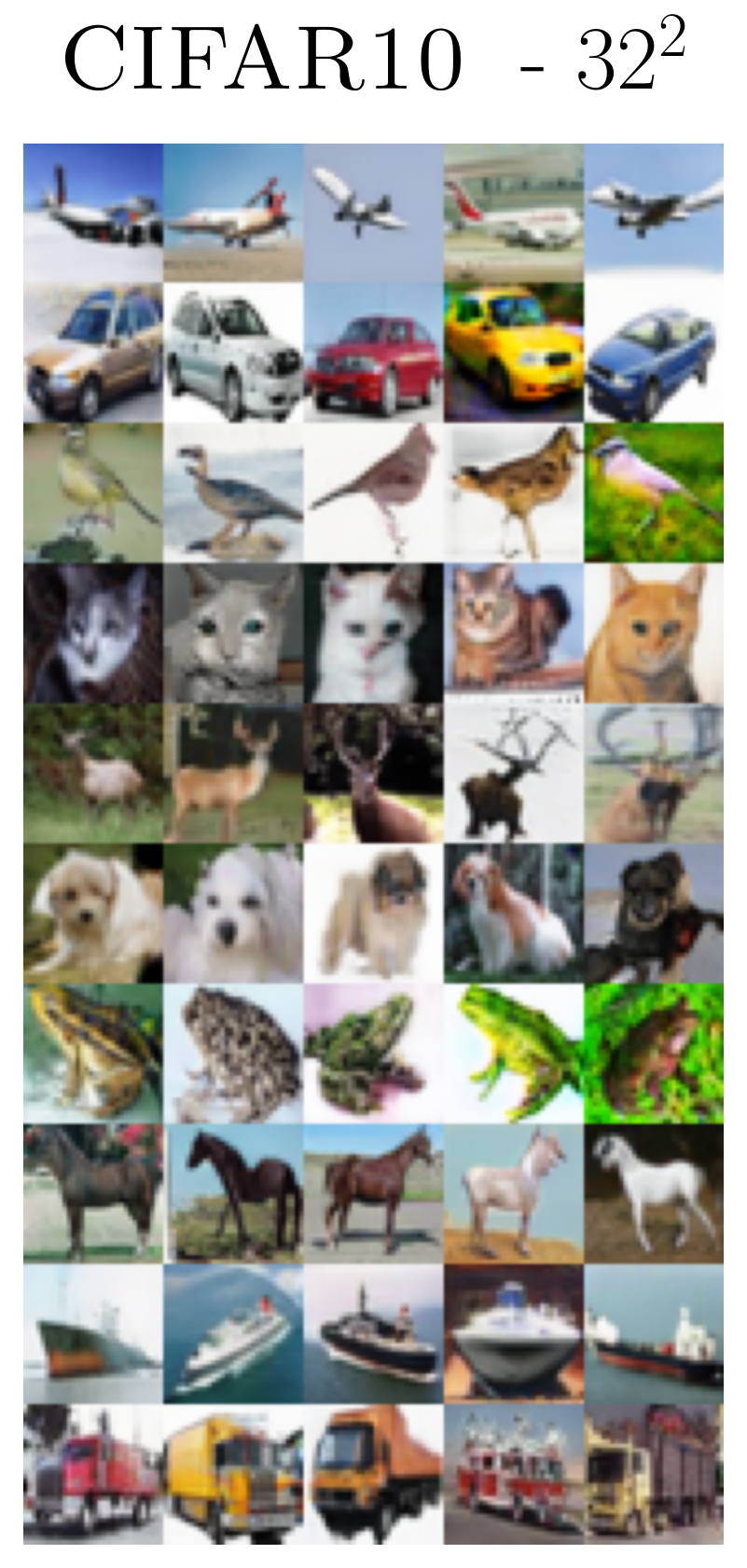}
    \caption{Curated images generated by the proposed ReACGAN. More qualitative results on 
    ReACGAN, BigGAN~\cite{Brock2019LargeSG}, ContraGAN~\cite{kang2020contragan}, and ACGAN~\cite{Odena2017ConditionalIS} are attached in Appendix~\ref{qualitative_all}.}
    \label{fig:Figure_qualitative}
\end{figure}
\begin{table}[t!]
\centering
\caption{Ablation study on feature map normalization, data-to-data consideration, and easy positive and negative sample suppression.}
\setlength\tabcolsep{4.0pt}
\vspace{2mm}
  \resizebox{0.96\textwidth}{!}{
\begin{tabular}{lcccccccccccc}
\cmidrule[1.0pt]{1-13}
\multirow{2}*[-0.5ex]{\large{Ablation}} & \multicolumn{4}{c}{\text{Tiny-ImageNet~\cite{Tiny}}} & \multicolumn{4}{c}{\text{CUB200~\cite{WelinderEtal2010}}} & \multicolumn{4}{c}{\text{ImageNet~\cite{deng2019arcface}}} \\
\cmidrule[1.0pt]( r){2-5}
\cmidrule[1.0pt](lr){6-9}
\cmidrule[1.0pt](l ){10-13}
& \text{IS}~$\uparrow$ & \text{FID}~$\downarrow$ & $\text{F}_{\text{0.125}}$~$\uparrow$ & $\text{F}_{\text{8}}$ ~$\uparrow$ &\text{IS}~$\uparrow$ & \text{FID}~$\downarrow$ & $\text{F}_{\text{0.125}}$~$\uparrow$ & $\text{F}_{\text{8}}$~$\uparrow$ & \text{IS}~$\uparrow$ & \text{FID}~$\downarrow$ & $\text{F}_{\text{0.125}}$~$\uparrow$ & $\text{F}_{\text{8}}$ ~$\uparrow$ \\
\cmidrule[1.0pt]( r){1-5}
\cmidrule[1.0pt](lr){6-9}
\cmidrule[1.0pt](l ){10-13}
ACGAN~\cite{Odena2017ConditionalIS} & 6.00 & 96.04 & 0.656 & 0.368 & \textbf{6.09} & 60.73 & 0.726 & 0.891 & 62.99 & 26.35 & 0.935 &  0.963\\
+ Normalization  & 13.46 & 30.33 & 0.955 & 0.889 & 4.78 & 25.54 & 0.883 & \underline{0.952} & 18.16 & 36.40 & 0.879 & 0.787 \\
+ Data-to-data~(Eq.~(\ref{eq:eq5}))  & 12.96 & 28.71 & \underline{0.967} & 0.863 & 5.08 & 25.12 & 0.894 & 0.946 & - & - & - & - \\
\rowcolor{yellow!20}+ Suppression~(Eq.~(\ref{eq:eq6}))  & \textbf{14.06} & \textbf{27.10} & \textbf{0.970} & \textbf{0.894} & 4.91 & \textbf{15.40} & \textbf{0.970} & \textbf{0.954} & \textbf{63.16} & \textbf{14.59} & \textbf{0.974} & \textbf{0.974} \\ \cmidrule[0.4pt]{1-13}
- Data-to-data & 12.96 & 30.79 & 0.960 & 0.857 & 5.39 & 30.36 & 0.863 & 0.947 & - & - & - & - \\
\cmidrule[1.0pt]{1-13}
\end{tabular}}
\label{table:ReAC_ablation}
\vspace{-0.3cm}
\end{table}
\section{Conclusion}
\label{sec:Conclusion}
In this paper, we have analyzed why training ACGAN becomes unstable as the number of classes in the dataset increases. By deriving the analytic form of gradient in the classifier and numerically checking the gradient values, we have discovered that the unstable training comes from a gradient exploding problem caused by the unboundedness of input feature vectors and poor classification ability of the classifier in the early training stage. To alleviate the instability and reinforce ACGAN, we have proposed the Data-to-Data Cross-Entropy loss~(D2D-CE) and the Rebooted Auxiliary Classifier Generative Adversarial Network~(ReACGAN). The experimental results verify the superiority of ReACGAN compared with the existing classifier- and projection-based GANs on five benchmark datasets. Moreover, exhaustive analyses on ReACGAN prove that ReACGAN is robust to hyperparameter selection and harmonizes with various architectures and differentiable augmentations.
\newpage
\begin{ack}
This work was supported by the IITP
grants (No.2019-0-01906: AI Graduate School Program - POSTECH, No.2021-0-00537: Visual Common Sense, No.2021-0-02068: AI Innovation Hub) funded by Ministry of Science and ICT, Korea.

\end{ack}
{\small
    \bibliographystyle{unsrt}
    \bibliography{paper}

\begin{thebibliography}{10}

\bibitem{Goodfellow2014GenerativeAN}
Ian Goodfellow, Jean Pouget-Abadie, Mehdi Mirza, Bing Xu, David Warde-Farley,
  Sherjil Ozair, Aaron Courville, and Yoshua Bengio.
\newblock {Generative Adversarial Nets}.
\newblock In {\em Advances in Neural Information Processing Systems (NeurIPS)},
  pages 2672--2680, 2014.

\bibitem{Radford2016UnsupervisedRL}
Alec Radford, Luke Metz, and Soumith Chintala.
\newblock {Unsupervised Representation Learning with Deep Convolutional
  Generative Adversarial Networks}.
\newblock {\em arXiv preprint arXiv 1511.06434}, 2016.

\bibitem{Miyato2018SpectralNF}
Takeru Miyato, Toshiki Kataoka, Masanori Koyama, and Yuichi Yoshida.
\newblock {Spectral Normalization for Generative Adversarial Networks}.
\newblock In {\em Proceedings of the International Conference on Learning
  Representations (ICLR)}, 2018.

\bibitem{Brock2019LargeSG}
Andrew Brock, Jeff Donahue, and Karen Simonyan.
\newblock {Large Scale {GAN} Training for High Fidelity Natural Image
  Synthesis}.
\newblock In {\em Proceedings of the International Conference on Learning
  Representations (ICLR)}, 2019.

\bibitem{karras2019style}
Tero Karras, Samuli Laine, and Timo Aila.
\newblock {A style-based generator architecture for generative adversarial
  networks}.
\newblock In {\em Proceedings of the IEEE International Conference on Computer
  Vision and Pattern Recognition (CVPR)}, pages 4401--4410, 2019.

\bibitem{Wu2019LOGANLO}
Yan Wu, Jeff Donahue, David Balduzzi, Karen Simonyan, and Timothy~P. Lillicrap.
\newblock {LOGAN: Latent Optimisation for Generative Adversarial Networks}.
\newblock {\em arXiv preprint arXiv 1912.00953}, 2019.

\bibitem{karras2020analyzing}
Tero Karras, Samuli Laine, Miika Aittala, Janne Hellsten, Jaakko Lehtinen, and
  Timo Aila.
\newblock {Analyzing and improving the image quality of stylegan}.
\newblock In {\em Proceedings of the IEEE International Conference on Computer
  Vision and Pattern Recognition (CVPR)}, pages 8110--8119, 2020.

\bibitem{Karras2020TrainingGA}
Tero Karras, Miika Aittala, Janne Hellsten, Samuli Laine, Jaakko Lehtinen, and
  Timo Aila.
\newblock {Training Generative Adversarial Networks with Limited Data}.
\newblock In {\em Advances in Neural Information Processing Systems (NeurIPS)},
  2020.

\bibitem{zhao2020differentiable}
Shengyu Zhao, Zhijian Liu, Ji~Lin, Jun-Yan Zhu, and Song Han.
\newblock {Differentiable augmentation for data-efficient gan training}.
\newblock {\em arXiv preprint arXiv 2006.10738}, 2020.

\bibitem{Mao2017LeastSG}
Xudong Mao, Qing Li, Haoran Xie, Raymond Y.~K. Lau, Zhixiang Wang, and
  Stephen~Paul Smolley.
\newblock {Least Squares Generative Adversarial Networks}.
\newblock In {\em Proceedings of the International Conference on Computer
  Vision (ICCV)}, pages 2813--2821, 2017.

\bibitem{Arjovsky2017WassersteinG}
Mart{\'i}n Arjovsky, Soumith Chintala, and L{\'e}on Bottou.
\newblock {Wasserstein GAN}.
\newblock {\em arXiv preprint arXiv 1701.07875}, 2017.

\bibitem{Arjovsky2017TowardsPM}
Mart{\'i}n Arjovsky and L{\'e}on Bottou.
\newblock {Towards Principled Methods for Training Generative Adversarial
  Networks}.
\newblock In {\em Proceedings of the International Conference on Learning
  Representations (ICLR)}, 2017.

\bibitem{Gulrajani2017ImprovedTO}
Ishaan Gulrajani, Faruk Ahmed, Martin Arjovsky, Vincent Dumoulin, and Aaron~C
  Courville.
\newblock {Improved Training of Wasserstein GANs}.
\newblock In {\em Advances in Neural Information Processing Systems (NeurIPS)},
  pages 5767--5777, 2017.

\bibitem{Kodali2018OnCA}
Naveen Kodali, James Hays, Jacob~D. Abernethy, and Zsolt Kira.
\newblock {On Convergence and Stability of GANs}.
\newblock {\em arXiv preprint arXiv 1705.07215}, 2018.

\bibitem{zhou2019lipschitz}
Zhiming Zhou, Jiadong Liang, Yuxuan Song, Lantao Yu, Hongwei Wang, Weinan
  Zhang, Yong Yu, and Zhihua Zhang.
\newblock {Lipschitz generative adversarial nets}.
\newblock In {\em Proceedings of the International Conference on Machine
  Learning (ICML)}, pages 7584--7593, 2019.

\bibitem{Zhang2019ConsistencyRF}
Han Zhang, Zizhao Zhang, Augustus Odena, and Honglak Lee.
\newblock {Consistency Regularization for Generative Adversarial Networks}.
\newblock In {\em Proceedings of the International Conference on Learning
  Representations (ICLR)}, 2020.

\bibitem{Zhao2020ImprovedCR}
Zhengli Zhao, Sameer Singh, Honglak Lee, Zizhao Zhang, Augustus Odena, and Han
  Zhang.
\newblock {Improved Consistency Regularization for GANs}.
\newblock {\em arXiv preprint arXiv 2002.04724}, 2020.

\bibitem{Odena2017ConditionalIS}
Augustus Odena, Christopher Olah, and Jonathon Shlens.
\newblock {Conditional Image Synthesis with Auxiliary Classifier {GAN}s}.
\newblock In {\em Proceedings of the International Conference on Machine
  Learning (ICML)}, pages 2642--2651, 2017.

\bibitem{Miyato2018cGANsWP}
Takeru Miyato and Masanori Koyama.
\newblock {c{GAN}s with Projection Discriminator}.
\newblock In {\em Proceedings of the International Conference on Learning
  Representations (ICLR)}, 2018.

\bibitem{NIPS2019_8414}
Mingming Gong, Yanwu Xu, Chunyuan Li, Kun Zhang, and Kayhan Batmanghelich.
\newblock {Twin Auxilary Classifiers GAN}.
\newblock In {\em Advances in Neural Information Processing Systems (NeurIPS)},
  2019.

\bibitem{Siarohin2019WhiteningAC}
Aliaksandr Siarohin, Enver Sangineto, and Nicu Sebe.
\newblock {Whitening and Coloring Batch Transform for GANs}.
\newblock In {\em Proceedings of the International Conference on Learning
  Representations (ICLR)}, 2019.

\bibitem{Liu_2019_ICCV}
Ming-Yu Liu, Xun Huang, Arun Mallya, Tero Karras, Timo Aila, Jaakko Lehtinen,
  and Jan Kautz.
\newblock {Few-Shot Unsupervised Image-to-Image Translation}.
\newblock In {\em Proceedings of the International Conference on Computer
  Vision (ICCV)}, 2019.

\bibitem{kang2020contragan}
Minguk Kang and Jaesik Park.
\newblock {ContraGAN: Contrastive Learning for Conditional Image Generation}.
\newblock In {\em Advances in Neural Information Processing Systems (NeurIPS)},
  2020.

\bibitem{shim2020circlegan}
Woohyeon Shim and Minsu Cho.
\newblock {CircleGAN: Generative Adversarial Learning across Spherical
  Circles}.
\newblock In {\em Advances in Neural Information Processing Systems (NeurIPS)},
  2020.

\bibitem{zhou2020omni}
Peng Zhou, Lingxi Xie, Bingbing Ni, and Qi~Tian.
\newblock {Omni-GAN: On the Secrets of cGANs and Beyond}.
\newblock {\em arXiv preprint arXiv:2011.13074}, 2020.

\bibitem{kavalerov2021multi}
Ilya Kavalerov, Wojciech Czaja, and Rama Chellappa.
\newblock {A multi-class hinge loss for conditional gans}.
\newblock In {\em Proceedings of the IEEE/CVF Winter Conference on Applications
  of Computer Vision (WACV)}, 2021.

\bibitem{hou2021cgans}
Liang Hou, Qi~Cao, Huawei Shen, and Xueqi Cheng.
\newblock {cGANs with Auxiliary Discriminative Classifier}.
\newblock {\em arXiv preprint arXiv:2107.10060}, 2021.

\bibitem{Han_2021_ICCV}
{Han, Ligong and Min, Martin Renqiang and Stathopoulos, Anastasis and Tian, Yu
  and Gao, Ruijiang and Kadav, Asim and Metaxas, Dimitris N.}
\newblock {Dual Projection Generative Adversarial Networks for Conditional
  Image Generation}.
\newblock In {\em Proceedings of the International Conference on Computer
  Vision (ICCV)}, 2021.

\bibitem{Zhang2019SelfAttentionGA}
Han Zhang, Ian Goodfellow, Dimitris Metaxas, and Augustus Odena.
\newblock {Self-Attention Generative Adversarial Networks}.
\newblock In {\em Proceedings of the International Conference on Machine
  Learning (ICML)}, pages 7354--7363, 2019.

\bibitem{Krizhevsky2009LearningML}
Alex Krizhevsky.
\newblock {\em {Learning Multiple Layers of Features from Tiny Images}}.
\newblock PhD thesis, University of Toronto, 2012.

\bibitem{Deng2009ImageNetAL}
Jia Deng, Wei Dong, Richard Socher, Li-Jia Li, Kai Li, and Fei-Fei Li.
\newblock {ImageNet: A large-scale hierarchical image database}.
\newblock In {\em Proceedings of the IEEE International Conference on Computer
  Vision and Pattern Recognition (CVPR)}, pages 248--255, 2009.

\bibitem{Tiny}
Johnson et~al.
\newblock {Tiny ImageNet Visual Recognition Challenge}.
\newblock \url{https://tiny-imagenet.herokuapp.com}.

\bibitem{WelinderEtal2010}
P.~Welinder, S.~Branson, T.~Mita, C.~Wah, F.~Schroff, S.~Belongie, and
  P.~Perona.
\newblock {Caltech-UCSD Birds 200}.
\newblock Technical report, California Institute of Technology, 2010.

\bibitem{Heusel2017GANsTB}
Martin Heusel, Hubert Ramsauer, Thomas Unterthiner, Bernhard Nessler, and Sepp
  Hochreiter.
\newblock {GANs Trained by a Two Time-Scale Update Rule Converge to a Local
  Nash Equilibrium}.
\newblock In {\em Advances in Neural Information Processing Systems (NeurIPS)},
  pages 6626--6637, 2017.

\bibitem{Nowozin2016fGANTG}
Sebastian Nowozin, Botond Cseke, and Ryota Tomioka.
\newblock {f-{GAN}: Training Generative Neural Samplers using Variational
  Divergence Minimization}.
\newblock In {\em Advances in Neural Information Processing Systems (NeurIPS)},
  pages 271--279, 2016.

\bibitem{srivastava2017veegan}
Akash Srivastava, Lazar Valkov, Chris Russell, Michael~U Gutmann, and Charles~A
  Sutton.
\newblock {VEEGAN: Reducing Mode Collapse in GANs using Implicit Variational
  Learning}.
\newblock In {\em Advances in Neural Information Processing Systems (NeurIPS)},
  2017.

\bibitem{Mirza2014ConditionalGA}
Mehdi Mirza and Simon Osindero.
\newblock {Conditional Generative Adversarial Nets}.
\newblock {\em arXiv preprint arXiv 1411.1784}, 2014.

\bibitem{zhou2018activation}
Zhiming Zhou, Han Cai, Shu Rong, Yuxuan Song, Kan Ren, Weinan Zhang, Jun Wang,
  and Yong Yu.
\newblock {Activation Maximization Generative Adversarial Nets}.
\newblock In {\em Proceedings of the International Conference on Learning
  Representations (ICLR)}, 2018.

\bibitem{Chen2020ASF}
Ting Chen, Simon Kornblith, Mohammad Norouzi, and Geoffrey~E. Hinton.
\newblock {A Simple Framework for Contrastive Learning of Visual
  Representations}.
\newblock {\em arXiv preprint arXiv 2002.05709}, 2020.

\bibitem{studiogan}
Minguk Kang and Jaesik Park.
\newblock {Pytorch-StudioGAN}.
\newblock \url{https://github.com/POSTECH-CVLab/PyTorch-StudioGAN}, 2020.

\bibitem{choi2020starganv2}
Yunjey Choi, Youngjung Uh, Jaejun Yoo, and Jung-Woo Ha.
\newblock {StarGAN v2: Diverse Image Synthesis for Multiple Domains}.
\newblock In {\em Proceedings of the IEEE International Conference on Computer
  Vision and Pattern Recognition (CVPR)}, 2020.

\bibitem{Salimans2016ImprovedTF}
Tim Salimans, Ian Goodfellow, Wojciech Zaremba, Vicki Cheung, Alec Radford,
  Xi~Chen, and Xi~Chen.
\newblock {Improved Techniques for Training GANs}.
\newblock In {\em Advances in Neural Information Processing Systems (NeurIPS)},
  pages 2234--2242, 2016.

\bibitem{sajjadi2018assessing}
Mehdi~SM Sajjadi, Olivier Bachem, Mario Lucic, Olivier Bousquet, and Sylvain
  Gelly.
\newblock {Assessing generative models via precision and recall}.
\newblock In {\em Advances in Neural Information Processing Systems (NeurIPS)},
  2018.

\bibitem{Dumoulin2017ALR}
Vincent Dumoulin, Jonathon Shlens, and Manjunath Kudlur.
\newblock {A Learned Representation For Artistic Style}.
\newblock In {\em Proceedings of the International Conference on Learning
  Representations (ICLR)}, 2017.

\bibitem{de_Vries}
Harm de~Vries, Florian Strub, Jeremie Mary, Hugo Larochelle, Olivier Pietquin,
  and Aaron~C Courville.
\newblock {Modulating early visual processing by language}.
\newblock In {\em Advances in Neural Information Processing Systems (NeurIPS)},
  pages 6594--6604, 2017.

\bibitem{Lim2017GeometricG}
Jae~Hyun Lim and Jong~Chul Ye.
\newblock {Geometric GAN}.
\newblock {\em arXiv preprint arXiv 1705.02894}, 2017.

\bibitem{nichol2021improved}
Alex Nichol and Prafulla Dhariwal.
\newblock {Improved denoising diffusion probabilistic models}.
\newblock In {\em Proceedings of the International Conference on Machine
  Learning (ICML)}, 2021.

\bibitem{kim2021score}
Dongjun Kim, Seungjae Shin, Kyungwoo Song, Wanmo Kang, and Il-Chul Moon.
\newblock {Score Matching Model for Unbounded Data Score}.
\newblock {\em arXiv preprint arXiv:2106.05527}, 2021.

\bibitem{deng2019arcface}
Jiankang Deng, Jia Guo, Niannan Xue, and Stefanos Zafeiriou.
\newblock Arcface: Additive angular margin loss for deep face recognition.
\newblock In {\em Proceedings of the IEEE International Conference on Computer
  Vision and Pattern Recognition (CVPR)}, 2019.

\bibitem{Kingma2015AdamAM}
Diederik~P. Kingma and Jimmy Ba.
\newblock {Adam: A Method for Stochastic Optimization}.
\newblock {\em arXiv preprint arXiv 1412.6980}, 2015.

\bibitem{Sinha2020TopkTO}
Samarth Sinha, Zhengli Zhao, Anirudh Goyal, Colin Raffel, and Augustus Odena.
\newblock {Top-k Training of GANs: Improving GAN Performance by Throwing Away
  Bad Samples}.
\newblock In {\em Advances in Neural Information Processing Systems (NeurIPS)},
  2020.

\bibitem{Ravuri2019ClassificationAS}
Suman~V. Ravuri and Oriol Vinyals.
\newblock {Classification Accuracy Score for Conditional Generative Models}.
\newblock In {\em Advances in Neural Information Processing Systems (NeurIPS)},
  2019.

\bibitem{Kynknniemi2019ImprovedPA}
Tuomas Kynk{\"a}{\"a}nniemi, Tero Karras, Samuli Laine, Jaakko Lehtinen, and
  Timo Aila.
\newblock {Improved Precision and Recall Metric for Assessing Generative
  Models}.
\newblock In {\em Advances in Neural Information Processing Systems (NeurIPS)},
  2019.

\bibitem{ferjad2020icml}
Muhammad~Ferjad Naeem, Seong~Joon Oh, Youngjung Uh, Yunjey Choi, and Jaejun
  Yoo.
\newblock {Reliable Fidelity and Diversity Metrics for Generative Models}.
\newblock In {\em Proceedings of the International Conference on Machine
  Learning (ICML)}, 2020.

\bibitem{Morozov2021OnSI}
Stanislav Morozov, Andrey Voynov, and Artem Babenko.
\newblock On self-supervised image representations for gan evaluation.
\newblock In {\em Proceedings of the International Conference on Learning
  Representations (ICLR)}, 2021.

\bibitem{jeong2021training}
Jongheon Jeong and Jinwoo Shin.
\newblock {Training GANs with Stronger Augmentations via Contrastive
  Discriminator}.
\newblock In {\em Proceedings of the International Conference on Learning
  Representations (ICLR)}, 2021.

\bibitem{grill2020bootstrap}
Jean-Bastien Grill, Florian Strub, Florent Altch{\'e}, Corentin Tallec, Pierre
  Richemond, Elena Buchatskaya, Carl Doersch, Bernardo Pires, Zhaohan Guo,
  Mohammad Azar, et~al.
\newblock {Bootstrap Your Own Latent: A new approach to self-supervised
  learning}.
\newblock In {\em Advances in Neural Information Processing Systems (NeurIPS)},
  2020.

\bibitem{micikevicius2018mixed}
Paulius Micikevicius, Sharan Narang, Jonah Alben, Gregory Diamos, Erich Elsen,
  David Garcia, Boris Ginsburg, Michael Houston, Oleksii Kuchaiev, Ganesh
  Venkatesh, et~al.
\newblock {Mixed Precision Training}.
\newblock In {\em Proceedings of the International Conference on Learning
  Representations (ICLR)}, 2018.

\bibitem{pmlr-v37-ioffe15}
Sergey Ioffe and Christian Szegedy.
\newblock {Batch Normalization: Accelerating Deep Network Training by Reducing
  Internal Covariate Shift}.
\newblock In {\em Proceedings of the International Conference on Machine
  Learning (ICML)}, pages 448--456, 2015.

\bibitem{mo2020freeze}
Sangwoo Mo, Minsu Cho, and Jinwoo Shin.
\newblock {Freeze the discriminator: a simple baseline for fine-tuning gans}.
\newblock {\em arXiv preprint arXiv:2002.10964}, 2020.

\bibitem{che2020your}
Tong Che, Ruixiang Zhang, Jascha Sohl-Dickstein, Hugo Larochelle, Liam Paull,
  Yuan Cao, and Yoshua Bengio.
\newblock {Your GAN is secretly an energy-based model and you should use
  discriminator driven latent sampling}.
\newblock In {\em Advances in Neural Information Processing Systems (NeurIPS)},
  2020.

\bibitem{shen2021closedform}
Yujun Shen and Bolei Zhou.
\newblock Closed-form factorization of latent semantics in gans.
\newblock In {\em Proceedings of the IEEE International Conference on Computer
  Vision and Pattern Recognition (CVPR)}, 2021.

\bibitem{yazc2018the}
Yasin Yaz{\i}c{\i}, Chuan-Sheng Foo, Stefan Winkler, Kim-Hui Yap, Georgios
  Piliouras, and Vijay Chandrasekhar.
\newblock {The Unusual Effectiveness of Averaging in {GAN} Training}.
\newblock In {\em Proceedings of the International Conference on Learning
  Representations (ICLR)}, 2019.

\bibitem{chuang2020debiased}
Ching-Yao Chuang, Joshua Robinson, Lin Yen-Chen, Antonio Torralba, and Stefanie
  Jegelka.
\newblock {Debiased contrastive learning}.
\newblock In {\em Advances in Neural Information Processing Systems (NeurIPS)},
  2020.

\bibitem{huynh2020boosting}
Tri Huynh, Simon Kornblith, Matthew~R Walter, Michael Maire, and Maryam
  Khademi.
\newblock {Boosting Contrastive Self-Supervised Learning with False Negative
  Cancellation}.
\newblock {\em arXiv preprint arXiv:2011.11765}, 2020.

\bibitem{robinson2020contrastive}
Joshua Robinson, Ching-Yao Chuang, Suvrit Sra, and Stefanie Jegelka.
\newblock {Contrastive learning with hard negative samples}.
\newblock In {\em Proceedings of the International Conference on Learning
  Representations (ICLR)}, 2021.

\bibitem{Szegedy2016RethinkingTI}
Christian Szegedy, Vincent Vanhoucke, Sergey Ioffe, Jon Shlens, and Zbigniew
  Wojna.
\newblock {Rethinking the Inception Architecture for Computer Vision}.
\newblock In {\em Proceedings of the IEEE International Conference on Computer
  Vision and Pattern Recognition (CVPR)}, pages 2818--2826, 2016.

\bibitem{kupyn2018deblurgan}
Orest Kupyn, Volodymyr Budzan, Mykola Mykhailych, Dmytro Mishkin, and
  Ji{\v{r}}{\'\i} Matas.
\newblock {Deblurgan: Blind motion deblurring using conditional adversarial
  networks}.
\newblock In {\em Proceedings of the IEEE International Conference on Computer
  Vision and Pattern Recognition (CVPR)}, 2018.

\bibitem{yang2019controllable}
Shuai Yang, Zhangyang Wang, Zhaowen Wang, Ning Xu, Jiaying Liu, and Zongming
  Guo.
\newblock Controllable artistic text style transfer via shape-matching gan.
\newblock In {\em Proceedings of the International Conference on Computer
  Vision (ICCV)}, 2019.

\bibitem{shetty2018adversarial}
Rakshith Shetty, Mario Fritz, and Bernt Schiele.
\newblock Adversarial scene editing: Automatic object removal from weak
  supervision.
\newblock In {\em Advances in Neural Information Processing Systems (NeurIPS)},
  2018.

\bibitem{chen2018sketchygan}
Wengling Chen and James Hays.
\newblock {Sketchygan: Towards diverse and realistic sketch to image
  synthesis}.
\newblock In {\em Proceedings of the IEEE International Conference on Computer
  Vision and Pattern Recognition (CVPR)}, 2018.

\bibitem{chen2018cartoongan}
Yang Chen, Yu-Kun Lai, and Yong-Jin Liu.
\newblock {Cartoongan: Generative adversarial networks for photo
  cartoonization}.
\newblock In {\em Proceedings of the IEEE International Conference on Computer
  Vision and Pattern Recognition (CVPR)}, 2018.

\bibitem{yeh2017semantic}
Raymond~A Yeh, Chen Chen, Teck Yian~Lim, Alexander~G Schwing, Mark
  Hasegawa-Johnson, and Minh~N Do.
\newblock Semantic image inpainting with deep generative models.
\newblock In {\em Proceedings of the IEEE International Conference on Computer
  Vision and Pattern Recognition (CVPR)}, pages 5485--5493, 2017.

\bibitem{nguyen2019deep}
Thanh~Thi Nguyen, Cuong~M Nguyen, Dung~Tien Nguyen, Duc~Thanh Nguyen, and Saeid
  Nahavandi.
\newblock Deep learning for deepfakes creation and detection: A survey.
\newblock {\em arXiv preprint arXiv:1909.11573}, 2019.

\bibitem{Xu2018FairGANFG}
Depeng Xu, Shuhan Yuan, Lu~Zhang, and Xintao Wu.
\newblock {FairGAN: Fairness-aware Generative Adversarial Networks}.
\newblock {\em International Conference on Big Data (Big Data)}, pages
  570--575, 2018.

\bibitem{masi2020two}
Iacopo Masi, Aditya Killekar, Royston~Marian Mascarenhas, Shenoy~Pratik
  Gurudatt, and Wael AbdAlmageed.
\newblock Two-branch recurrent network for isolating deepfakes in videos.
\newblock In {\em Proceedings of the European Conference on Computer Vision
  (ECCV)}, 2020.

\bibitem{naseer2020self}
Muzammal Naseer, Salman Khan, Munawar Hayat, Fahad~Shahbaz Khan, and Fatih
  Porikli.
\newblock A self-supervised approach for adversarial robustness.
\newblock In {\em Proceedings of the IEEE International Conference on Computer
  Vision and Pattern Recognition (CVPR)}, pages 262--271, 2020.

\end{thebibliography}
}
\clearpage
\appendix
\section*{\Large{Appendices}}
\addcontentsline{toc}{section}{Appendices}
\renewcommand\thefigure{A\arabic{figure}}
\renewcommand{\thetable}{A\arabic{table}}
\setcounter{figure}{0}
\setcounter{table}{0}
\setcounter{property}{0}
\section{Algorithm}
\label{algorithm}
\renewcommand{\algorithmicrequire}{\textbf{Input:}}
\renewcommand{\algorithmicprocedure}{\textbf{For}}
\renewcommand{\algorithmicensure}{\textbf{Output:}}
\begin{algorithm}[H]
\caption{: Training ReACGAN}\label{alg:algorithm}
\begin{algorithmic}[1]
\Require{Batch size: $N$. Temperature: $\tau$. Margin values: $m_p$, $m_n$. Balance coefficient:~$\lambda$.}
\Statex \quad \; Parameters of the generator: $\theta$.
\Statex \quad \; Parameters of the discriminator (feature extractor $+$ adversarial head $+$ linear head): $\phi$.  
\Statex \quad \;  \# of discriminator updates per single generator update: $n_{dis}$.
\Statex \quad \;  Adversarial loss: $\mathcal{L}_{\text{Adv}}$~\cite{Goodfellow2014GenerativeAN, Mao2017LeastSG, Gulrajani2017ImprovedTO, Lim2017GeometricG}.
\Statex \quad \; Learning rate: $\alpha_{1}, \alpha_{2}$. Adam hyperparameters~\cite{Kingma2015AdamAM}: $\beta_{1}, \beta_{2}$.
\Ensure{Optimized $(\theta, \phi)$.}
\Statex
\State Initialize $(\theta, \phi)$
\For{$\{1, ..., $ \# of training iterations$\}$} 
    \For{$\{1, ..., n_\mathrm{dis}\}$}
    \State Sample $\mX, \vy^{\mathrm{real}} = \{\vx_{i}\}_{i=1}^{N}, \{y_{i}\}_{i=1}^{N} \sim p_\mathrm{real}(\rvx,\rvy)$
    \State Sample $\mZ = \{\vz_{i}\}_{i=1}^{N} \sim p(\rvz)$ and $\vy^{\mathrm{fake}} = \{y_{i}^\mathrm{fake}\}_{i=1}^{N} \sim P(\rvy)$
    \State $\mathcal{L}_{\text{D\_Adv}} \longleftarrow \mathcal{L}_{\text{Adv}}(\mX, \vy^{\mathrm{real}}, G(\mZ, \vy^{\mathrm{fake}}), \vy^{\mathrm{fake}})$
    \State $\mathcal{L}_{\text{D\_Cond}} \longleftarrow \mathcal{L}_{\text{D2D-CE}}(\mX, \vy^{\mathrm{real}}; \tau, m_p, m_n)$ 
    \Comment{Eq. (\ref{eq:eq6}) with real images.}
    \State $\phi \longleftarrow \text{Adam}(\mathcal{L}_{\text{D\_Adv}} + \lambda\mathcal{L}_{\text{D\_Cond}}, \alpha_{1}, \beta_1, \beta_2)$
    \EndFor
\State Sample $\mZ = \{\vz_{i}\}_{i=1}^{N} \sim p(\rvz)$ and $\vy^{\mathrm{fake}} = \{y_{i}^\mathrm{fake}\}_{i=1}^{N} \sim P(\rvy)$
\State $\mathcal{L}_{\text{G\_Adv}} \longleftarrow \mathcal{L}_{\text{Adv}}(G(\mZ, \vy^{\mathrm{fake}}), \vy^{\mathrm{fake}})$
\State $\mathcal{L}_{\text{G\_Cond}} \longleftarrow \mathcal{L}_{\text{D2D-CE}}(G(\mZ, \vy^{\mathrm{fake}}), \vy^{\mathrm{fake}}; \tau, m_p, m_n)$ 
\Comment{Eq. (\ref{eq:eq6}) with fake images.}
\State $\theta \longleftarrow \text{Adam}(\mathcal{L}_{\text{G\_Adv}} + \lambda\mathcal{L}_{\text{G\_Cond}}, \alpha_{2}, \beta_1, \beta_2)$
\EndFor
\end{algorithmic}
\end{algorithm}
\section{Software: PyTorch-StudioGAN}
Generative Adversarial Network~(GAN) is one of the popular generative models for realistic image generation. Although GAN has been actively studied in the machine learning community, only a few open-source libraries provide reliable implementations for GAN training. In addition, the existing libraries do not support various training and test configurations for loss functions, backbone architectures, regularizations, differentiable augmentations, and evaluation metrics. In this paper, we expand StudioGAN~\cite{studiogan} library, and the StudioGAN provides about 40 implementations of GAN-related papers as follows:

\textbf{GANs}: DCGAN~\cite{Radford2016UnsupervisedRL}, LSGAN~\cite{Mao2017LeastSG}, GGAN~\cite{Lim2017GeometricG}, WGAN-WC~\cite{Arjovsky2017WassersteinG}, WGAN-GP~\cite{Gulrajani2017ImprovedTO}, WGAN-DRA~\cite{Kodali2018OnCA}, ACGAN~\cite{Odena2017ConditionalIS}, Projection discriminator~\cite{Miyato2018cGANsWP}, SNGAN~\cite{Miyato2018SpectralNF}, SAGAN~\cite{Zhang2019SelfAttentionGA}, TACGAN~\cite{NIPS2019_8414},  LGAN~\cite{zhou2019lipschitz}, BigGAN~\cite{Brock2019LargeSG}, BigGAN-deep~\cite{Brock2019LargeSG}, StyleGAN2~\cite{karras2020analyzing}, CRGAN~\cite{Zhang2019ConsistencyRF}, ICRGAN~\cite{Zhao2020ImprovedCR}, LOGAN~\cite{Wu2019LOGANLO}, MHGAN~\cite{kavalerov2021multi}, ContraGAN~\cite{kang2020contragan}, ADCGAN~\cite{hou2021cgans}, ReACGAN~(ours).

\textbf{Adversarial losses}: Logistic loss~\cite{karras2020analyzing}, Non-saturation loss~\cite{Goodfellow2014GenerativeAN}, Least square loss~\cite{Mao2017LeastSG}, Wasserstein loss~\cite{Arjovsky2017WassersteinG}, Hinge loss~\cite{Lim2017GeometricG}, Multiple discriminator loss~\cite{Liu_2019_ICCV}, Multi-hinge loss~\cite{kavalerov2021multi}.

\textbf{Regularizations}: Feature matching regularization~\cite{Salimans2016ImprovedTF}, R1 regularization~\cite{Liu_2019_ICCV}, Weight clipping regularization~\cite{Arjovsky2017WassersteinG}, Spectral normalization~\cite{Miyato2018SpectralNF}, Path length regularization~\cite{karras2019style, karras2020analyzing}, Top-k training~\cite{Sinha2020TopkTO}.

\textbf{Metrics}: IS~\cite{Salimans2016ImprovedTF}, FID~\cite{Heusel2017GANsTB}, Intra-class FID, CAS~\cite{Ravuri2019ClassificationAS}, Precision and recall~\cite{sajjadi2018assessing}, Improved precision and recall~\cite{Kynknniemi2019ImprovedPA}, Density and coverage~\cite{ferjad2020icml}, SwAV backbone FID~\cite{Morozov2021OnSI}.

\textbf{Differentiable augmentations}: SimCLR augmentation~\cite{Chen2020ASF, jeong2021training}, BYOL augmentation~\cite{grill2020bootstrap, jeong2021training}, DiffAugment~\cite{zhao2020differentiable}, Adaptive discriminator augmentation~(ADA)~\cite{Karras2020TrainingGA}.

\textbf{Miscellaneous}: Mixed precision training~\cite{micikevicius2018mixed}, Distributed data parallel~(DDP), Data parallel~(DP), Synchronized batch normalization~\cite{pmlr-v37-ioffe15}, Standing statistics~\cite{Brock2019LargeSG}, Truncation trick~\cite{Brock2019LargeSG, karras2020analyzing}, Freeze discriminator~(FreezeD)~\cite{mo2020freeze}, Discriminator driven latent sampling~(DDLS)~\cite{che2020your}, Closed-form factorization (SeFa)~\cite{shen2021closedform}.
\section{Training Details}
\subsection{\large{Datasets}}
\label{dataset}
\textbf{\text{CIFAR10}}~\cite{Krizhevsky2009LearningML} is a widely used benchmark dataset for evaluating cGANs. The dataset contains 60k $32\times32$ RGB images which belong to 10 different classes. The dataset is split into 50k images for training and 10k images for testing.

\textbf{\text{Tiny-ImageNet}}~\cite{Tiny} contains 120k $64\times64$ RGB images and is split into 100k training, 10k validation, and 10k test images. \text{Tiny-ImageNet} consists of 200 categories, and training GANs on \text{Tiny-ImageNet} is more challenging than CIFAR10 since there is less data~(500 images) per class.

\textbf{\text{CUB200}}~\cite{WelinderEtal2010} provides around 12k fine-grained RGB images for 200 bird classes. We apply the center crop to each image using a square box whose lengths are the same as the short side of the image, and we resize the images to 128$\times$128 pixels. We train cGANs on CUB200 dataset to identify the generation ability of cGANs on images with fine-grained characteristics in a limited data situation.

\textbf{\text{ImageNet}}~\cite{Deng2009ImageNetAL} provides around 1,281k and 50k RGB images for training and validation. We preprocess each image in the same way as applied to CUB200.

\textbf{\text{AFHQ}}~\cite{choi2020starganv2} consists of 14,630 and 1,500 numbers of $512\times512$ RGB images for training and validation. The dataset is divided into 3 different animal classes~(cat, dog, and wild animals).

For training and testing, we apply horizontal flip augmentation for all datasets and normalize image pixel values to a range between -1 and 1.\\

\subsection{\large{Hyperparameter Setup}}
\label{hyperparameter_setup}

Selecting proper hyperparameter values greatly affects GAN training. So it might be helpful to specify details of hyperparameter setups used in our work for future study. In this section, we aim to provide training specifications as much as possible, and if there exists a missing experimental setup, it follows configurations and details of StudioGAN implementation~\cite{studiogan}.

\begin{table}[!ht]
\centering
\vspace{-2.5mm}
\caption{Hyperparameter setups for cGAN training. The settings (A, C, E, I, K, M, P, R) are commonly used practices in previous studies~\cite{Zhang2019SelfAttentionGA, Brock2019LargeSG, kang2020contragan, Karras2020TrainingGA}} 
\vspace{2mm}
  \resizebox{0.98\textwidth}{!}{
\begin{tabular}{lcccccccc}
\cmidrule[1.0pt]{1-9}
Setting & Batch size& Adam$(\alpha_1, \alpha_2, \beta_1$, $\beta_2)$~\cite{Kingma2015AdamAM} & $n_{dis}$ & ($\lambda, \tau$) & $m_p$ & G\_Ema~\cite{yazc2018the} & Ema start & Total iterations\\
\cmidrule[1.0pt]{1-9}
A & 64 & (2e-4, 2e-4, 0.5, 0.999) & 5 & - & - & True & 1k &  100k\\
B & 128 & (2.82e-4, 2.82e-4, 0.5, 0.999) & 5 & (0.5, 0.5) & 0.98 & True & 1k &  100k\\
C & 64 & (2e-4, 2e-4, 0.5, 0.999) & 5 & - & - & True & 1k &  200k\\
D & 128 & (2.82e-4, 2.82e-4, 0.5, 0.999) & 5 & (0.5, 0.5) & 0.98 & True & 1k &  200k\\
E & 64 & (2.5e-3, 2.5e-3, 0.0, 0.99) & 1 & - & - & True & 0 &  200k\\
F & 64 & (2.5e-3, 2.5e-3, 0.0, 0.99) & 2 & (0.25, 0.25) & 0.98 & True & 0 &  200k\\
G & 64 & (2.5e-3, 2.5e-3, 0.0, 0.99) & 1 & - & - & True & 0 &  800k\\
H & 64 & (2.5e-3, 2.5e-3, 0.0, 0.99) & 2 & (0.25, 0.25) & 0.98 & True & 0 &  800k\\
\midrule
I & 1024 & (4e-4, 1e-4, 0.0, 0.999) & 1 & - & - & True & 20k &  100k\\
J & 1024 & (4e-4, 1e-4, 0.0, 0.999) & 1 & (0.75, 0.75) & 1.0 & True & 20k &  100k\\
\midrule
K & 256 & (2e-4, 5e-5, 0.0, 0.999) & 2 & - & - & True & 4k &  40k\\
L & 256 & (2e-4, 5e-5, 0.0, 0.999) & 2 & (0.25, 0.25) & 0.95 & True & 4k &  40k\\
\midrule
M & 256 & (2e-4, 5e-5, 0.0, 0.999) & 2 & - & - & True & 20k & 200k\\
N & 256 & (2e-4, 5e-5, 0.0, 0.999) & 2 & - & - & True & 20k &  600k\\
O & 256 & (2e-4, 5e-5, 0.0, 0.999) & 2 & (1.0, 0.5) & 0.98 & True & 20k &  600k\\
P & 2048 & (2e-4, 5e-5, 0.0, 0.999) & 2 & - & - & True & 20k &  200k\\
Q & 2048 & (2e-4, 5e-5, 0.0, 0.999) & 2 & (0.5, 0.25) & 0.90 & True & 20k &  200k\\
\midrule
R & 64 & (2.5e-3, 2.5e-3, 0.0, 0.99) & 1 & - & - & True & 0 &  200k\\
S & 64 & (2.5e-3, 2.5e-3, 0.0, 0.99) & 2 & (0.5, 0.5) & 0.95 & True & 0 &  200k\\
\cmidrule[1.0pt]{1-9}
\end{tabular}}
\label{table:hyper}
\end{table}
\begin{figure}[ht]
    \vspace{-1mm}
    \centering
    \includegraphics[width=0.48\linewidth]{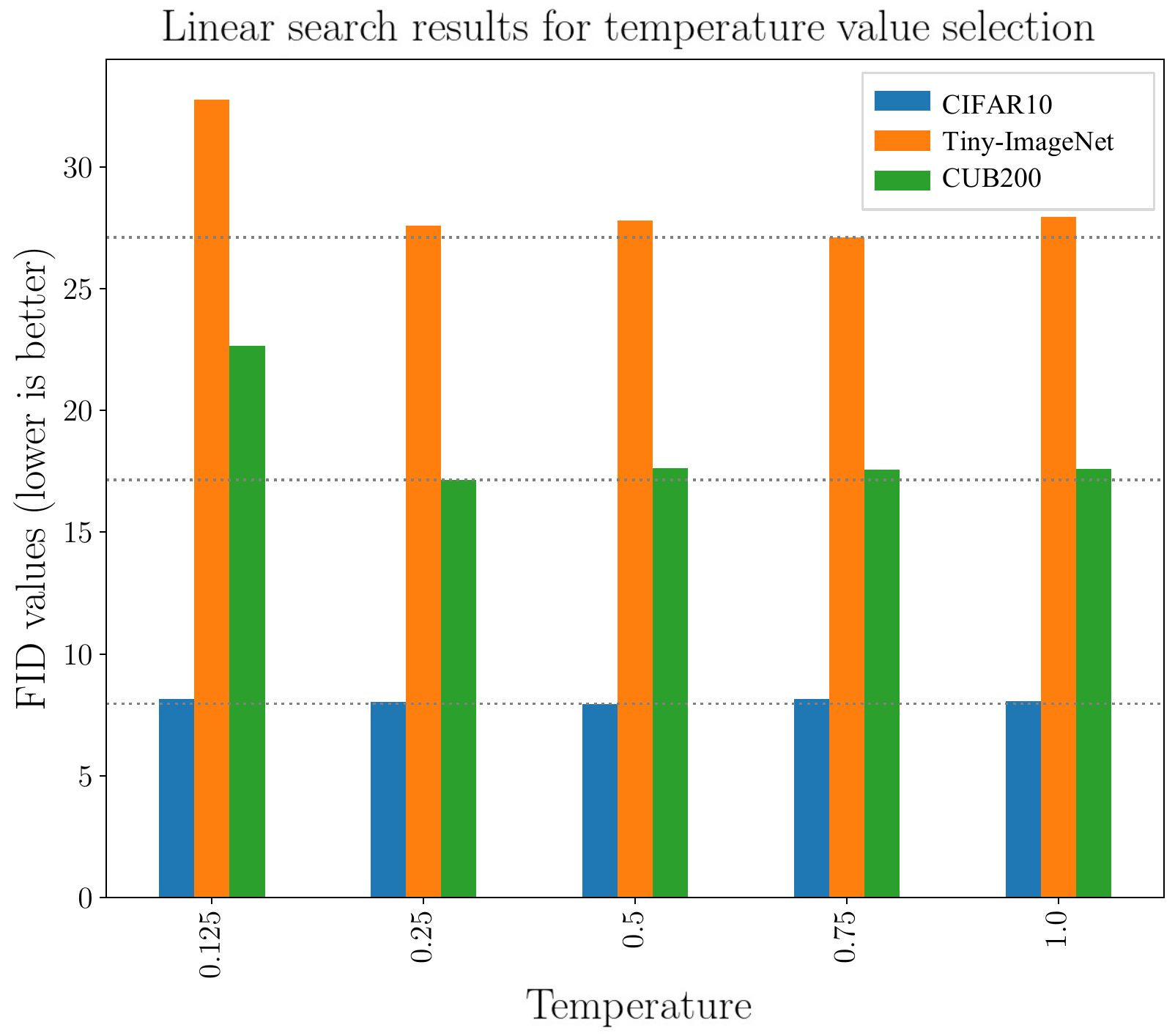}
    \includegraphics[width=0.47\linewidth]{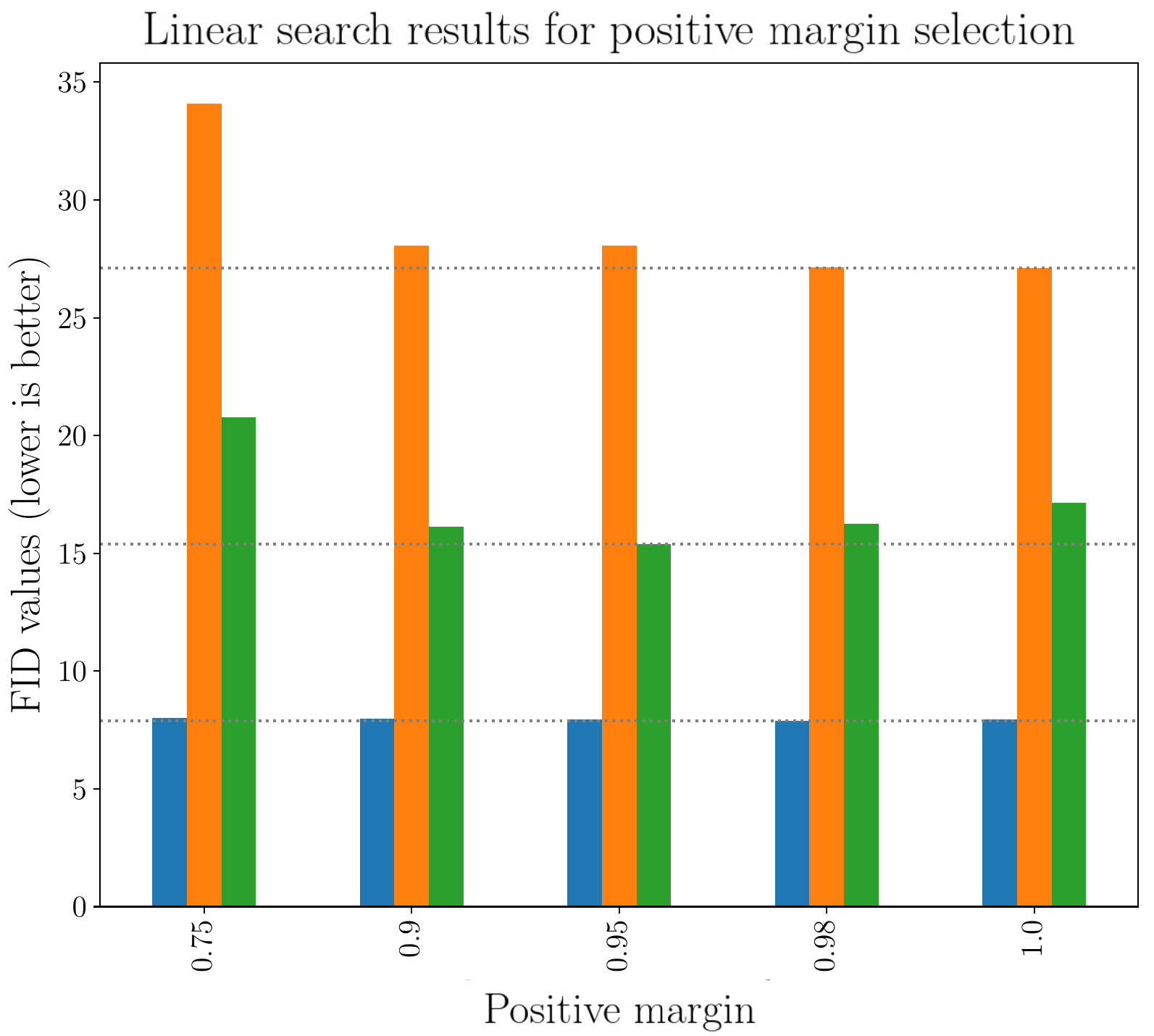}
    \caption{FID~\cite{Heusel2017GANsTB} values of ReACGANs with different temperatures and positive margins. The dotted lines indicate the best performances on each dataset.}
    \label{fig:gridsearch}
\end{figure}
Table~\ref{table:hyper} shows hyperparameter setups used in our experiments. The settings (A, C, E, G) are used for baseline experiments on CIFAR10: BigGAN~\cite{Brock2019LargeSG}, BigGAN with DiffAug~\cite{zhao2020differentiable}, StyleGAN2~\cite{karras2020analyzing}, and StyleGAN2 with ADA~\cite{Karras2020TrainingGA}, the setting (I) on Tiny-ImageNet: BigGAN and BigGAN with DiffAug, the setting (K) on CUB200: BigGAN and BigGAN with DiffAug, the settings (M, N, P) on ImageNet: ACGAN/SNGAN/ContraGAN, BigGAN/ReACGAN/DiffAug-BigGAN/DiffAug-ReACGAN with a batch size of 256, and BigGAN with a batch size of 2048, and the setting (R) on AFHQ: StyleGAN2 with ADA. For ReACGAN experiments, we utilize the settings (B, D, F, H, J, L, O, Q, S) for experiments on the datasets stated above. To select an appropriate temperature~$\tau$ and positive margin~$m_p$, we conduct two-stage linear search with the candidates of a temperature $\tau \in \{0.125, 0.25, 0.5, 0.75, 1.0\}$ and a positive margin $m_p \in~\{0.5, 0.75, 0.9, 0.95, 0.98, 1.0\}$ while fixing the dimensionalities of feature embeddings to 512, 768, 1024, and 2048 for CIFAR10, Tiny-ImageNet, CUB200, and ImageNet experiments. We set the balance coefficient~$\lambda$ equal to the temperature~$\tau$ except for ImageNet generation experiments. Specifically, we explore the best temperature value on each dataset and fix the temperature for the linear search on the positive margin.
Linear search results are summarized in Fig.~\ref{fig:gridsearch}, and the results show that ReACGAN provides stable performances across various temperature and positive margin values.
\section{Proofs of Properties of D2D-CE}
\label{proof_of_D2DCE}
In the main paper, we introduce a new objective, the D2D-CE to train ACGAN stably. In this section, we provide proofs of the properties of D2D-CE. Using the notations defined in the main paper, our proposed D2D-CE loss can be expressed as follows:
\begin{align*}
    \mathcal{L_{\text{D2D-CE}}}=-\frac{1}{N}\sum_{i=1}^{N}\log{\bigg(\frac{\exp{\big([\vf_i^{\top}\vv_{y_i} -m_p]_{-}/\tau}\big)}
    {\exp{\big([\vf_i^{\top}\vv_{y_i}-m_p]_{-}/\tau}\big)+\sum_{j\in \mathcal{N}(i)}\exp{\big([\vf_i^{\top}\vf_j -m_n]_{+}/\tau\big)}}\bigg)},
    \label{eq:Aeq1}\tag{A1}
\end{align*}
where $i$ is a sample index and $\mathcal{N}(i)$ is the set of indices that indicate the locations of negative samples in the mini-batch. To understand properties of D2D-CE loss, we can rewrite Eq.~(\ref{eq:Aeq1}) as follows:
\begin{align*}
    \mathcal{L_{\text{D2D-CE}}}=-\frac{1}{N}\sum_{i=1}^{N}\log{\bigg(\frac{\exp{\big([s_i-m_p]_{-}/\tau}\big)}
    {\exp{\big([s_i-m_p]_{-}/\tau}\big)+\sum_{j\in \mathcal{N}(i)}\exp{\big([s_{i,j}-m_n]_{+}/\tau\big)}}\bigg)}.
    \label{eq:Aeq2}\tag{A2}
\end{align*}

Let $s_{q}$ be a similarity between a normalized reference sample embedding $\vf_q$ and the corresponding normalized proxy $\vv_{y_q}$, $s_{q, r}$ be a similarity between $\vf_q$ and one of the its negative samples $\vf_r$, and $a,b\in \mathcal{N\text{($q$)}}$ be arbitrary indices of negative samples. Then, we can summarize the four properties of D2D-CE loss as follows:

\begin{property}
Hard negative mining.
If the value of $s_{q, a}$ is greater than $s_{q, b}$, the derivative of $\mathcal{L}_{\text{D2D-CE}}$ w.r.t $s_{q, a}$ is greater than or equal to the derivative w.r.t $s_{q, b}$; that is $\frac{\partial{\mathcal{L}_{\text{D2D-CE}}}}{\partial{s_{q, a}}}$ $\geq$ $\frac{\partial{\mathcal{L}_{\text{D2D-CE}}}}{\partial{s_{q, b}}}$ $\geq 0$.
\end{property}

\begin{property}
Positive suppression.
If $s_{q}$  $ - m_p \geq 0$, the derivative of  $\mathcal{L}_{\text{D2D-CE}}$ w.r.t $s_{q}$ is $0$.
\end{property}

\begin{property}
Negative suppression. 
If $s_{q, r}$  $ - m_n \leq 0$, the derivative of  $\mathcal{L}_{\text{D2D-CE}}$ w.r.t $s_{q, r}$ is $0$.
\end{property}

\begin{property}
If $s_{q} - m_p \geq 0$ and $s_{q, r} - m_n \leq 0$ are satisfied, $\mathcal{L}_{\text{D2D-CE}}$ has the global minima of $~\frac{1}{N}\sum_{i=1}^{N}\log{(1 + |\mathcal{N}(i)|)}$.
\end{property}

\textbf{\textit{Proof of Property 1.}}
By expanding Eq.~(\ref{eq:Aeq2}), we have the following equation:
\begin{align*}
    \mathcal{L_{\text{D2D-CE}}}=
    &-\underbrace{\frac{1}{\tau N}\sum_{i=1}^{N}[s_i-m_p]_{-}}_{\text{Positive attraction}} \\
    & + \underbrace{\frac{1}{N}\sum_{i=1}^{N}\log{\bigg(\exp{\big([s_i-m_p]_{-}/\tau}\big)+\sum_{j\in \mathcal{N}(i)}\exp{\big([s_{i,j}-m_n]_{+}/\tau\big)}}\bigg)}_{\text{Negative repulsion}}.
    \label{eq:Aeq3}\tag{A3}
\end{align*}
\\
Based on this, we calculate the derivative of $\mathcal{L}_{\text{D2D-CE}}$ w.r.t $s_{q,r}$ as follows:
\begin{align*}
    \frac{\partial \mathcal{L}_{\text{D2D-CE}}}{\partial s_{q,r}} =&
    \frac{\1_{s_{i,j} - m_n >0}(i=q, j=r)\exp{\big((s_{q,r}-m_n)/\tau\big)}}
    {\tau N\bigg(\exp{\big([s_q-m_p]_{-}/\tau}\big)+\sum_{j\in \mathcal{N}(q)}\exp{\big([s_{q,j}-m_n]_{+}/\tau\big)}\bigg)}.
    \label{eq:Aeq4}\tag{A4}
\end{align*}

Since we assume $s_{q, a} > s_{q, b}$ is satisfied, the derivative of $\frac{\partial{\mathcal{L_{\text{D2D-CE}}}}}{\partial{s_{q,a}}}$ is greater than $\frac{\partial{\mathcal{L_{\text{D2D-CE}}}}}{\partial{s_{q,b}}}$ except when the indicator functions $\1_{s_{i,j} -m_{n}>0}(i=q,j=a)$ and $\1_{s_{i,j} -m_{n}>0}(i=q,j=b)$ are 0. Note that the value of the derivative $\frac{\partial \mathcal{L}_{\text{D2D-CE}}}{\partial s_{q,r}}$ exponentially increases as the similarity $s_{q,r}$ linearly increases, and this means that $\mathcal{L}_{\text{D2D-CE}}$ conducts hard negative mining.\\

\textbf{\textit{Proof of Property 2.}} Based on Eq.~(\ref{eq:Aeq3}), we can derive the derivative of $\mathcal{L}_{\text{D2D-CE}}$ w.r.t $s_q$ as follows:
\begin{align*}
    \frac{\partial \mathcal{L}_{\text{D2D-CE}}}{\partial s_{q}} =
    &-\frac{1}{\tau N}\1_{s_i - m_p < 0}(i=q)\\
    &+ \frac{\1_{s_i - m_p < 0}(i=q){\exp{\big((s_q-m_p)/\tau}\big)}}
    {\tau N \bigg(\exp{\big([s_q-m_p]_{-}/\tau}\big)+\sum_{j\in \mathcal{N}(q)}\exp{\big([s_{q,j}-m_n]_{+}/\tau\big)}\bigg)}.
    \label{eq:Aeq5}\tag{A5}
\end{align*}
Since the indicator function $\1_{s_i -m_p<0}(i=q)$ gives 0 value when $s_q -m_p \geq 0$, the derivative of $\mathcal{L}_{\text{D2D-CE}}$ w.r.t $s_{q}$ is 0 when $s_q - m_p \geq0$ is satisfied.

\textbf{\textit{Proof of Property 3.}} Based on Eq.~(\ref{eq:Aeq4}), the derivative of $\mathcal{L}_{\text{D2D-CE}}$ w.r.t $s_{q, r}$ is $0$ when $s_{q,r} - m_n  \leq 0$.

\textbf{\textit{Proof of Property 4.}} We can get the global minima of $\mathcal{L}_{\text{D2D-CE}}$ by plugging in 0 values inside of the exponential components in Eq.~(\ref{eq:Aeq3}) as follows:
\begin{align*}
    \mathcal{L_{\text{D2D-CE}}}
    &= \frac{1}{N}\sum_{i=1}^{N}\log{\bigg(\exp{\big(0}\big)+\sum_{j\in \mathcal{N}(i)}\exp{\big(0\big)}}\bigg) \\
    &= \frac{1}{N}\sum_{i=1}^{N}\log{(1 +|\mathcal{N}(i)|}).
    \label{eq:Aeq6}\tag{A6}
\end{align*}
\section{Additional Experimental Results}
\label{additional_experiments}
\begin{figure}[t!]
    \centering
    \begin{subfigure}{0.32\textwidth}
    \includegraphics[width=\linewidth]{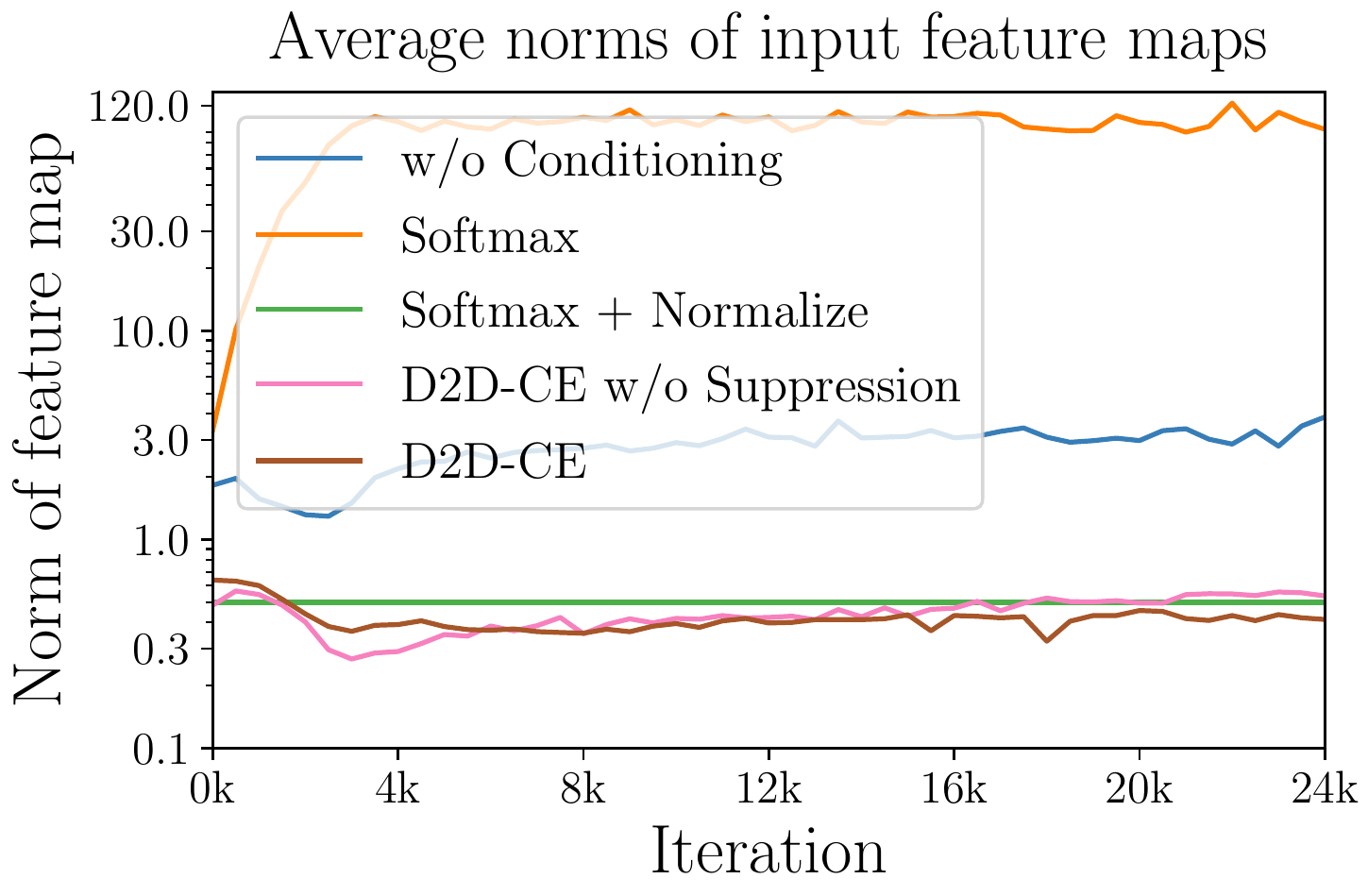}
    \caption{Feature norm} \label{fig:acgan_fnorm_sp}
    \end{subfigure}
    \begin{subfigure}{0.32\textwidth}
    \includegraphics[width=\linewidth]{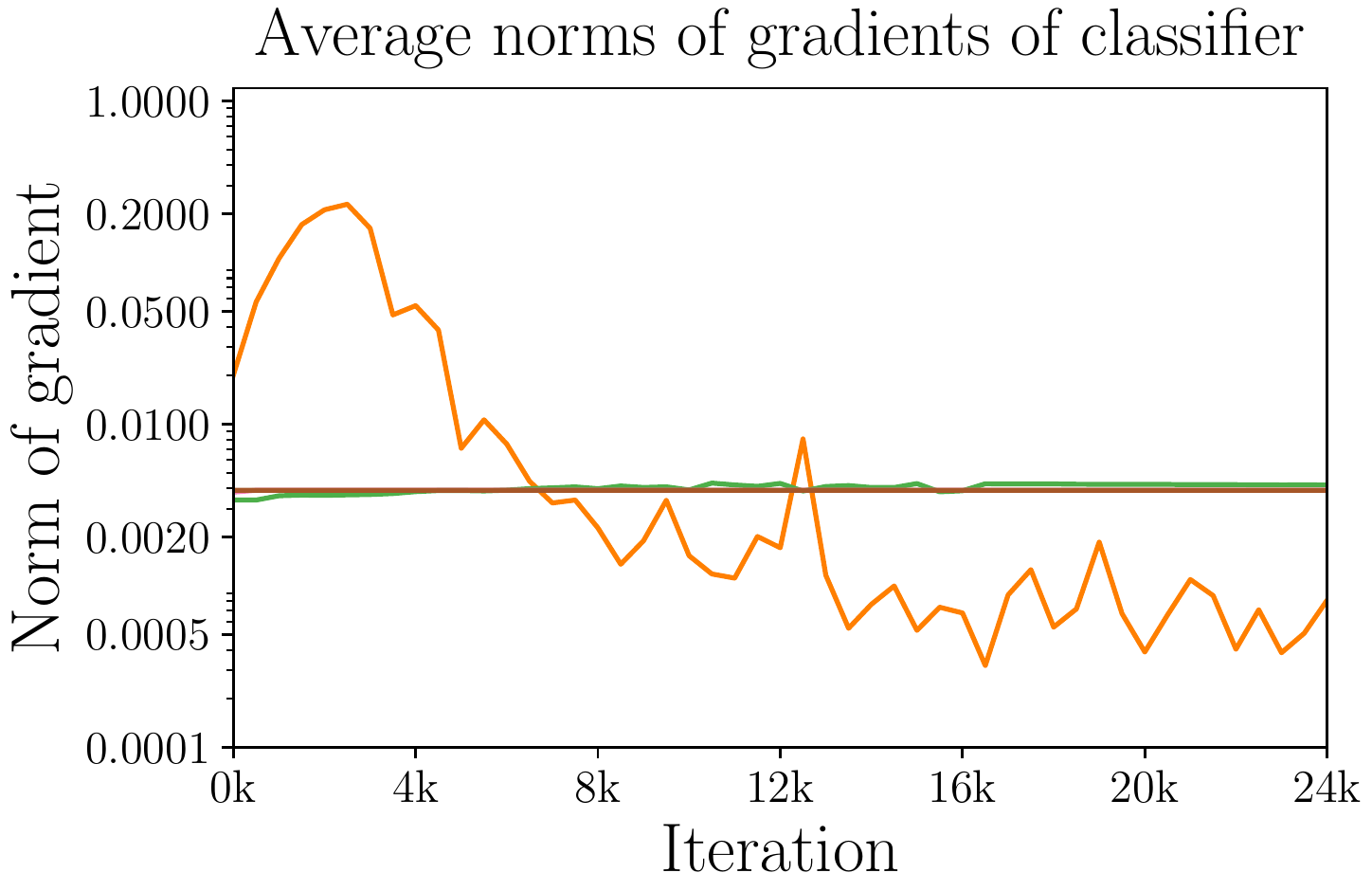}
    \caption{Gradient norm} \label{fig:acgan_gnorm_sp}
    \end{subfigure}
    \begin{subfigure}{0.32\textwidth}
    \includegraphics[width=\linewidth]{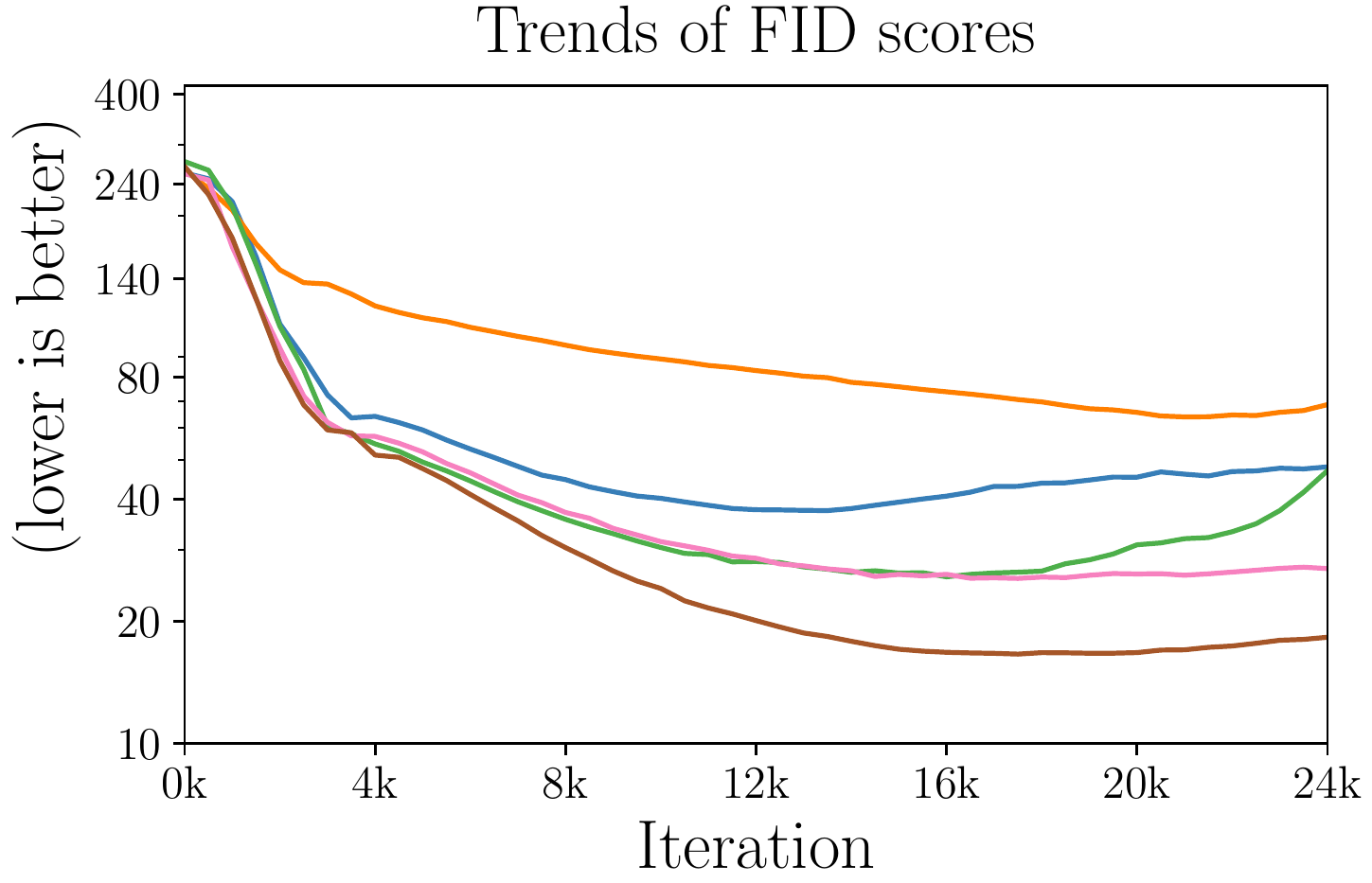}
    \caption{FID score} \label{fig:acgan_fid_sp}
    \end{subfigure}
    \caption{Merits of integrating feature normalization and data-to-data relationship consideration. The experiments are conducted using CUB200~\cite{WelinderEtal2010} dataset. (a) Average norms of input feature maps, (b) average norms of gradients of classification losses, and (c) trends of FID scores. Compared to ACGAN~\cite{Odena2017ConditionalIS}, the proposed ReACGAN does not experience the early-training collapse problem caused by excessively large norms of feature maps and gradients. In addition, ReACGAN can converge to a better equilibrium by considering data-to-data relationships with easy sample suppression.}
    \label{fig:Figure_exploding_sp}
\end{figure}

\textbf{Gradients Exploding Problem in ACGAN.} We conduct additional experiments regarding the gradient exploding problem of ACGAN using CUB200 dataset, and the results can be seen in Fig.~\ref{fig:Figure_exploding_sp}. Same as the experimental results using Tiny-ImageNet~(Fig.~\ref{fig:Figure_exploding} in the main paper), normalizing feature maps resolves the early-training collapse problem. Also, considering data-to-data relationships brings extra performance gain with easy positive and negative sample suppression.

\begin{figure}[t!]
    \centering
    \begin{subfigure}{0.40\textwidth}
    \includegraphics[width=\linewidth]{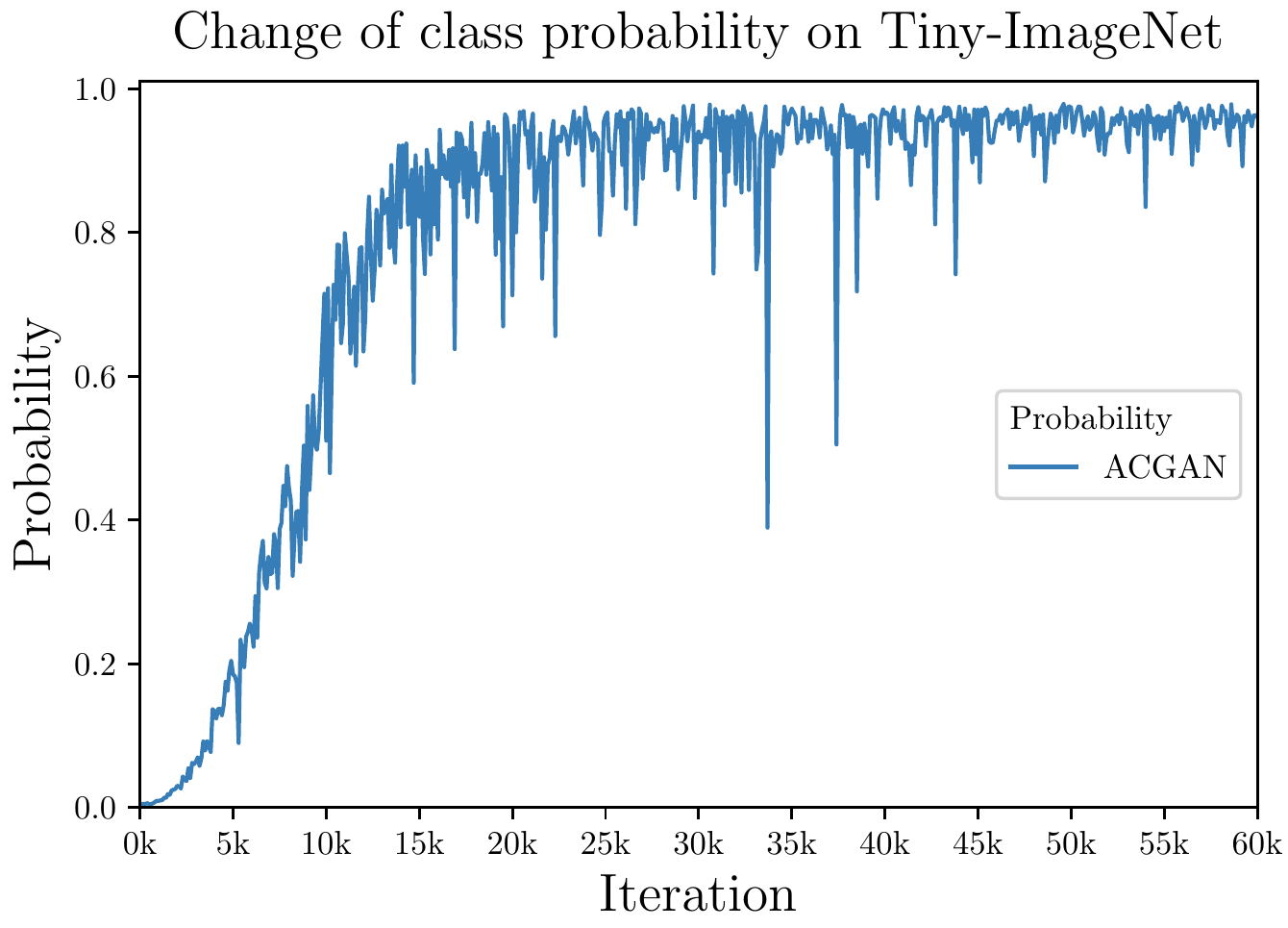}
    \end{subfigure}
    \begin{subfigure}{0.40\textwidth}
    \includegraphics[width=\linewidth]{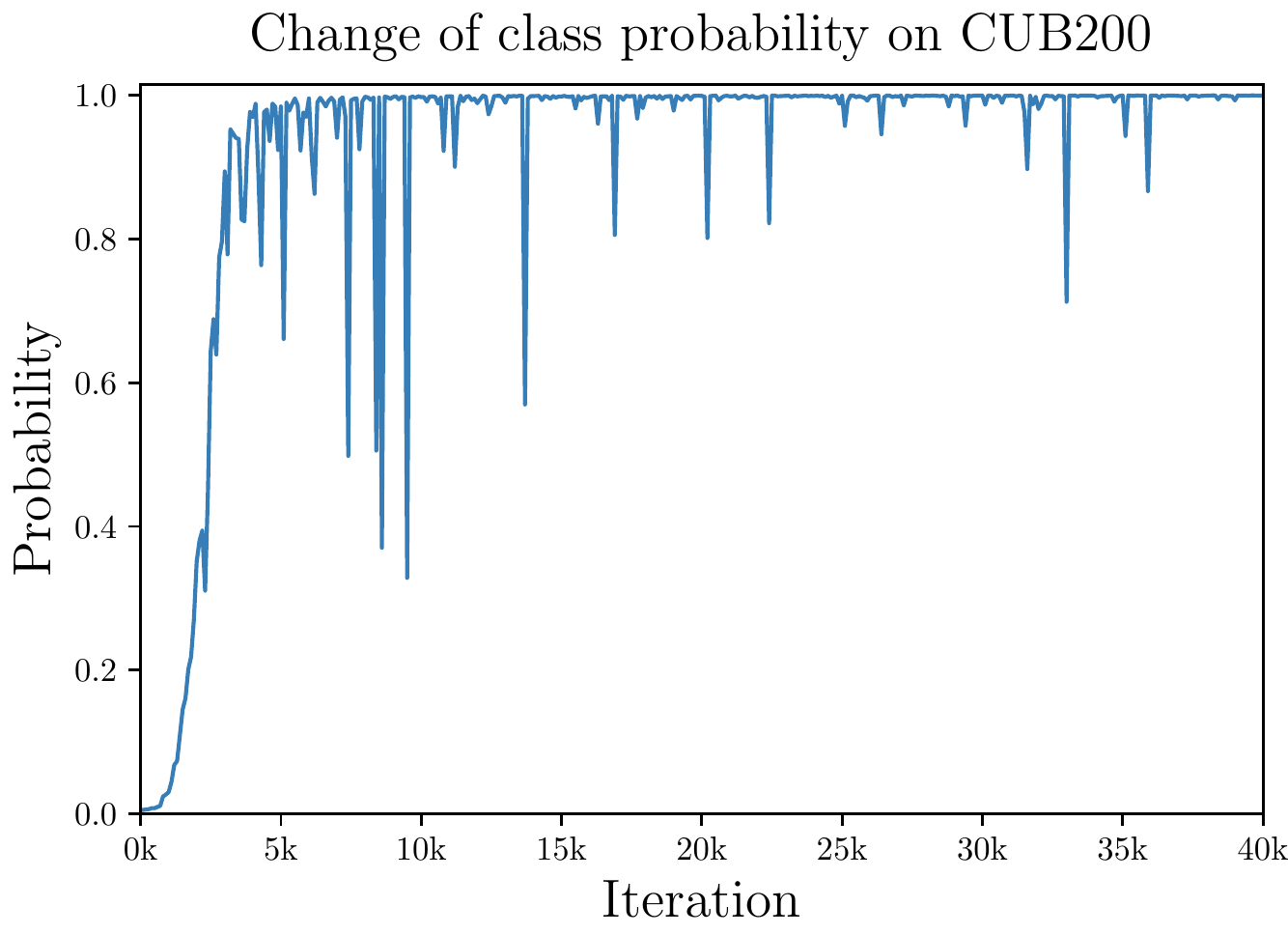}
    \end{subfigure}
    \caption{Trends of the probability values from the classifiers trained on Tiny-ImageNet~\cite{Tiny} and CUB200~\cite{WelinderEtal2010} datasets.}
    \label{fig:biased_p}
\end{figure}
\textbf{ACGAN Focuses on Classifier Training instead of Adversarial Learning.} To identify ACGAN is prone to being biased toward classifying categories of images instead of discriminating the authenticity of given samples, we track the trend of classifier's target probabilities as the training progresses using Tiny-ImageNet and CUB200 datasets. As can be seen in Fig.~\ref{fig:Figure_exploding_sp}, \ref{fig:biased_p}, and Fig.~\ref{fig:Figure_exploding} in the main paper, classifier's target probabilities continuously increase, but the FID scores do not decrease as of certain points in time. Therefore the experimental results imply that ACGAN training is likely to become biased toward label classification instead of adversarial training.

\begin{figure}[t!]
    \centering
    \includegraphics[width=0.96\linewidth]{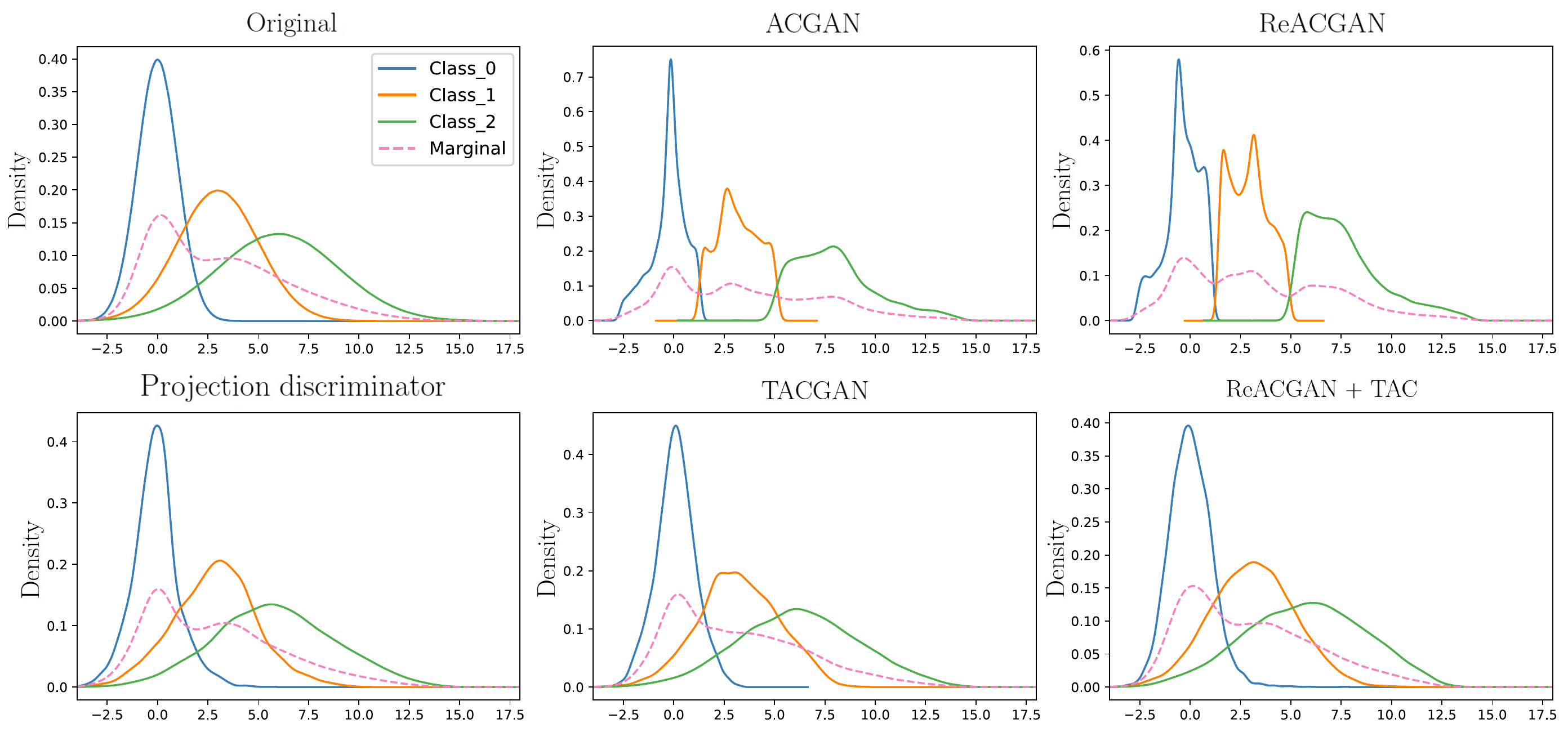}
    \caption{Comparison with ACGAN~\cite{Odena2017ConditionalIS}, Projection discriminator~\cite{Miyato2018cGANsWP}, TACGAN~\cite{NIPS2019_8414}, ReACGAN, and ReACGAN + TAC on a synthetic 1-D MoG dataset~\cite{NIPS2019_8414}. We conduct all experiments using the same setup specified in the paper~\cite{NIPS2019_8414}.}
    \label{fig:MoG-TAC}
\end{figure}

\textbf{Can ReACGAN Approximate a Mixture of Gaussian Distributions Whose Supports Overlap?} We conduct distribution approximation experiments using a 1-D mixture of Gaussian distributions~(MoG). The experiments are proposed by Gong~\etal~\cite{NIPS2019_8414} and are devised to identify if a given GAN can estimate any true data distribution, even the mixture of Gaussians with overlapped supports.  Although ReACGAN has shown successful outputs on real images, ReACGAN fails to estimate the 1-D MoG, resulting in poor approximation ability similar to ACGAN (see Fig.~\ref{fig:MoG-TAC}). However, this phenomenon is not weird because ReACGAN follows the same optimization process as ACGAN does, which inherently reduces a conditional entropy $\mathcal{H}(\rvy|\rvx)$. As the result, ACGAN can only accurately approximate marginal distributions generated by conditional distributions with non-overlapped supports.

To deal with this problem, Gong~\etal~\cite{NIPS2019_8414} have suggested using a twin auxiliary classifier~(TAC) on the top of the discriminator and have demonstrated that TAC enables ACGAN to exactly estimate the 1-D MoG. Therefore, ReACGAN can be reinforced by introducing TAC, and the experimental result verifies that ReACGAN can approximate the 1-D MoG exactly~(see Fig.~\ref{fig:MoG-TAC}).
\clearpage
\begin{table}[h!]
\centering
\caption{Experiments to identify the effectiveness of ReACGAN with TAC~\cite{NIPS2019_8414} on CIFAR10~\cite{Krizhevsky2009LearningML} and Tiny-ImageNet~\cite{Tiny} datasets.}
\setlength\tabcolsep{4.0pt}
\vspace{2mm}
  \resizebox{0.7\textwidth}{!}{
\begin{tabular}{lcccccccc}
\cmidrule[1.0pt]{1-9}
\multirow{2}*[-0.5ex]{\large{Dataset}} & \multicolumn{2}{c}{\text{ACGAN~\cite{Odena2017ConditionalIS}}} & \multicolumn{2}{c}{\text{TACGAN~\cite{NIPS2019_8414}}} & \multicolumn{2}{c}{\cellcolor{yellow!20}\text{ReACGAN}} & \multicolumn{2}{c}{\cellcolor{yellow!20}\text{ReACGAN + TAC~\cite{NIPS2019_8414}}}\\
\cmidrule[1.0pt](r){2-3}
\cmidrule[1.0pt](lr){4-5}
\cmidrule[1.0pt](lr){6-7}
\cmidrule[1.0pt](lr){8-9}
& \text{IS}~$\uparrow$ & \text{FID}~$\downarrow$ & \text{IS}~$\uparrow$ & \text{FID}~$\downarrow$ &\text{IS}~$\uparrow$ & \text{FID}~$\downarrow$ & \text{IS}~$\uparrow$ & \text{FID}~$\downarrow$ \\
\cmidrule[1.0pt]{1-9}
CIFAR10~\cite{Krizhevsky2009LearningML}  & \underline{9.84} & 8.45 & 9.78 & 8.01 & \cellcolor{yellow!20}\textbf{9.89} & \cellcolor{yellow!20}\textbf{7.88} & \cellcolor{yellow!20}9.70 & \cellcolor{yellow!20}\underline{7.94} \\
Tiny-ImageNet~\cite{Tiny} & 6.00 & 96.04 & 7.62 & 65.99 & \cellcolor{yellow!20}\textbf{14.06} & \cellcolor{yellow!20}27.10 & \cellcolor{yellow!20}13.71 & \cellcolor{yellow!20}\textbf{26.14}  \\
\cmidrule[1.0pt]{1-9}
\end{tabular}}
\label{table:tac_table}
\end{table}
Finally, to validate the effectiveness of TAC for ReACGAN on real datasets, we perform CIFAR10 and Tiny-ImageNet generation experiments. Contrary to our expectations, Table~\ref{table:tac_table} shows that ReACGAN + TAC provides comparable or marginally better results on CIFAR10 and Tiny-ImageNet datasets over the ReACGAN. We speculate that this is because benchmark datasets are highly refined and might follow a mixture of non-overlapped conditional distributions.

\textbf{Effect of D2D-CE for Different GAN Architectures.} We perform additional experiments to identify the effect of D2D-CE loss for different GAN architectures. We utilize a deep convolutional neural network~(Deep CNN)~\cite{Radford2016UnsupervisedRL} and a ResNet-style network~(ResNet)~\cite{Gulrajani2017ImprovedTO} to train GANs on CIFAR10 and Tiny-IamgeNet datasets. The experimental results show that ReACGAN provides consistent generation results on different architectures (see Table~\ref{table:sup_architecture}).
\begin{table}[h!]
\centering
\caption{Experiments for investigating the effect of D2D-CE for different architectures using CIFAR10~\cite{Krizhevsky2009LearningML} and Tiny-ImageNet~\cite{Tiny} datasets. We report only FID~\cite{Heusel2017GANsTB} for a compact expression.}
\setlength\tabcolsep{4.0pt}
\vspace{2mm}
  \resizebox{1.0\textwidth}{!}{
\begin{tabular}{lccc}
\cmidrule[1.0pt]{1-4}
Conditioning method & Deep CNN~\cite{Radford2016UnsupervisedRL} on CIFAR10~\cite{Krizhevsky2009LearningML} & ResNet~\cite{Gulrajani2017ImprovedTO} on CIFAR10~\cite{Krizhevsky2009LearningML} & ResNet~\cite{Gulrajani2017ImprovedTO} on Tiny-ImageNet~\cite{Tiny} \\
\cmidrule[1.0pt]( r){1-4}
AC~\cite{Odena2017ConditionalIS}  & 20.35 & 13.04 & 87.84 \\
PD~\cite{Miyato2018cGANsWP}  & 19.49 & 13.47 & 47.88 \\
2C~\cite{kang2020contragan}  & 21.47 & 14.38 & \textbf{40.56} \\
\rowcolor{yellow!20}D2D-CE (ReACGAN) & \textbf{18.94} & \textbf{12.47} & \underline{40.89} \\
\cmidrule[1.0pt]{1-4}
\end{tabular}}
\label{table:sup_architecture}
\end{table}


\begin{table}[t!]
\centering
\caption{Ablation study on the number of negative samples. FID score~\cite{Heusel2017GANsTB} is used for evaluation.}
\vspace{2mm}
  \resizebox{0.65\textwidth}{!}{
\begin{tabular}{lcccccc}
\cmidrule[1.0pt]{1-7}
\multirow{2}*[-0.5ex]{\large{Dataset}} & \multicolumn{6}{c}{Masking probability $p$ for negative samples in Eq.~(6)} \\
\cmidrule[1.0pt]{2-7}
 & $p=1.0$ & $0.8$ & $0.6$ & $0.4$ & $0.2$ & \cellcolor{yellow!20}$0.0$ \\
\cmidrule[1.0pt]{1-7}
\text{CIFAR10}~\cite{Krizhevsky2009LearningML} & 11.15 & 8.07 & 8.11 & 7.83 & 8.01 & \cellcolor{yellow!20}7.88 \\
\text{Tiny-ImageNet}~\cite{Tiny} & 60.74 & 30.04 & 28.64 & 28.59 & 28.68 & \cellcolor{yellow!20}27.10 \\
\cmidrule[1.0pt]{1-7}
\end{tabular}}
\label{table:negatives_ablation}
\vspace{-2.5mm}
\end{table}
\textbf{Effect of Number of Negative Samples on ReACGAN Training.} We investigate how the number of negative samples affects the generation performance of ReACGAN using CIFAR10 and Tiny-ImageNet datasets. First, we compute pairwise similarities between all negative samples in the mini-batch. Then, we drop similarities between negative samples using a randomly generated mask whose element has a value of 0 according to the pre-defined probability $p$ and otherwise has a value of 1. For example, $p= 0.1$ will lead approximately 10\% of the total similarities between negative samples not to account for calculating the denominator part of D2D-CE loss. As can be seen in Table~\ref{table:negatives_ablation}, D2D-CE loss benefits from more negative samples. This implies that the more the data-to-data relations are provided, the richer supervision signals for conditioning become, resulting in better image generation results. In addition, from the optimization point of view, the generator and discriminator can receive gradient signals at an exponential rate as the number of negative samples increases linearly.

\textbf{Are There Any Possible Prescriptions for Preventing the Gradient Exploding Problem?} In the main paper, we verify that simply normalizing feature embeddings onto a unit hypersphere resolves ACGAN's early-training collapse problem. In this section, we explore if there exist other cures for resolving the early-training collapse problem: (1) lowering classification strength, (2) gradient clipping, and (3) feature clipping. The experimental results (Table~\ref{table:exploding}) indicate that lowering classification strength and normalizing feature maps can prevent ACGAN training from collapsing at the early training phase. However, we cannot succeed in training ACGAN by clipping gradients of the classifier. We speculate that this is because gradient clipping restricts not only the norms of feature maps but also the class probability values; thus, ACGAN can be updated by inaccurate gradients and ends up collapsing. Among those methods, the normalization and D2D-CE present better performances than the others, demonstrating the effectiveness of our proposals.
\begin{table}[h!]
\centering
\caption{Experiments for studying available cures for preventing the gradient exploding problem in ACGAN. FID~\cite{Heusel2017GANsTB} scores are reported for evaluation. $\lambda$ is a balance coefficient between adversarial learning and classifier training.}
\vspace{2mm}
  \resizebox{1.0\textwidth}{!}{
\begin{tabular}{lcccccccc}
\cmidrule[1.0pt]{1-9}
Dataset & $\lambda=0.25$ & $\lambda=0.5$ & $\lambda=0.75$ & $\lambda=1$ & Normalization & Feature clipping & Gradient clipping & \cellcolor{yellow!20}D2D-CE \\
\cmidrule[1.0pt]{1-9}
\text{CIFAR100}~\cite{Krizhevsky2009LearningML}  & \underline{12.30} & 13.61 & 15.60 & 16.92 & 13.17 & 17.47 & 40.23 & \cellcolor{yellow!20}\textbf{12.25} \\
\text{Tiny-ImageNet}~\cite{Tiny}  & 62.86 & 92.05 & 104.34 & 98.75 & 28.04 & 57.65 &108.30& \cellcolor{yellow!20}\textbf{27.10} \\
\cmidrule[1.0pt]{1-9}
\end{tabular}}
\label{table:exploding}
\vspace{-2.5mm}
\end{table}

\textbf{Training Time per 100 Generator Updates.} We investigate training times of BigGAN, ContraGAN, and ReACGAN on ImageNet using 8 Nvidia V100 GPUs. The batch size is set to 2048. We identify that ReACGAN brings in a slight computational overhead and takes about 1.05$\sim$1.1 longer time than the other GANs. Specifically, BigGAN takes 17m 37s, ContraGAN 18m 24s, and ReACGAN 18m 52s per 100 generator updates.
\section{Analysis of the differences between ReACGAN and ContraGAN}
\label{difference_reac_contra}
This section explains the differences between ReACGAN and ContraGAN~\cite{kang2020contragan} from a mathematical point of view. Kang and Park~\cite{kang2020contragan} have proposed the conditional contrastive loss~(2C loss), which is formulated from NT-Xent loss~\cite{Chen2020ASF}, and developed contrastive generative adversarial networks~(ContraGAN) for conditional image generation. Using the notations used in our main paper, we can write down 2C loss as follows:
\begin{align*}
    \mathcal{L_{\text{2C}}}=-\frac{1}{N}\sum_{i=1}^{N}\log{\bigg(\frac{\exp{\big(\vf_{i}^\top\vv_{y_i}/\tau})  + \sum_{n_p \in \mathcal{P}(i)}\exp(\vf_{i}^{\top}\vf_{n_p}/\tau)}
    {\exp{\big(\vf_{i}^\top\vv_{y_i}/\tau}\big)+\sum_{j\in \{1,...,N\}\backslash\{i\} }\exp{\big(\vf_{i}^\top\vf_{j}/\tau\big)}}\bigg)},
    \label{eq:Aeq7}\tag{A7}
\end{align*}

where $\mathcal{P}(i)$ is the set of indices that indicate locations of positive samples in the mini-batch. To clearly identify how each sample embedding updates, we start by considering 2C loss on a single sample. We can rewrite a single sample version of Eq.~(\ref{eq:Aeq7}) as follows:
\begin{align*}
    \mathcal{L_{\text{2C}}^{'}}=
    &-\underbrace{\frac{1}{N}\log\bigg(\exp{\big(\vf_{q}^\top\vv_{y_q}/\tau})  + \sum_{n_p \in \mathcal{P}(q)}\exp(\vf_{q}^{\top}\vf_{n_p}/\tau)\bigg)}_{\text{Positive attraction}} \\
    & + \underbrace{\frac{1}{N}\log{\bigg(\exp{\big(\vf_{q}^\top\vv_{y_q}/\tau}\big)+\sum_{j\in \{1,...,N\}\backslash\{q\}}\exp{\big(\vf_{q}^\top\vf_{j}/\tau\big)}}\bigg)}_{\text{Negative repulsion}}.
    \label{eq:Aeq8}\tag{A8}
\end{align*}

Using Eq.~(\ref{eq:Aeq8}), we can calculate the derivative of $\mathcal{L_{\text{2C}}^{'}}$ w.r.t $\vf_q$ as follows:
\begin{align*}
    \frac{\partial \mathcal{L_{\text{2C}}^{'}}}{\partial \vf_{q}} =
    &- \frac{\exp(\vf_q^\top\vv_{y_q}/\tau)\vv_{y_q} 
    + \sum_{n_p \in \mathcal{P}(q)}\exp{(\vf_q^{\top}\vf_{n_p}/\tau)\vf_{n_p}}}
    {\tau N \bigg(\exp(\vf_q^\top\vv_{y_q}/\tau)+
    \sum_{n_p \in \mathcal{P}(q)}\exp{(\vf_q^{\top}\vf_{n_p}/\tau)}\bigg)}\\
    & + \frac{\exp(\vf_q^\top\vv_{y_q}/\tau)\vv_{y_q} + \sum_{j\in \{1,...,N\}\backslash\{q\}}\exp{(\vf_{q}^\top\vf_{j}/\tau)\vf_j}}{\tau N\bigg(\exp(\vf_q^\top\vv_{y_q}/\tau) + \sum_{j\in \{1,...,N\}\backslash\{q\}}\exp{(\vf_{q}^\top\vf_{j}/\tau)}\bigg)}.
    \label{eq:Aeq9}\tag{A9}
\end{align*}

To understand Eq.~(\ref{eq:Aeq9}) more intuitively, we replace the denominator terms of Eq.~(\ref{eq:Aeq9}) with $A$ and $B$ and re-organize the equation as follows:
\begin{align*}
    \frac{\partial \mathcal{L}_{\text{2C}}}{\partial \vf_{q}} =
    &- \frac{\exp(\vf_q^\top\vv_{y_q}/\tau)\vv_{y_q} 
    + \sum_{n_p \in \mathcal{P}(q)}\exp{(\vf_q^{\top}\vf_{n_p}/\tau)\vf_{n_p}}}
    {\tau NA}\\
    & + \frac{\exp(\vf_q^\top\vv_{y_q}/\tau)\vv_{y_q} + \sum_{j\in \{1,...,N\}\backslash\{q\}}\exp{(\vf_{q}^\top\vf_{j}/\tau)\vf_j}}{\tau NB}\\
    =& -\underbrace{\frac{1}{\tau N}\bigg(\frac{\exp(\vf_q^{\top}\vv_{y_q}/\tau)\vv_{y_q} + \overbrace{\sum_{n_p \in \mathcal{P}(q)}\exp{(\vf_q^{\top}\vf_{n_p}/\tau)\vf_{n_p}}}^{\text{(P) Positive samples}}}{A} -\frac{\exp(\vf_q^{\top}\vv_{y_q}/\tau)\vv_{y_q}}{B} \bigg)}_{\text{Positive attraction}}\\
    & +\underbrace{\frac{1}{\tau N}\bigg(\frac{\sum_{n_q \in \mathcal{N}(q)}\exp{(\vf_{q}^\top\vf_{n_q}/\tau)\vf_{n_q}} + \overbrace{\sum_{j\in \{1,...,N\}\backslash(\{q\}\cup\mathcal{N}(q))}\exp{(\vf_{q}^\top\vf_{j}/\tau)\vf_j}}^{\text{(F) False negative samples}}}{B} \bigg)}_{\text{Negative repulsion}}.
    \label{eq:Aeq10}\tag{A10}
\end{align*}
The above equation implies that the positive samples (P) in Eq~(\ref{eq:Aeq10}) can cause easy positive mining, \textit{i.e.}, if a similarity $\vf_q^{\top}\vf_{n_p}$ has a large value, the gradient $\frac{\partial\mathcal{L}_{\text{2C}}}{\partial\vf_q}$ can be biased towards $\vf_{n_p}$ direction with large magnitude. In addition, the false-negative samples (F) can attenuate the negative repulsion force, which is already being addressed in the contrastive learning community~\cite{chuang2020debiased, huynh2020boosting, robinson2020contrastive}. Unlike 2C loss, our D2D-CE loss does not experience the easy positive mining and the attenuation caused by false-negative samples. To demonstrate this, we write down D2D-CE loss as follows:

\begin{align*}
    \mathcal{L_{\text{D2D-CE}}}=-\frac{1}{N}\sum_{i=1}^{N}\log{\bigg(\frac{\exp{([\vf_{i}^\top\vv_{y_i} - m_p]_{-}/\tau})}
    {\exp{\big([\vf_{i}^\top\vv_{y_i} - m_p]_{-}/\tau}\big)+\sum_{j\in \mathcal{N}(i) }\exp{\big([\vf_{i}^\top\vf_{j} - m_n]_{+}/\tau\big)}}\bigg)}.
    \label{eq:Aeq11}\tag{A11}
\end{align*}

In the same way as before, we can expand a single sample version of Eq.~(\ref{eq:Aeq11}) as follows:
\begin{align*}
    \mathcal{L_{\text{D2D-CE}}}=
    &-\underbrace{\frac{1}{\tau N}[\vf_{q}^\top\vv_{y_q} - m_p]_{-}}_{\text{Positive attraction}} \\
    & + \underbrace{\frac{1}{N}\log{\bigg(\exp{\big([\vf_{q}^\top\vv_{y_q} - m_p]_{-}/\tau}\big)+\sum_{j\in \mathcal{N}(q)}\exp{\big([\vf_{q}^\top\vf_{j} - m_n]_{+}/\tau}}\bigg)}_{\text{Negative repulsion}}.
    \label{eq:Aeq12}\tag{A12}
\end{align*}
Based on Eq.~(\ref{eq:Aeq12}), we can calculate the derivative of $\mathcal{L_{\text{D2D-CE}}}$ w.r.t $\vf_q$ as follows:

\begin{align*}
    \frac{\partial \mathcal{L}_{\text{D2D-CE}}}{\partial \vf_{q}}=
    &-\frac{1}{\tau N}\1_{\vf_i^\top\vv_{y_i} -m_p < 0}(i=q)\vv_{y_q}  \\
    & + \frac{1}{\tau N}
    \bigg(\frac{\1_{\vf_i^\top\vv_{y_i} -m_p < 0}(i=q)\exp\big((\vf_q^\top\vv_{y_q} -m_p)/\tau\big)\vv_{y_q}}
    {\exp{\big([\vf_{q}^\top\vv_{y_q} - m_p]_{-}/\tau}\big)+\sum_{j\in \mathcal{N}(q)}\exp{\big([\vf_{q}^\top\vf_{j} - m_n]_{+}/\tau\big)}}\bigg)\\
    & + \frac{1}{\tau N}
    \bigg(\frac{\sum_{j\in\mathcal{N}(q)}\1_{\vf_{i}^\top\vf_{j} - m_n > 0}(i=q) \exp\big((\vf_{q}^\top\vf_{j} - m_n)/\tau\big)\vf_j}
    {\exp{\big([\vf_{q}^\top\vv_{y_q} - m_p]_{-}/\tau}\big)+\sum_{j\in \mathcal{N}(q)}\exp{\big([\vf_{q}^\top\vf_{j} - m_n]_{+}/\tau\big)}}\bigg)\\
    = &-\underbrace{\frac{\1_{\vf_i^\top\vv_{y_i} -m_p < 0}(i=q)}{\tau N}
    \bigg(\vv_{y_q} - 
    \frac{\exp\big((\vf_q^\top\vv_{y_q} -m_p)/\tau\big)\vv_{y_q}}
    {C} \bigg)}_{\text{Positive attraction}}\\
    & + \underbrace{\frac{1}{\tau N}\bigg(\frac{\sum_{j\in\mathcal{N}(q)}\1_{\vf_{i}^\top\vf_{j} - m_n > 0}(i=q) \exp\big((\vf_{q}^\top\vf_{j} - m_n)/\tau\big)\vf_j}
    {C}\bigg)}_{\text{Negative repulsion}},
    \label{eq:Aeq13}\tag{A13}
\end{align*}

where $C = \exp{\big([\vf_{q}^\top\vv_{y_q} - m_p]_{-}/\tau}\big)+\sum_{j\in \mathcal{N}(q)}\exp{\big([\vf_{q}^\top\vf_{j} - m_n]_{+}/\tau\big)}$. Unlike 2C loss, D2D-CE loss does not contain multiple positive samples in the positive attraction bracket and only consists of negative samples in the negative repulsion part, which means that D2D-CE loss does not perform easy-positive mining and does not attenuate the negative repulsion force. 

\begin{table}[h!]
\centering
\caption{Top-1 and Top-5 ImageNet classification accuracies on generated images from ACGAN~\cite{Odena2017ConditionalIS}, BigGAN~\cite{Brock2019LargeSG}, ContraGAN~\cite{kang2020contragan}, and ReACGAN~(ours). We use ImageNet pre-trained Inception-V3 model~\cite{Szegedy2016RethinkingTI} as a classifier. To generate images from GANs, we use the best checkpoints saved during 200k generator updates.} 
\vspace{2mm}
  \resizebox{0.95\textwidth}{!}{
\begin{tabular}{lccccc}
\cmidrule[1.0pt]{1-6}
 & Real data~(validation) & ACGAN~\cite{Odena2017ConditionalIS} & BigGAN~\cite{Brock2019LargeSG} & ContraGAN~\cite{kang2020contragan} & \cellcolor{yellow!20}ReACGAN \\
\cmidrule[1.0pt]{1-6}
Top-1 Accuracy (\%) & 70.822 & \textbf{62.412} & 29.994 & 2.866 &  \cellcolor{yellow!20}23.210\\
Top-5 Accuracy (\%) & 89.574 & 84.899 & \textbf{53.842} & 11.482 & \cellcolor{yellow!20}51.602\\
\cmidrule[1.0pt]{1-6}
IS~\cite{Salimans2016ImprovedTF}~$\uparrow$ & 173.33 & \textbf{62.99} & 28.63 & 25.25 & \cellcolor{yellow!20}50.30\\
FID~\cite{Heusel2017GANsTB}~$\downarrow$ & - & 26.35 & 24.68 & 25.16 & \cellcolor{yellow!20}\textbf{16.32} \\
\cmidrule[1.0pt]{1-6}
\end{tabular}}
\label{table:classification}
\end{table}

\begin{figure}[ht]
    \centering
    \includegraphics[width=0.99\linewidth]{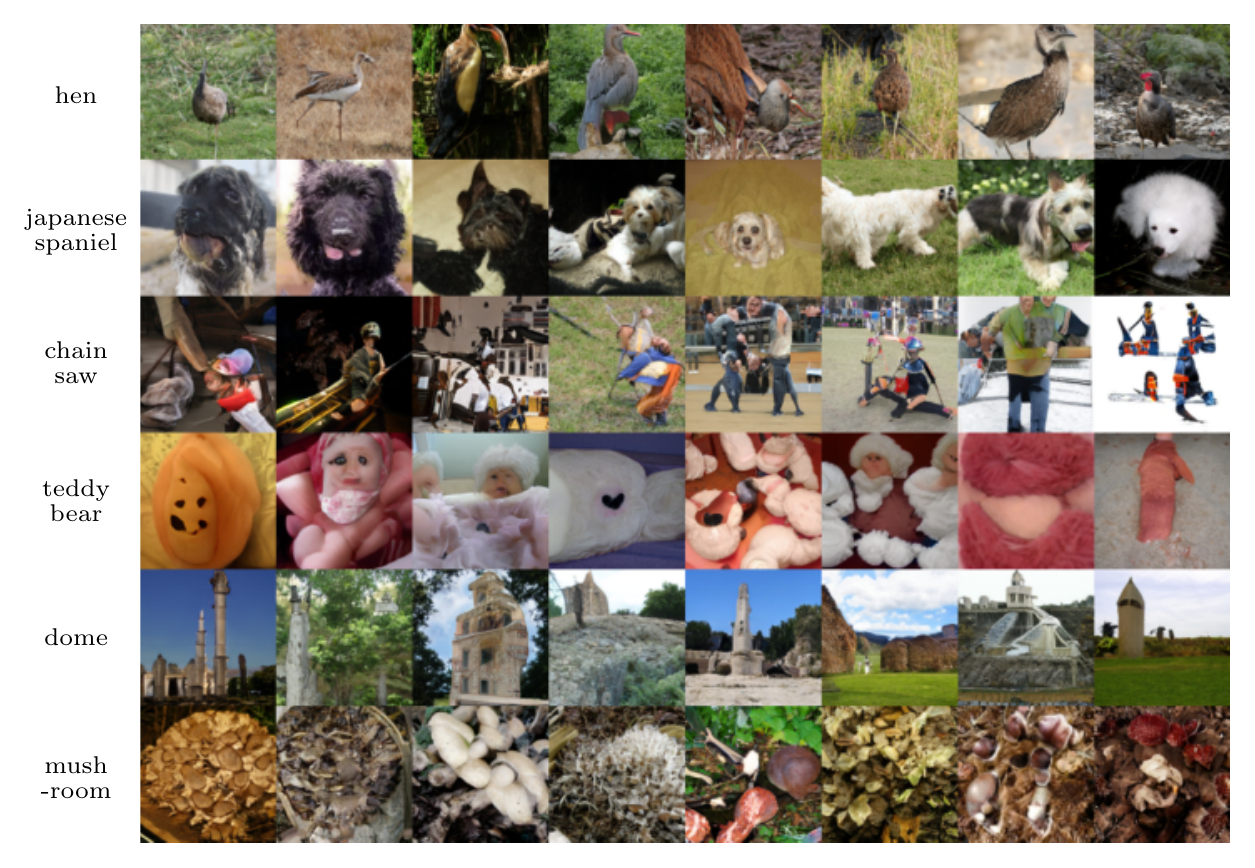}
    \hspace{-0.25cm}
    \caption{Generated images on ImageNet~\cite{Deng2009ImageNetAL} dataset using ContraGAN~\cite{kang2020contragan} and the batch size of 256~(FID=25.16).} 
    \label{fig:Figure_qualitative_contra_img}
\end{figure}

\begin{figure}[ht]
    \centering
    \includegraphics[width=0.99\linewidth]{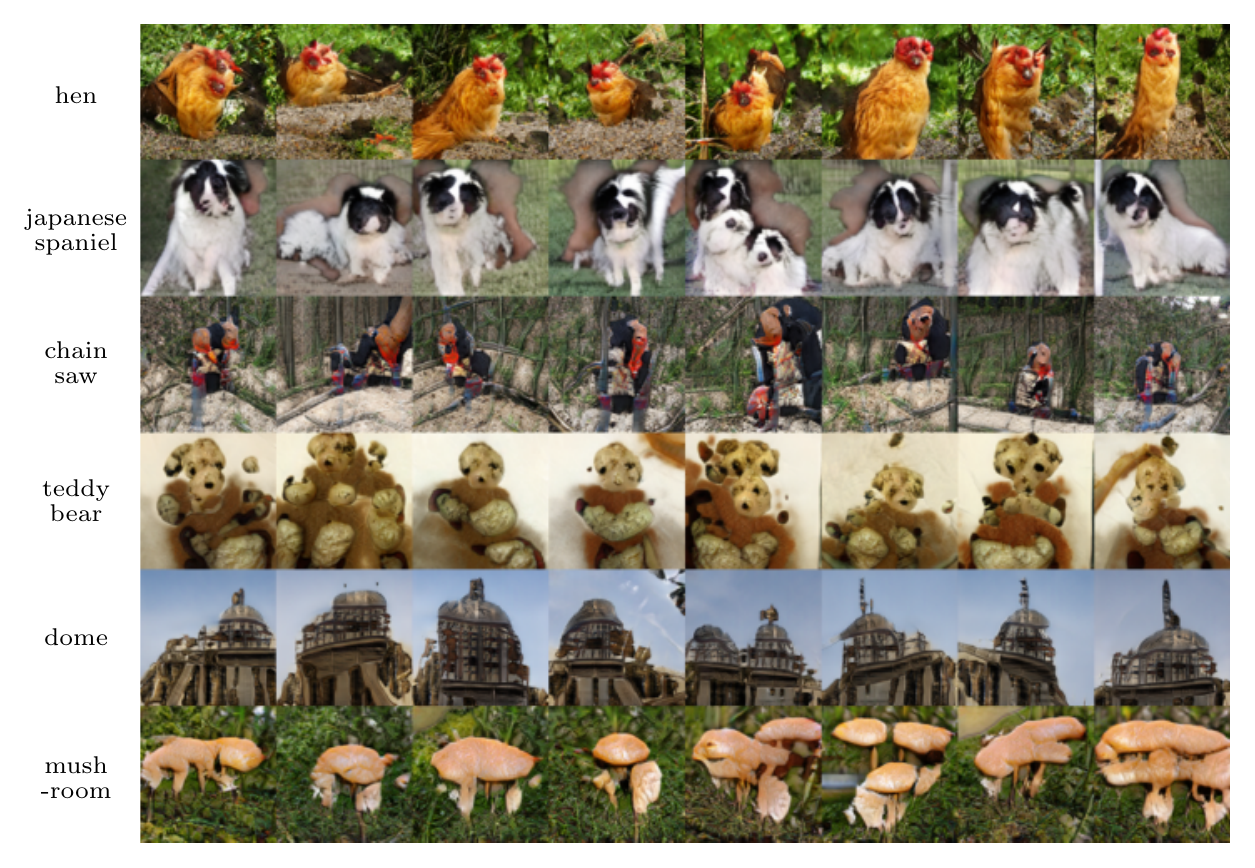}
    \hspace{-0.25cm}
    \caption{Generated images on ImageNet~\cite{Deng2009ImageNetAL} dataset using ACGAN~\cite{Odena2017ConditionalIS} and the batch size of 256 (FID=26.35).}
    \label{fig:Figure_qualitative_ac_img}
\end{figure}

\begin{figure}[ht]
    \centering
    \includegraphics[width=0.99\linewidth]{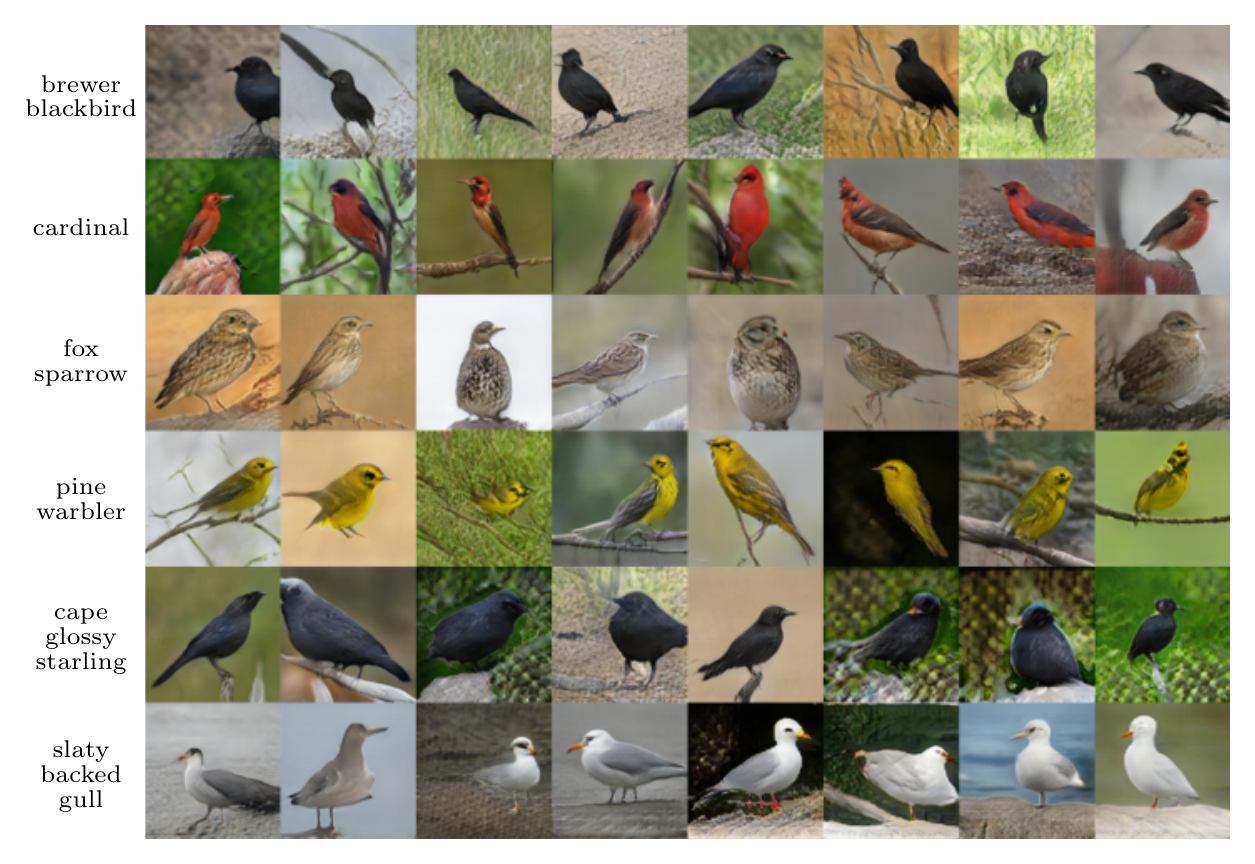}
    \hspace{-0.25cm}
    \caption{Generated images on CUB200~\cite{WelinderEtal2010} dataset using ContraGAN~\cite{kang2020contragan} (FID=20.89).}
    \label{fig:Figure_qualitative_contra_cub}
\end{figure}

\begin{figure}[ht]
    \centering
    \includegraphics[width=1.0\linewidth]{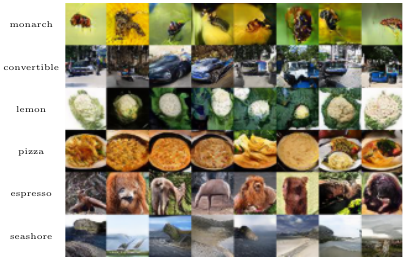}
    \hspace{-0.25cm}
    \caption{Generated images on Tiny-ImageNet~\cite{Tiny} dataset using ContraGAN~\cite{kang2020contragan} (FID=28.41).} 
    \label{fig:Figure_qualitative_contra_tiny}
\end{figure}

\begin{figure}[ht]
    \centering
    \includegraphics[width=0.90\linewidth]{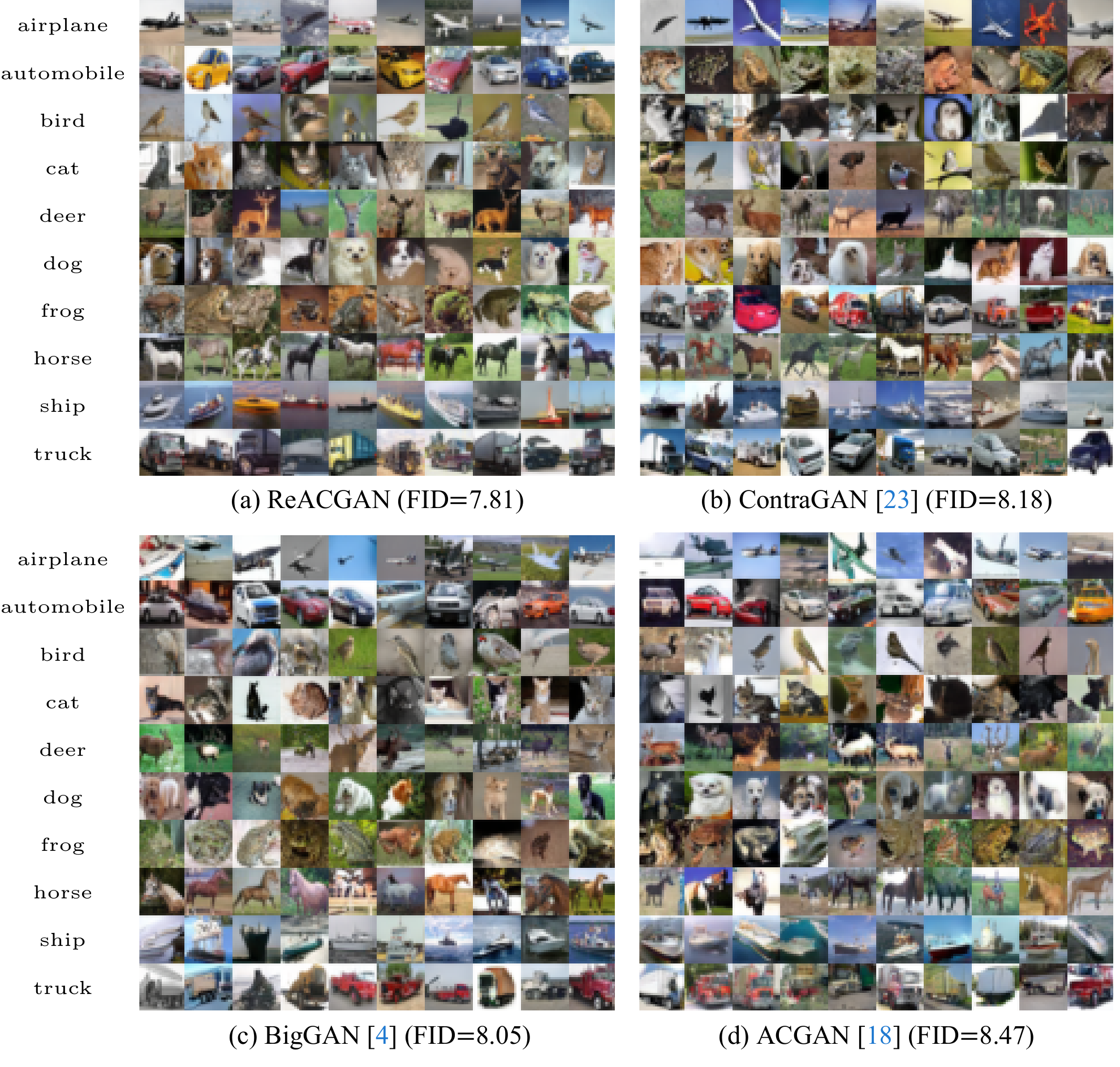}
    \caption{Generated images on CIFAR10~\cite{Krizhevsky2009LearningML} dataset.} 
    \label{fig:Figure_qualitative_cifar}
\end{figure}

To compare the conditioning performance of ReACGAN with other cGANs, we calculate Top-1 and Top-5 classification accuracies on ImageNet~\cite{Deng2009ImageNetAL} using the pre-trained Inception-V3 network~\cite{Szegedy2016RethinkingTI}. The results are summarized in Table~\ref{table:classification}. Although ReACGAN has a lower FID value and higher IS score compared with BigGAN~\cite{Brock2019LargeSG} and ContraGAN~\cite{kang2020contragan}, the top-1 and top-5 accuracies of ReACGAN are slightly below that of BigGAN. This implies that ReACGAN tends to approximate overall distribution with a slight loss of the exact conditioning. On the other hand, ContraGAN fails to perform conditional image generation, and it provides $2.866~\%$ Top-1 accuracy on ImageNet dataset. This indicates that ContraGAN is likely to generate undesirably conditioned but visually satisfactory images (see~Fig.~\ref{fig:Figure_qualitative_contra_img}, \ref{fig:Figure_qualitative_contra_cub}, \ref{fig:Figure_qualitative_contra_tiny}, and \ref{fig:Figure_qualitative_cifar} for quantitative results). One more interesting point is that although generated images from ACGAN give the best classification accuracy, they show a poor FID value compared with the others. This implies that ACGAN generates well-classifiable images without considering the diversity and fidelity of generated samples~(see~Fig.~\ref{fig:Figure_qualitative_ac_img}).
\clearpage
\section{Potential Negative Societal Impacts}
The success in generating photo-realistic images in GANs~\cite{Brock2019LargeSG, karras2019style, karras2020analyzing} has attracted a myriad of applications to be developed, such as photo editing (filtering~\cite{kupyn2018deblurgan}, stylization~\cite{yang2019controllable} and object removal~\cite{shetty2018adversarial}), image translation (sketch $\to$ clip art~\cite{chen2018sketchygan}, photo $\to$ cartoon~\cite{chen2018cartoongan}), image in-painting~\cite{yeh2017semantic}, and image extrapolation to arbitrary resolutions~\cite{zhou2020omni}.
While, in most cases, GANs are helpful for content creation or fast prototyping, there exist potential threats that one can maliciously use the synthesized results to deceive others.
A well-known example is deepfake~\cite{nguyen2019deep}, where a person in the video appears with the voice and appearance of a celebrity and conveys a message to deceive or confuse others, \textit{e.g.}, fake news.
Other examples include sexual harnesses~\cite{Xu2018FairGANFG} and hacking machine vision applications.

As an effort to circumvent the negative issues, a number of techniques have been proposed. Masi~\etal~\cite{masi2020two} have utilized color and frequency information to detect deepfake. Naseer~\etal~\cite{naseer2020self} have developed a general defense method from self-attacking via feature perturbation. We anticipate that further development of synthetic image detection techniques, well-established policies on the technique, and ethical awareness of researchers/developers will enable us to enjoy the broad applicability and benefits of GANs.
\section{Computation resources}
In this section, we provide a summary of the total number of performed experiments, computing resources, and approximated training time spent on our research in Table~\ref{table:gpu_usage}. Since we have conducted a lot of experiments with various configurations using different resources, we divide our experiments into 16 divisions and calculate \emph{approximate time spent} on each division of experiments.
\begin{table}[h]
\centering
\caption{Approximate total training time~(days) provided for reference.}
\vspace{2mm}
  \resizebox{1.0\textwidth}{!}{
\begin{tabular}{lcccc}
\cmidrule[1.0pt]{1-5}
Division of experiments & GPU Type & Days & \# of experiments & Approximate Time~(days)\\
\cmidrule[1.0pt]{1-5}
\text{CIFAR10}~\cite{Krizhevsky2009LearningML} & RTX 2080 Ti & 0.75 & 100 & 75 \\
\text{CIFAR10}~\cite{Krizhevsky2009LearningML} + CR~\cite{Zhang2019ConsistencyRF} & RTX 2080 Ti & 1.17 & 9 & 10.53  \\
\text{CIFAR10}~\cite{Krizhevsky2009LearningML} + DiffAug~\cite{zhao2020differentiable} & RTX 2080 Ti & 2.04 & 9 & 18.36  \\
\text{CIFAR10}~\cite{Krizhevsky2009LearningML} + StyleGAN2~\cite{karras2020analyzing} & TITAN Xp$\times2$ & 2.58 & 2 & 5.16  \\
\text{CIFAR10}~\cite{Krizhevsky2009LearningML} + StyleGAN2~\cite{karras2020analyzing} + ADA~\cite{Karras2020TrainingGA} & TITAN Xp$\times2$ & 9.52 & 2 & 19.04  \\
\midrule
\text{Tiny-ImageNet}~\cite{Tiny} & TITAN RTX$\times4$ & 1.54 & 84 & 132.44  \\
\text{Tiny-ImageNet}~\cite{Tiny} + CR~\cite{Zhang2019ConsistencyRF} & TITAN RTX$\times4$ & 1.42 & 9 & 12.78  \\
\text{Tiny-ImageNet}~\cite{Tiny} + DiffAug~\cite{zhao2020differentiable} & TITAN RTX$\times4$ & 2.83 & 9 & 25.47  \\
\midrule
\text{CUB200}~\cite{WelinderEtal2010} & TITAN RTX$\times4$ & 1.63 & 24 & 39.12  \\
\text{CUB200}~\cite{WelinderEtal2010} + CR~\cite{Zhang2019ConsistencyRF}& TITAN RTX$\times4$ & 0.92 & 9 & 8.28  \\
\text{CUB200}~\cite{WelinderEtal2010} + DiffAug~\cite{zhao2020differentiable}& TITAN RTX$\times4$ & 0.67 & 9 & 6.03  \\
\midrule
\text{ImageNet}~\cite{Deng2009ImageNetAL} (200k iter., B.S.=256) & Tesla V100$\times4$ & 4.17 & 6 & 25.02  \\
\text{ImageNet}~\cite{Deng2009ImageNetAL} (600k iter., B.S.=256) & Tesla V100$\times4$ & 12.51 & 2 & 25.02  \\
\text{ImageNet}~\cite{Deng2009ImageNetAL} + DiffAug~\cite{zhao2020differentiable} (600k iter., B.S.=256) & A100$\times4$ & 12.43 & 2 & 24.86  \\
\text{ImageNet}~\cite{Deng2009ImageNetAL} (B.S.=2048) & Tesla V100$\times8$ & 26.90 & 2 & 53.80  \\
\midrule
\text{AFHQ}~\cite{choi2020starganv2} + StyleGAN2~\cite{karras2020analyzing} + ADA~\cite{Karras2020TrainingGA} & A100$\times4$ & 1.96 & 2 & 3.92  \\
\cmidrule[1.0pt]{1-5}
Total &  &  & 280 & 484.83  \\
\cmidrule[1.0pt]{1-5}
\end{tabular}}
\label{table:gpu_usage}
\end{table}
\section{Standard Deviations of Experiments}
We run all the experiments three times with random seeds and report the averaged best performances for reliable evaluation with the lone exception of ImageNet experiments. This section provides standard deviations for reference.
\begin{table}[ht]
\caption{Comparisons with classifier-based GANs~\cite{Odena2017ConditionalIS, kang2020contragan} and projection-based GANs~\cite{Miyato2018SpectralNF,Brock2019LargeSG,Zhang2019SelfAttentionGA} on CIFAR10~\cite{Krizhevsky2009LearningML}, Tiny-ImageNet~\cite{Tiny}, and CUB200~\cite{WelinderEtal2010} datasets using IS~\cite{Salimans2016ImprovedTF}, FID~\cite{Heusel2017GANsTB}, $\text{F}_{\text{0.125}}$ and $\text{F}_{\text{8}}$~\cite{sajjadi2018assessing} metrics. We report the standard deviations of three different runs in this table.}
\setlength\tabcolsep{4.0pt}
\vspace{2mm}
  \resizebox{1.0\textwidth}{!}{
\begin{tabular}{lcccccccccccc}
\cmidrule[1.0pt]{1-13}
\multirow{2}*[-0.5ex]{\large{Method}} & \multicolumn{4}{c}{\text{CIFAR10~\cite{Krizhevsky2009LearningML}}} & \multicolumn{4}{c}{\text{Tiny-ImageNet~\cite{Tiny}}} & \multicolumn{4}{c}{\text{CUB200~\cite{WelinderEtal2010}}} \\
\cmidrule[1.0pt]( r){2-5}
\cmidrule[1.0pt](lr){6-9}
\cmidrule[1.0pt](l ){10-13}
& \text{IS}~$\uparrow$ & \text{FID}~$\downarrow$ & $\text{F}_{\text{0.125}}$~$\uparrow$ & $\text{F}_{\text{8}}$ ~$\uparrow$ & \text{IS}~$\uparrow$ & \text{FID}~$\downarrow$ & $\text{F}_{\text{0.125}}$~$\uparrow$ & $\text{F}_{\text{8}}$~$\uparrow$ & \text{IS}~$\uparrow$ & \text{FID}~$\downarrow$ & $\text{F}_{\text{0.125}}$~$\uparrow$ & $\text{F}_{\text{8}}$ ~$\uparrow$ \\ 
\cmidrule[1.0pt]( r){1-5}
\cmidrule[1.0pt](lr){6-9}
\cmidrule[1.0pt](l ){10-13}
SNGAN$^{*}$~\cite{Miyato2018SpectralNF} & - & - & - & - & - & - & - & - & - & - & - & -\\
BigGAN$^{*}$~\cite{Brock2019LargeSG} & - & - & - & - & - & - & - & - & - & - & - & -\\
ContraGAN$^{*}$~\cite{kang2020contragan} & - & - & - & - & - & - & - & - & - & - & - & -\\
ACGAN~\cite{Odena2017ConditionalIS} & 0.06 & \textbf{0.02} & 0.001 & \textbf{0.000} & 0.12 & 8.04 & 0.021 & 0.048 & 0.73 & 9.20 & 0.036 & 0.102 \\
SNGAN~\cite{Miyato2018SpectralNF} & \textbf{0.02} & 0.20 & 0.002 & 0.004 & 0.16 & 0.84 & 0.009 & 0.022 & 0.35 & 2.08 & 0.020 & 0.010 \\
SAGAN~\cite{Zhang2019SelfAttentionGA} & 0.06 & 0.45 & 0.003 & 0.004 & 0.52 & 6.97 & 0.056 & 0.034 & 0.20 & 11.62 & 0.070 & 0.039 \\
BigGAN~\cite{Brock2019LargeSG} & 0.08 & 0.04 & 0.001 & 0.004 & 2.35 & 4.48 & 0.014 & 0.068 & 0.11 & 2.83 & 0.020 & \textbf{0.008}\\
ContraGAN~\cite{kang2020contragan} & 0.07 & 0.09 & \textbf{0.000} & 0.001 & 0.34 & 0.57 & \textbf{0.001} & 0.011 & 0.13 & 1.36 & 0.008 & 0.021 \\
\rowcolor{yellow!20}ReACGAN & 0.10 & 0.07 & 0.001 & 0.002 & \textbf{0.07} & \textbf{0.56} & 0.003 & \textbf{0.009} & \textbf{0.09} & \textbf{0.80} & \textbf{0.007} & 0.009\\
\midrule
BigGAN + CR$^{*}$~\cite{Zhang2019ConsistencyRF} & - & - & - & - & - & - & - & - & - & - & - & -\\
BigGAN~\cite{Brock2019LargeSG} + CR~\cite{Zhang2019ConsistencyRF} & \textbf{0.15} & 0.09 & \textbf{0.001} & \textbf{0.001} & 0.47 & \textbf{0.15} & 0.003 & 0.002 & \textbf{0.00} & 0.85 & 0.004 & \textbf{0.003} \\
ContraGAN~\cite{kang2020contragan} + CR~\cite{Zhang2019ConsistencyRF} & 0.28 & 0.94 & 0.003 & 0.015 & 0.41 & 0.46 & \textbf{0.001} & \textbf{0.001} & 0.07 & \textbf{0.15} & 0.004 & \textbf{0.003} \\
\rowcolor{yellow!20}ReACGAN + CR~\cite{Zhang2019ConsistencyRF} & 0.16 & \textbf{0.02} & \textbf{0.001} & \textbf{0.001} & \textbf{0.16} & 0.82 & \textbf{0.001} & 0.005  & 0.07 & 0.45 & \textbf{0.003} & \textbf{0.003} \\
\midrule
BigGAN + DiffAug$^{*}$~\cite{zhao2020differentiable} & - & - & - & - & - & - & - & - & - & - & - & -\\
BigGAN~\cite{Brock2019LargeSG} + DiffAug~\cite{zhao2020differentiable} & 0.16 & \textbf{0.06} & 0.001 & \textbf{0.000} & 0.87 & 0.81 & \textbf{0.001} & 0.002 & 0.18 & 1.10 & 0.004 & \textbf{0.001} \\
ContraGAN~\cite{kang2020contragan} + DiffAug~\cite{zhao2020differentiable} & 0.07 & 0.08 & 0.001 & \textbf{0.000} & 0.78 & 0.31 & 0.002 & \textbf{0.001} & 0.07 & \textbf{0.15} & 0.004 & 0.003 \\ 
\rowcolor{yellow!20}ReACGAN + DiffAug~\cite{zhao2020differentiable} & \textbf{0.04} & 0.12 & \textbf{0.000} & 0.002 & \textbf{0.12} & \textbf{0.14} & 0.002 & \textbf{0.001} & \textbf{0.05} & 0.66 & \textbf{0.002} & \textbf{0.001}\\
\cmidrule[1.0pt]{1-13}
\end{tabular}}
\end{table}

\begin{table}[ht]
\centering
\caption{Experiments on the effectiveness of D2D-CE loss compared with other conditioning losses.}
\setlength\tabcolsep{4.0pt}
\vspace{2mm}
  \resizebox{1.0\textwidth}{!}{
\begin{tabular}{lcccccccccccc}
\cmidrule[1.0pt]{1-13}
\multirow{2}*[-0.5ex]{\normalsize{Conditioning Method}} & \multicolumn{4}{c}{\text{CIFAR10~\cite{Krizhevsky2009LearningML}}} & \multicolumn{4}{c}{\text{Tiny-ImageNet~\cite{Tiny}}} & \multicolumn{4}{c}{\text{CUB200~\cite{WelinderEtal2010}}} \\
\cmidrule[1.0pt]( r){2-5}
\cmidrule[1.0pt](lr){6-9}
\cmidrule[1.0pt](l ){10-13}
& \textsc{IS}~$\uparrow$ & \textsc{FID}~$\downarrow$ & $\textsc{F}_{\text{0.125}}$~$\uparrow$ & $\textsc{F}_{\text{8}}$ ~$\uparrow$ &\textsc{IS}~$\uparrow$ & \textsc{FID}~$\downarrow$ & $\textsc{F}_{\text{0.125}}$~$\uparrow$ & $\textsc{F}_{\text{8}}$~$\uparrow$ & \textsc{IS}~$\uparrow$ & \textsc{FID}~$\downarrow$ & $\textsc{F}_{\text{0.125}}$~$\uparrow$ & $\textsc{F}_{\text{8}}$ ~$\uparrow$ \\
\cmidrule[1.0pt]( r){1-5}
\cmidrule[1.0pt](lr){6-9}
\cmidrule[1.0pt](l ){10-13}
\multirow{2}*[1.0ex]{BigGAN w/o Condition~\cite{Brock2019LargeSG}} & \multirow{2}*[-0.5ex]{0.12} & \multirow{2}*[-0.5ex]{0.20} & \multirow{2}*[-0.5ex]{0.001} & \multirow{2}*[-0.5ex]{0.001}& \multirow{2}*[-0.5ex]{0.79} & \multirow{2}*[-0.5ex]{11.69} & \multirow{2}*[-0.5ex]{0.055} & \multirow{2}*[-0.5ex]{0.097}& \multirow{2}*[-0.5ex]{0.23} & \multirow{2}*[-0.5ex]{7.00} & \multirow{2}*[-0.5ex]{0.031} & \multirow{2}*[-0.5ex]{0.015}\\
\multirow{2}*[1.0ex]{(Abbreviated to Big)} &  &  &  &  & & &  &  &  &  &  & \\
\midrule
Big + AC~\cite{Odena2017ConditionalIS} & \textbf{0.06} & \textbf{0.02} & 0.001 & \textbf{0.000} & 0.12 & 8.04 & 0.021 & 0.048 & 0.73 & 9.20 & 0.036 & 0.102 \\
Big + PD~\cite{Miyato2018cGANsWP} & 0.08 & 0.04 & 0.001 & 0.004 & 2.35 & 4.48 & 0.014 & 0.068 & 0.11 & 2.83 & 0.020 & \textbf{0.008}\\
Big + MH~\cite{kavalerov2021multi} & 0.12 & 0.04 & \textbf{0.000} & 0.002 & 0.72 & 15.42 & 0.102 & 0.056 & 0.42 & 40.23 & 0.124 & 0.087\\
Big + 2C~\cite{kang2020contragan} & 0.07 & 0.09 & \textbf{0.000} & 0.001 & 0.34 & 0.57 & \textbf{0.001} & 0.011 & 0.13 & 1.36 & 0.008 & 0.021 \\
\rowcolor{yellow!20}Big + D2D-CE~(ReACGAN) & 0.10 & 0.07 & 0.001 & 0.002 & \textbf{0.07} & \textbf{0.56} & 0.003 & \textbf{0.009} & \textbf{0.09} & \textbf{0.80} & \textbf{0.007} & 0.009\\ \cmidrule[1.0pt]{1-13}
\end{tabular}}
\end{table}
\begin{table}[ht]
\centering
\caption{Experiments to identify the consistent performance of D2D-CE on adversarial loss selection.}
\vspace{2mm}
  \resizebox{0.93\textwidth}{!}{
\begin{tabular}{lccccccccccc}
\cmidrule[1.0pt]{1-10}
\multirow{2}*[-0.5ex]{\normalsize{Adversarial Loss}} & \multirow{2}*[0.5ex]{Conditioning} & \multicolumn{4}{c}{\text{CIFAR10~\cite{Krizhevsky2009LearningML}}} & \multicolumn{4}{c}{\text{Tiny-ImageNet~\cite{Tiny}}} \\
\cmidrule[1.0pt]( r){3-6}
\cmidrule[1.0pt]( l){7-10}
& \multirow{2}*[1.5ex]{Method} & \text{IS}~$\uparrow$ & \text{FID}~$\downarrow$ & $\text{F}_{\text{0.125}}$~$\uparrow$ & $\text{F}_{\text{8}}$ ~$\uparrow$ & \text{IS}~$\uparrow$ & \text{FID}~$\downarrow$ & $\text{F}_{\text{0.125}}$~$\uparrow$ & $\text{F}_{\text{8}}$ ~$\uparrow$ \\
\cmidrule[1.0pt]{1-10}
\multirow{3}*[0.5ex]{Non-Saturation~\cite{Goodfellow2014GenerativeAN}} & PD~\cite{Miyato2018cGANsWP} & 0.10 & 0.12 & \textbf{0.001} & 0.002 & \textbf{0.41} & \textbf{1.56} & 0.029 & 0.032\\
& 2C~\cite{kang2020contragan} & \textbf{0.07} & 0.08 & \textbf{0.001} & 0.006 & 0.98 & 6.72 & 0.020 & 0.034\\
 & \cellcolor{yellow!20}D2D-CE & \cellcolor{yellow!20}\textbf{0.07} & \cellcolor{yellow!20}\textbf{0.06} & \cellcolor{yellow!20}\textbf{0.001} & \cellcolor{yellow!20}\textbf{0.001} & \cellcolor{yellow!20}0.67 & \cellcolor{yellow!20}3.09 & \cellcolor{yellow!20}\textbf{0.011} & \cellcolor{yellow!20}\textbf{0.028} \\
\cmidrule[0.4pt]{1-10}
\multirow{3}*[0.5ex]{Least Square~\cite{Mao2017LeastSG}} & PD~\cite{Miyato2018cGANsWP} & 0.08 & 0.26 & \textbf{0.001} & \textbf{0.002} & \textbf{0.49} & \textbf{0.72} & \textbf{0.008} & \textbf{0.006} \\
 & 2C~\cite{kang2020contragan} & 0.29 & 1.28 & 0.002 & 0.012 & 3.01 & 14.32 & 0.082 & 0.123\\
& \cellcolor{yellow!20}D2D-CE & \cellcolor{yellow!20}\textbf{0.05} & \cellcolor{yellow!20}\textbf{0.23} & \cellcolor{yellow!20}0.002 & \cellcolor{yellow!20}\textbf{0.002} & \cellcolor{yellow!20}2.56 & \cellcolor{yellow!20}16.55 & \cellcolor{yellow!20}0.105 & \cellcolor{yellow!20}0.172 \\
\cmidrule[0.4pt]{1-10}
\multirow{3}*[0.5ex]{W-GP~\cite{Gulrajani2017ImprovedTO}} & PD~\cite{Miyato2018cGANsWP} & 0.30 & \textbf{2.82} & \textbf{0.010} & \textbf{0.024} & \textbf{2.05} & 37.22 & 0.261 & 0.234 \\
& 2C~\cite{kang2020contragan} & \textbf{0.27} & 7.35 & 0.056 & 0.031 & 2.20 & 33.35 & 0.187 & 0.190\\
 & \cellcolor{yellow!20}D2D-CE & \cellcolor{yellow!20}0.41 & \cellcolor{yellow!20}6.75 & \cellcolor{yellow!20}0.015 & \cellcolor{yellow!20}0.043 & \cellcolor{yellow!20}2.43 & \cellcolor{yellow!20}\textbf{20.38} & \cellcolor{yellow!20}\textbf{0.086} & \cellcolor{yellow!20}\textbf{0.155} \\
\cmidrule[0.4pt]{1-10}
\multirow{3}*[0.5ex]{Hinge~\cite{Lim2017GeometricG}} & PD~\cite{Miyato2018cGANsWP} & 0.08 & \textbf{0.04} & 0.001 & 0.004 & 2.35 & 4.48 & 0.014 & 0.068\\
& 2C~\cite{kang2020contragan} & \textbf{0.07} & 0.09 & \textbf{0.000} & \textbf{0.001} & 0.34 & 0.57 & \textbf{0.001} & 0.011\\
& \cellcolor{yellow!20}D2D-CE & \cellcolor{yellow!20}0.10 & \cellcolor{yellow!20}0.07 & \cellcolor{yellow!20}0.001 & \cellcolor{yellow!20}0.002 & \cellcolor{yellow!20}\textbf{0.07} & \cellcolor{yellow!20}\textbf{0.56} & \cellcolor{yellow!20}0.003 & \cellcolor{yellow!20}\textbf{0.009}\\
\cmidrule[1.0pt]{1-10}
\end{tabular}}
\end{table}

\clearpage
\begin{table}[ht]
\centering
\caption{Ablation study on normalization, data-to-data consideration, and easy negative and positive sample suppression.}
\setlength\tabcolsep{4.0pt}
\vspace{2mm}
  \resizebox{0.70\textwidth}{!}{
\begin{tabular}{lcccccccc}
\cmidrule[1.0pt]{1-9}
\multirow{2}*[-0.5ex]{\large{Ablation}} & \multicolumn{4}{c}{\text{Tiny-ImageNet~\cite{Tiny}}} & \multicolumn{4}{c}{\text{CUB200~\cite{WelinderEtal2010}}} \\
\cmidrule[1.0pt]( r){2-5}
\cmidrule[1.0pt](lr){6-9}
& \text{IS}~$\uparrow$ & \text{FID}~$\downarrow$ & $\text{F}_{\text{0.125}}$~$\uparrow$ & $\text{F}_{\text{8}}$ ~$\uparrow$ &\text{IS}~$\uparrow$ & \text{FID}~$\downarrow$ & $\text{F}_{\text{0.125}}$~$\uparrow$ & $\text{F}_{\text{8}}$~$\uparrow$ \\
\cmidrule[1.0pt]( r){1-5}
\cmidrule[1.0pt](lr){6-9}
ACGAN~\cite{Odena2017ConditionalIS} & 0.12 & 8.04 & 0.021 & 0.048 & 0.73 & 9.20 & 0.036 & 0.102\\
+ Normalization  & - & - & - & - & - & - & - & - \\
+ Data-to-data~(Eq.~(\ref{eq:eq5}))  & - & - & - & - & - & - & - & - \\
\rowcolor{yellow!20}+ Suppression~(Eq.~(\ref{eq:eq6}))  & 0.07 & 0.56 & 0.003 & 0.009 & 0.09 & 0.80 & 0.007 & 0.009\\ \cmidrule[0.4pt]{1-9}
- Data-to-data & - & - & - & - & - & - & - & -\\
\cmidrule[1.0pt]{1-9}
\end{tabular}}
\end{table}
\begin{table}[ht]
\centering
\caption{Experiments for investigating the effect of D2D-CE for different architectures using CIFAR10~\cite{Krizhevsky2009LearningML} and Tiny-ImageNet~\cite{Tiny} datasets.}
\setlength\tabcolsep{4.0pt}
\vspace{2mm}
  \resizebox{0.95\textwidth}{!}{
\begin{tabular}{lccc}
\cmidrule[1.0pt]{1-4}
Conditioning method & Deep CNN~\cite{Radford2016UnsupervisedRL} on CIFAR10~\cite{Krizhevsky2009LearningML} & ResNet~\cite{Gulrajani2017ImprovedTO} on CIFAR10~\cite{Krizhevsky2009LearningML} & ResNet~\cite{Gulrajani2017ImprovedTO} on Tiny-ImageNet~\cite{Tiny} \\
\cmidrule[1.0pt]( r){1-4}
AC~\cite{Odena2017ConditionalIS}  & 0.02 & 0.11 & 0.34 \\
PD~\cite{Miyato2018cGANsWP}  & 0.67 & 0.25 & 1.29 \\
2C~\cite{kang2020contragan}  & 1.28 & 1.33 & 0.18 \\
\rowcolor{yellow!20}D2D-CE~(ReACGAN) & 0.03 & 0.23 & 0.29 \\
\cmidrule[1.0pt]{1-4}
\end{tabular}}
\end{table}
\begin{table}[ht]
\centering
\caption{Ablation study on the number of negative samples.}
\vspace{2mm}
  \resizebox{0.65\textwidth}{!}{
\begin{tabular}{lcccccc}
\cmidrule[1.0pt]{1-7}
\multirow{2}*[-0.5ex]{\large{Dataset}} & \multicolumn{6}{c}{Masking probability $p$ for negative samples in Eq.~(\ref{eq:eq6})} \\
\cmidrule[1.0pt]{2-7}
 & $p=1.0$ & $0.8$ & $0.6$ & $0.4$ & $0.2$ & \cellcolor{yellow!20}$0.0$ \\
\cmidrule[1.0pt]{1-7}
\text{CIFAR10}~\cite{Krizhevsky2009LearningML} & 0.33 & 0.04 & 0.11 & 0.13 & 0.05 & \cellcolor{yellow!20}0.07 \\
\text{Tiny-ImageNet}~\cite{Tiny} & 8.62 & 1.82 & 0.76 & 0.92 & 0.87 & \cellcolor{yellow!20}0.56 \\
\cmidrule[1.0pt]{1-7}
\end{tabular}}
\label{table:negative_ablation}
\end{table}

\section{Qualitative Results}
\label{qualitative_all}
We provide images that are generated by our ReACGAN and baseline approaches (ContraGAN~\cite{kang2020contragan}, BigGAN~\cite{Brock2019LargeSG}, and ACGAN~\cite{Odena2017ConditionalIS}).

\begin{figure}[h!]
    \centering
    \includegraphics[width=0.98\linewidth]{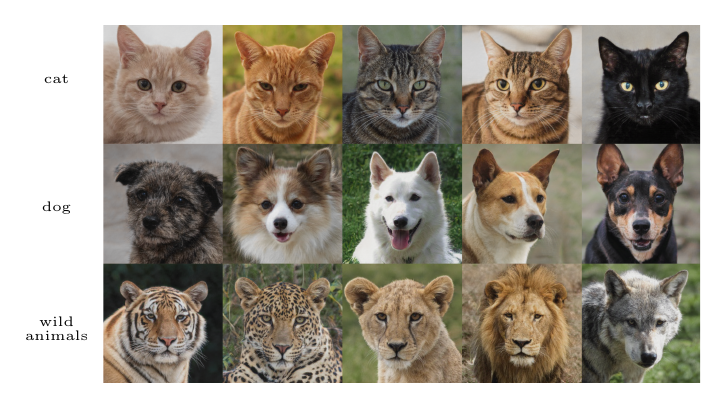}
    \hspace{-0.25cm}
    \caption{Generated images on AFHQ~\cite{choi2020starganv2} dataset using StyleGAN2~\cite{karras2020analyzing} +  ADA~\cite{Karras2020TrainingGA} + D2D-CE~(ReACGAN)~(FID=4.95).}
    \label{fig:Figure_qualitative_big_afhq}
\end{figure}
\begin{figure}[ht]
    \centering
    \includegraphics[width=0.98\linewidth]{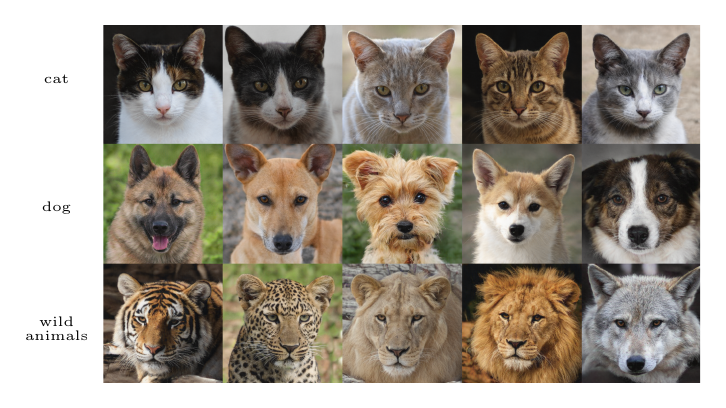}
    \hspace{-0.25cm}
    \caption{Generated images on AFHQ~\cite{choi2020starganv2} dataset using StyleGAN2~\cite{karras2020analyzing} + ADA~\cite{Karras2020TrainingGA} (FID=4.99).} 
    \label{fig:Figure_qualitative_reac_afhq}
\end{figure}
 \begin{figure}[ht]
    \centering
    \includegraphics[width=0.99\linewidth]{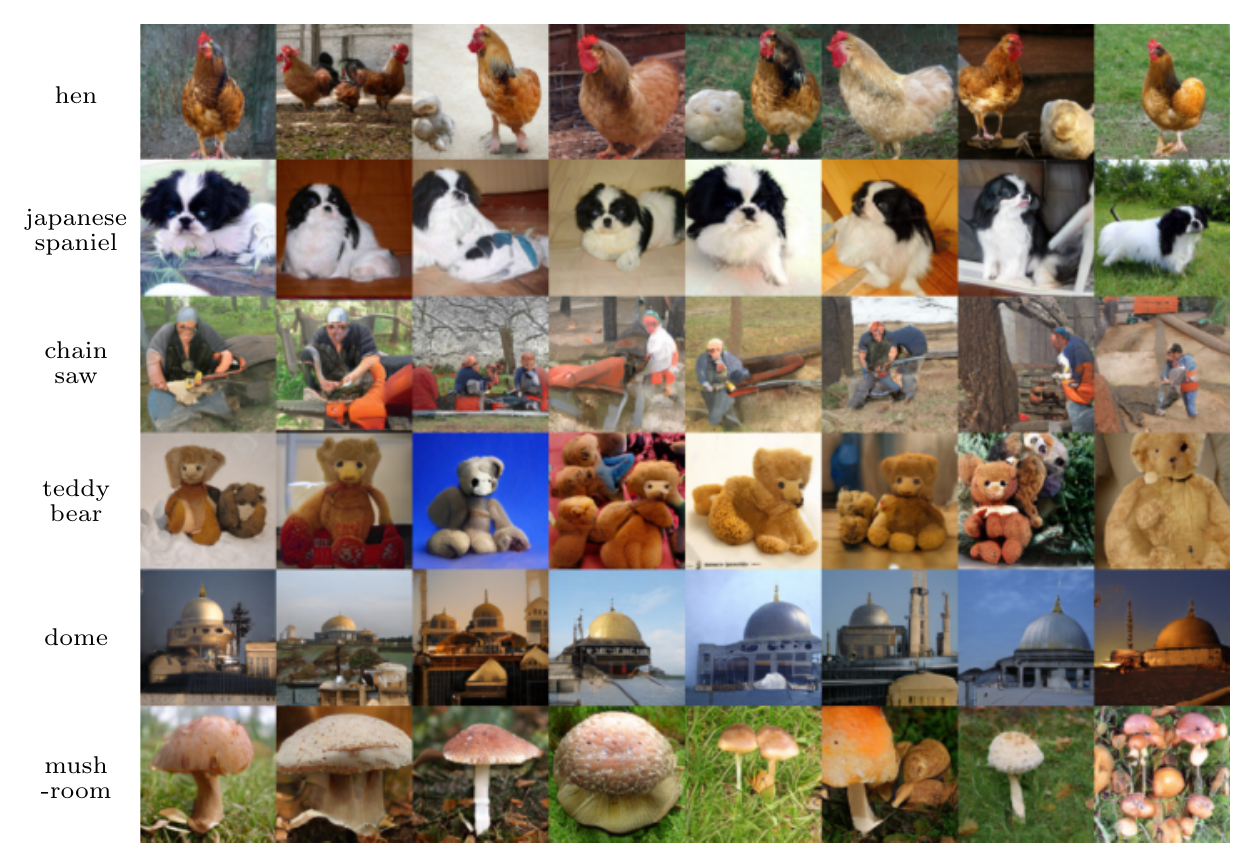}
    \hspace{-0.25cm}
    \caption{Generated images on ImageNet~\cite{Deng2009ImageNetAL} dataset using ReACGAN and the batch size of 2048~(FID=8.23).} 
    \label{fig:Figure_qualitative_reac_img2048}
\end{figure}
\begin{figure}[ht]
    \centering
    \includegraphics[width=0.99\linewidth]{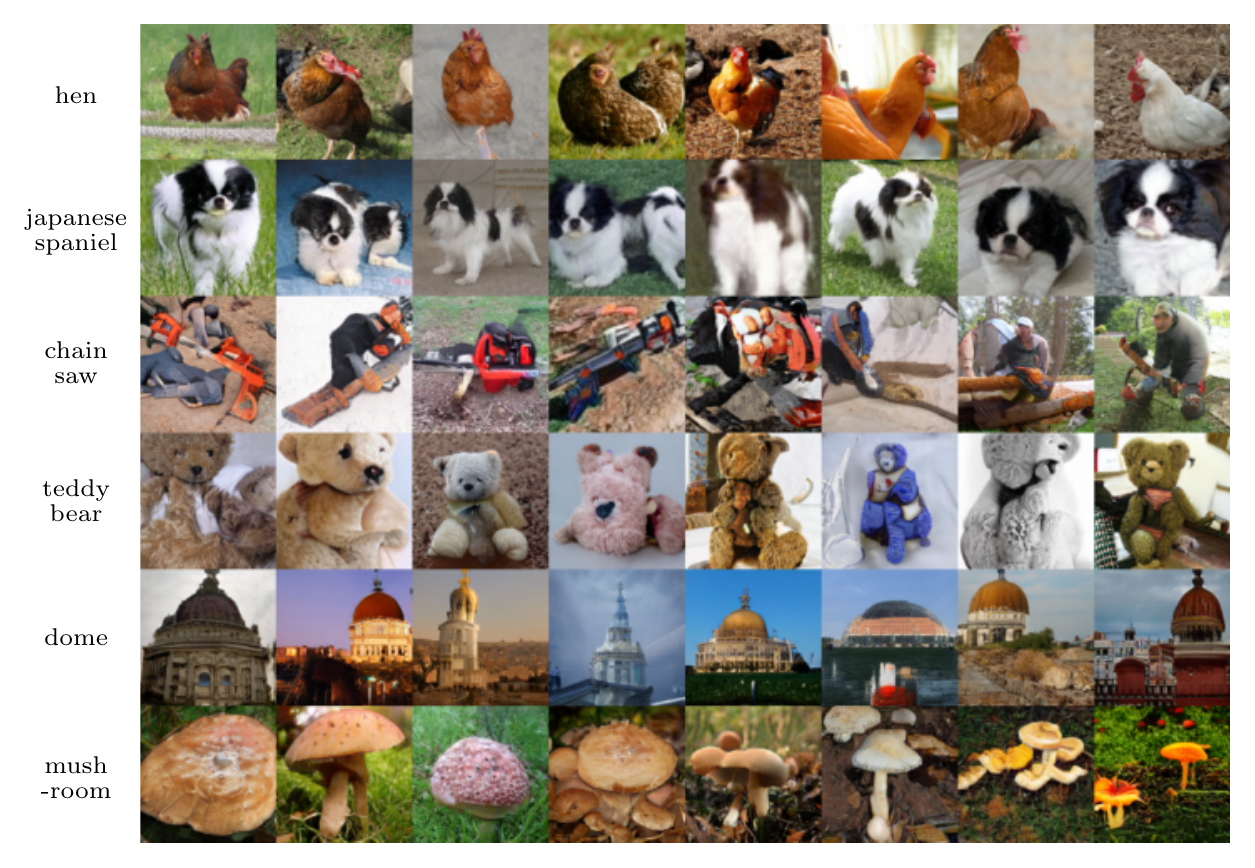}
    \hspace{-0.25cm}
    \caption{Generated images on ImageNet~\cite{Deng2009ImageNetAL} dataset using BigGAN~\cite{Brock2019LargeSG} and the batch size of 2048~(FID=7.89).} 
    \label{fig:Figure_qualitative_big_img2048}
\end{figure}

\begin{figure}[ht]
    \centering
    \includegraphics[width=0.99\linewidth]{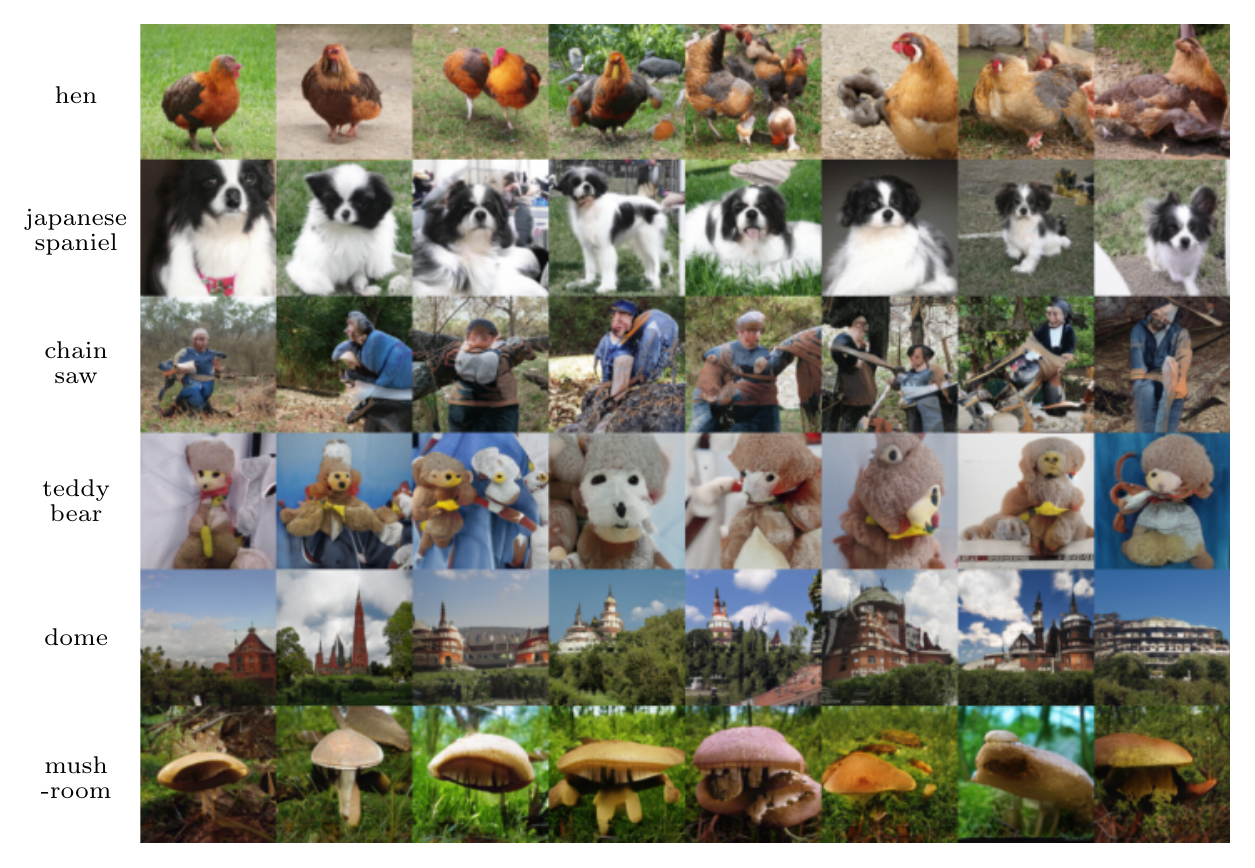}
    \hspace{-0.25cm}
    \caption{Generated images on ImageNet~\cite{Deng2009ImageNetAL} dataset using ReACGAN and the batch size of 256~(FID=13.98).} 
    \label{fig:Figure_qualitative_reac_img256}
\end{figure}

\begin{figure}[ht]
    \centering
    \includegraphics[width=0.99\linewidth]{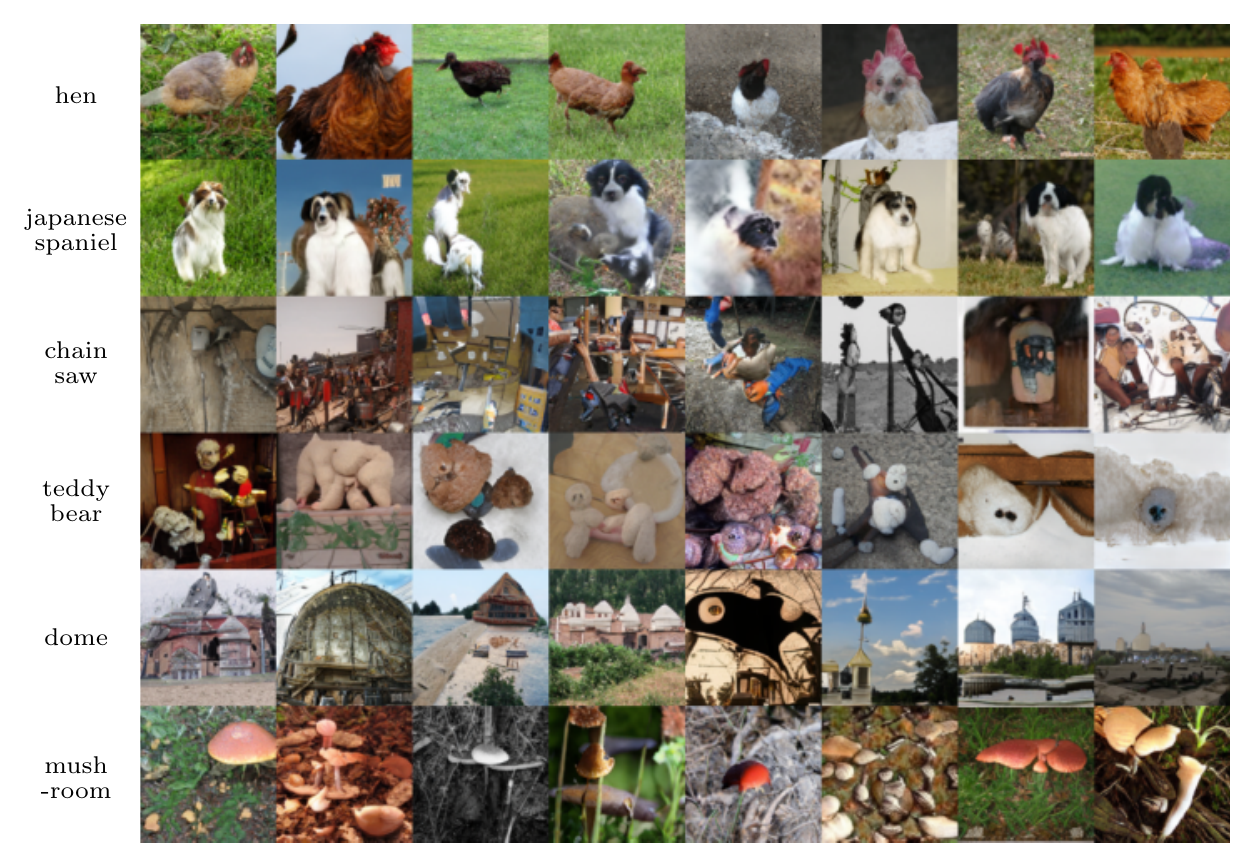}
    \hspace{-0.25cm}
    \caption{Generated images on ImageNet~\cite{Deng2009ImageNetAL} dataset using BigGAN~\cite{Brock2019LargeSG} and the batch size of 256 (FID=16.36).} 
    \label{fig:Figure_qualitative_big_img256}
\end{figure}

\begin{figure}[ht]
    \centering
    \includegraphics[width=0.99\linewidth]{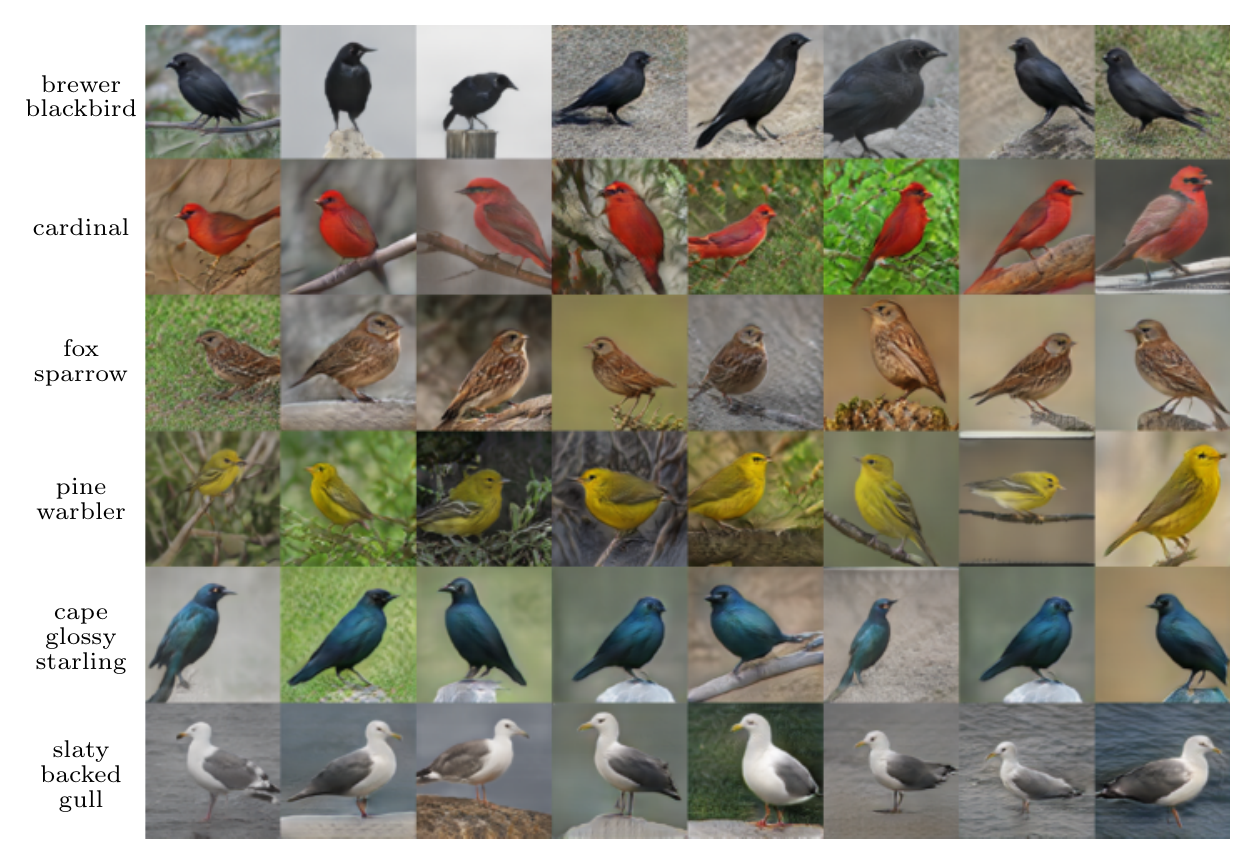}
    \hspace{-0.25cm}
    \caption{Generated images on CUB200~\cite{WelinderEtal2010} dataset using ReACGAN (FID=14.67).} 
    \label{fig:Figure_qualitative_reac_cub}
\end{figure}

\begin{figure}[ht]
    \centering
    \includegraphics[width=0.99\linewidth]{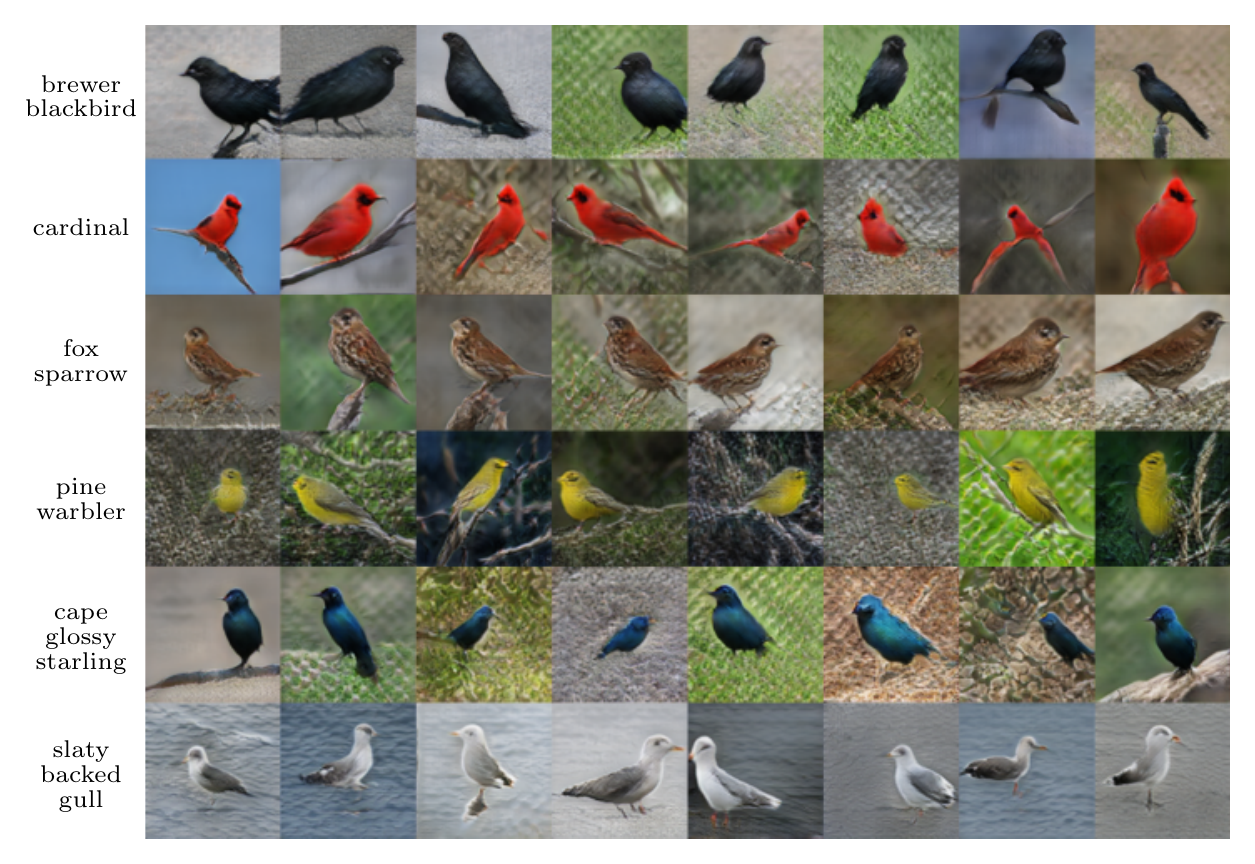}
    \hspace{-0.25cm}
    \caption{Generated images on CUB200~\cite{WelinderEtal2010} dataset using BigGAN~\cite{Brock2019LargeSG} (FID=17.80).}
    \label{fig:Figure_qualitative_big_cub}
\end{figure}

\begin{figure}[ht]
    \centering
    \includegraphics[width=0.99\linewidth]{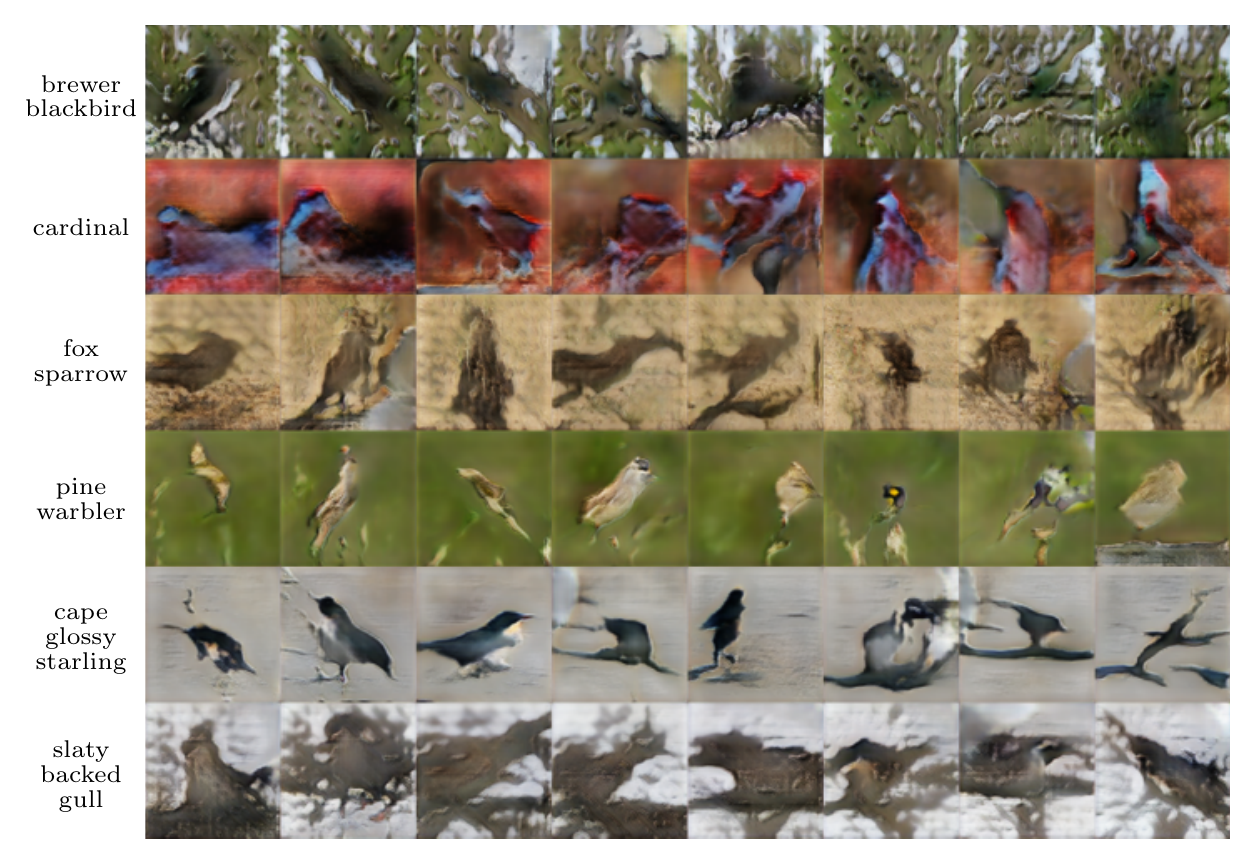}
    \hspace{-0.25cm}
    \caption{Generated images on CUB200~\cite{WelinderEtal2010} dataset using ACGAN~\cite{Odena2017ConditionalIS} (FID=61.29).}
    \label{fig:Figure_qualitative_ac_cub}
\end{figure}

\begin{figure}[ht]
    \centering
    \includegraphics[width=1.0\linewidth]{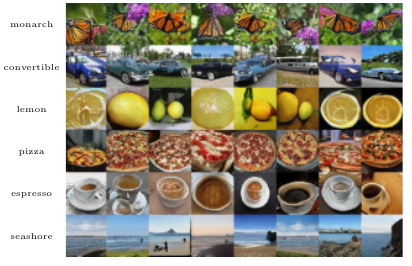}
    \hspace{-0.25cm}
    \caption{Generated images on Tiny-ImageNet~\cite{Tiny} dataset using ReACGAN (FID=26.82).} 
    \label{fig:Figure_qualitative_reac_tiny}
\end{figure}
\begin{figure}[ht]
    \centering
    \includegraphics[width=1.0\linewidth]{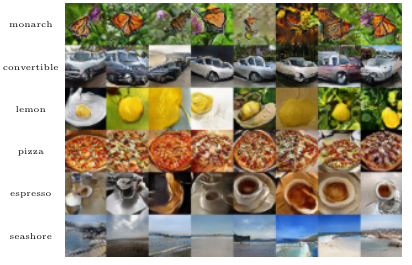}
    \hspace{-0.25cm}
    \caption{Generated images on Tiny-ImageNet~\cite{Tiny} dataset using BigGAN~\cite{Brock2019LargeSG} (FID=31.92).} 
    \label{fig:Figure_qualitative_big_tiny}
\end{figure}
\begin{figure}[ht]
    \centering
    \includegraphics[width=1.0\linewidth]{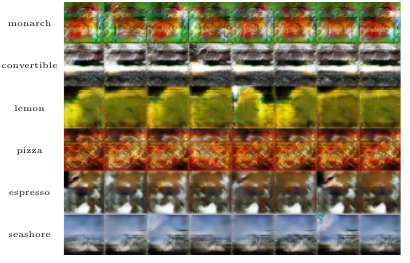}
    \hspace{-0.25cm}
    \caption{Generated images on Tiny-ImageNet~\cite{Tiny} dataset using ACGAN~\cite{Odena2017ConditionalIS} (FID=61.50).} 
    \label{fig:Figure_qualitative_ac_tiny}
\end{figure}
\clearpage
\end{document}